\documentclass[12pt,twoside]{article}

\usepackage[super,sort,comma]{natbib}
\usepackage{fancyhdr}	

\usepackage[section]{placeins} 

\usepackage{graphicx}

\makeatletter \renewcommand\@biblabel[1]{$^{#1}$} \makeatother
\newcommand{\cen}[1]{\begin{center} #1 \end{center}}

\usepackage[mathlines]{lineno}

\usepackage{hyperref}
\usepackage{xcolor}

\definecolor{gray}{rgb}{0.6,0.6,0.6}
\definecolor{red}{rgb}{0.85,0,0}
\definecolor{green}{rgb}{0,0.85,0}
\definecolor{blue}{rgb}{0,0,0.85}
\definecolor{beige}{rgb}{0.92,0.87,0.78}

\usepackage[all]{hypcap}    

\usepackage{graphicx}
\usepackage{xcolor}
\usepackage{amssymb,amsmath,amsfonts}
\usepackage{tikz}
\usepackage{multirow}
\usepackage{booktabs}

\usepackage{amssymb}
\usepackage{amsthm}
\usepackage{bm}
\usepackage{mathtools}
\usepackage{dsfont}

\usepackage{multicol}
\usepackage{tikzscale}
\usepackage{numprint}

\usepackage{float}
\usepackage{pict2e}
\usepackage{siunitx}
\usepackage[english]{babel}
\usepackage[percent]{overpic}

\newcommand{\Rd}{\mathbf{R}}                         
\newcommand{\XX}{\mathbf{x}}                         
\newcommand{\ZZ}{\mathbf{z}}                         
\newcommand{\RR}{\mathbf{r}}                         
\newcommand{\Ad}{\mathbf A}                         
\newcommand{\Fd}{\mathbf F}                         
\newcommand{\Ed}{\mathbf E} 
\newcommand{\Cd}{\mathbf C} 
\newcommand{\Id}{\mathbf I} 
\newcommand{\Hd}{\mathbf H}

\newcommand{\Au}{\mathbf{A}_I} 
  
\newcommand{\Xu}{\mathbf{x}_I}                         
\newcommand{\Yu}{\mathbf{y}_I}                         

\newcommand{\Xcnn}{\mathbf{x}_{\mathrm{CNN}}} 
\newcommand{\cnn}{{\mathrm{CNN}}}

\newcommand{\coloneqq}{{:=}}
\usepackage{graphicx}

\usepackage{xcolor}

\usepackage[percent]{overpic}

\usepackage{algpseudocode}
\usepackage{algorithm}
\usepackage{adjustbox}
\usepackage{siunitx}

\newcommand{\herm}{{\scriptstyle \boldsymbol{\mathsf{H}}}}
 \newcommand{\trans}{{\scriptstyle \boldsymbol{\mathsf{T}}}}

\begin{document}
	
\cen{\sf {\Large {\bfseries An End-To-End-Trainable Iterative Network Architecture for Accelerated Radial Multi-Coil 2D Cine MR Image Reconstruction} \\  
		\vspace*{10mm}
		Andreas~Kofler$^{1}$,
		Markus~Haltmeier$^{2}$,
		Tobias~Schaeffter$^{1,3,4}$,	
		Christoph~Kolbitsch$^{1,3}$}\\ \vspace{0.1cm} 	 
	$^{1}$Physikalisch-Technische Bundesanstalt (PTB), Braunschweig and Berlin, 10587, Germany\\
	$^{2}$Department of Mathematics, University of Innsbruck, Innsbruck, 6020, Austria\\
	$^{3}$School of Imaging Sciences and Biomedical Engineering, King's College London, London, SE1 7EH, UK\\
	$^{4}$Department of Biomedical Engineering, Technical University of Berlin, Berlin, 10623, Germany\\
	\vspace{5mm}
	Version typeset \today
}

\vspace{-0.3cm}
andreas.kofler@ptb.de

\pagenumbering{roman}
\pagestyle{plain}

\begin{abstract}
	\noindent {\bf Purpose:} Iterative Convolutional Neural Networks (CNNs) which resemble unrolled learned iterative schemes have shown to consistently deliver state-of-the-art results for image reconstruction problems across different imaging modalities. However, because these methodes include the forward model in the architecture, their applicability is often restricted to either relatively small reconstruction problems or to problems with operators which are computationally cheap to compute. As a consequence, they have so far not been applied to dynamic non-Cartesian multi-coil reconstruction problems. \\
	{\bf Methods:} In this work, we propose a CNN-architecture for image reconstruction of accelerated 2D radial cine MRI with multiple receiver coils. The network is based on a computationally light CNN-component and a subsequent conjugate gradient (CG) method which can be jointly trained end-to-end using an efficient training strategy. We investigate the proposed training-strategy and compare our method  to other well-known reconstruction techniques with learned and non-learned regularization methods.\\
	{\bf Results:} Our proposed method outperforms all other methods based on non-learned regularization. Further, it performs similar or better than a CNN-based method employing a 3D U-Net and a method using adaptive dictionary learning. In addition, we empirically demonstrate that even by training the network with only iteration, it is possible to increase the length  of the network at test time and further improve the results.  \\
	{\bf Conclusions:} End-to-end training allows to highly reduce the number of trainable parameters of and stabilize the reconstruction network. Further, because it is possible to change the length of the network at test time, the need to find a compromise between the complexity of the CNN-block and the number of  iterations in each CG-block becomes irrelevant.
	\end{abstract}

\newpage     

\tableofcontents

\newpage

\setlength{\baselineskip}{0.7cm}      

\pagenumbering{arabic}
\setcounter{page}{1}
\pagestyle{fancy}

\section{Introduction}
Magnetic Resonance Imaging (MRI) is an important tool for the assessment of different cardiovascular diseases. Cardiac cine MRI, for example, is used to assess the cardiac function  as well as left and right ventricular volumes and left ventricular mass \cite{puntmann2018society}.
However, MRI is also known to suffer from relatively long data-acquisition times. In cardiac cine MRI, the data-acquisition usually has to take place during a single breathhold of the patient in order to avoid artefacts arising from the respiratory motion. Since for patients with limited breathhold capabilities this can be challenging, undersampling techniques can be used to accelerate the measurement process. Undersampling  in Fourier-domain, the so-called $k$-space, leads to a violation of the Nyquist-Shannon sampling theorem and therefore, regularization techniques must be used to reconstruct artefact-free images.\\
Recently, image reconstruction has experienced a paradigm shift with the re-emergence of Neural Networks (NNs), and in particular, Convolutional NNs (CNNs) which can be employed as regularization methods \cite{wang2018image}. CNNs can be used in different ways as regularization methods for image reconstruction problems. A key element to categorize these methods is wheter or not the forward operator is used in the learning process, see  \cite{ongie2020deep}.  A straight-forward approach is to simply post-process an initial estimate of the solution which is impaired by noise and artefacts, see e.g. \cite{jin2017deep}, \cite{sandino2017deep}, \cite{Hauptmann2019}, \cite{kofler2019spatio}. Further, cross-/multi-domain methods which pre-process the $k$-space and subsequently process the initially obtained reconstruction have been investigated as well \cite{el2020multi}. 
However, the post-processed image might lack data-consistency, meaning that it is not clear how well the post-processed image matches the measured $k$-space data. Thus, approaches which used the output of a pre-trained CNN as image-priors have been considered as well, see e.g.\ \cite{hyun2018deep}, \cite{kofler2020neural}.
Thereby, the CNN-priors are used in the formulation of a Tikhonov functional which is subsequently minimized to increase the data-consistency of the solution. These methods have the (computational) advantage of decoupling the regularization step from increasing data-consistency and thus are in general applicable to arbitrary large-scale image reconstruction problems as well.
However, the network-training is typically carried out in the absence of the forward model and although this strategy was reported to be successful, image reconstruction methods based on NNs have been reported to possibly suffer from instabilities \cite{Antun201907377}. This issue could directly affect  the obtained CNN-prior and thus, directly affect the quality of the final reconstruction, since its solution depends on the CNN-prior. Interestingly, in \cite{Antun201907377}, the authors showed that the CNN-based reconstruction methods which include the forward and adjoint operators in the network architecture are the most stable with respect to small perturbations and adversarial attacks.  Further, including the  forward model  in the network architecture  has been reported to lower the maximum error-bound of the CNNs \cite{maier2019learning}. Thus, it is desirable to find a way to include the physical model in the CNN-architecture, even for computationally expensive/large-scale problems.\\
Methods in which the CNNs architectures resemble unrolled iterative schemes of finite length are referred to as variational/iterative/cascaded networks, see e.g.\ \cite{schlemper2017deep}, \cite{hammernik2018learning}, \cite{kobler2017variational}, \cite{qin2018convolutional}, \cite{aggarwal2018modl}, \cite{qin2019k}, \cite{gilton2019neumann}. Thereby, these methods typically consist of CNN-blocks  and data-consistency (DC)-blocks, which are given in the form of gradient-steps with respect to a data-fidelity term, see e.g.\ \cite{hammernik2018learning}, \cite{gilton2019neumann}, or as layers which implement the minimizer  of a functional involving a data-fidelity term, see e.g.\ \cite{schlemper2017deep}, \cite{qin2018convolutional}, \cite{qin2019k}. While the CNN-blocks can be interpreted as learned regularizers, the DC-blocks utilize the measured undersampled $k$-space data to increase/ensure the data-consistency of the intermediate CNN-outputs.\\
Unfortunately, including the forward and adjoint operators in the CNN can at the same time represent  the computational bottleneck of these methods, since the forward operator as well as its adjoint can be computationally expensive to evaluate. As a consequence, the applicability of iterative networks is currently still limited to either reconstruction problems with easy-to-compute forward operators, e.g.\ a  FFT-transform sampled on a Cartesian grid \cite{schlemper2017deep}, \cite{qin2018convolutional},\cite{qin2019k}, \cite{kustner2020cinenet} or to non-dynamic problems with non-Cartesian sampling schemes, see \cite{schlemper2019nonuniform} for a single-coil data-acquisition or \cite{malave2020reconstruction} for a multiple coil-acquisition. For example, in \cite{schlemper2017deep}, \cite{qin2018convolutional}, \cite{qin2019k}, only a single-coil is used for the encoding operator. In \cite{kustner2020cinenet}, multiple receiver coils were used for a dynamic 3D reconstruction problem with a Cartesian sampling grid. For non-dynamic problems, on the other hand, iterative CNNs for multi-coil data-acquisition on a Cartesian grid have been received more attention, see e.g.\ \cite{hammernik2018learning}, \cite{kobler2017variational}, \cite{duan2019vs}.\\
Acquisition protocols using non-Cartesian sampling trajectories such as radial or spiral sampling, can be an attractive alternative to standard Cartesian acquisitions, especially because the arising undersampling artefacts are much more incoherent compared to the ones arising from a Cartesian acquisition, see \cite{lustig2008compressed}.
However, radially acquired $k$-space data is computationally more demanding to reconstruct as the forward encoding operator involves gridding on a Cartesian grid before the fast Fourier-transform can be applied, see \cite{smith2019trajectory} for more details. 
In \cite{schlemper2019nonuniform}, an iterative network for a non-dynamic radial single-coil acquisition was proposed. However, the standard in the clinical routine, is in fact given by multi-coil data-acquisitions  \cite{knoll2020deep} through which the acquisition of the $k$-space data increases in terms of  computational complexity. Finally, acquiring a dynamic process is inherently computationally more demanding as for each time point, the encoding operator has to be applied to the image.\\
In this work, we present a novel computationally light and efficient CNN-block architecture as well as a training strategy which are tailored to image reconstruction of dynamic multi-coil radial MR data of the heart. The  combination of the proposed network architecture and training strategy allows to train the entire reconstruction network in an end-to-end manner, even for dynamic problems with non-uniformly sampled data as well as multiple receiver coils. To the best of our knowledge, this is the first work which overcomes these computational difficulties and presents such an end-to-end trainable network architecture for dynamic non-Cartesian multi-coil MR reconstruction problems.\\
The paper is structured as follows: In Section \ref{sec:mat_n_methods}, we formally introduce the reconstruction problem and the proposed method by discussing in more detail the used CNN-block as well as the training strategy. In Section \ref{sec:results}, we show extensive experiments to validate the efficacy of our approach and compare it to different state-of-the-art methods for dynamic cardiac radial MRI. We then discuss the main advantages and limitations of our work in Section \ref{sec:discussion} and conclude the work in Section \ref{sec:conclusion}.

\section{Materials and Methods}\label{sec:mat_n_methods}

\subsection{Problem Formulation}
Let $\XX \in \mathbb{C}^N$ with $N=N_x \times N_y \times N_t$ denote the vector representation of a complex-valued cine MR image, i.e.\ $\XX = [\XX_1, \ldots, \XX_{N_t}]^\trans$. The forward operator $\Ad$ maps the dynamic cine MR image to its corresponding $k$-space. In this work, we focus on a 2D radial encoding operator using multiple receiver coils. More precisely, the operator $\Ad$ is given by
\begin{equation}\label{multi_coil_prblem}
\Ad\coloneqq (\Id_{N_c} \otimes \Ed) \Cd,
\end{equation}
where $\Id_{N_c}$ denotes the identity operator and $\Cd$ consists of the concatenation of the $N_c$ different coil-sensitivity maps which are multiplied with the cine MR image, i.e.\ $\Cd =  [\Cd_1,\ldots,\Cd_{N_c}]^\trans$, with $\Cd_j = \mathrm{diag}(\mathbf{c}_j,\mathbf{c}_j,\ldots,\mathbf{c}_j) \in \mathbb{C}^{N \times N}$ and $\mathbf{c}_j \in \mathbb{C}^{N_x \times N_y}$. The operator $\Ed = \mathrm{diag}(\Ed_1, \ldots, \Ed_{N_t})$ denotes a composition of 2D radial encoding operators $\Ed_t$ which for each temporal point $t \in \{1,\ldots, N_t\}$ sample a 2D image $ \XX_t \in \mathbb{C}^{N_x \times N_y}$ along a non-Cartesian grid in the Fourier domain.
In particular, for this work, we consider trajectories given by the radial golden-angle method \cite{winkelmann2006optimal} but point out that other trajectories can be considered as well.
By $J=\{1,\ldots,N_{\mathrm{rad}}\}$ we denote the set of indices the $k$-space coefficients which are needed to sample a 2D image $\XX_t$ at Nyquist limit. 
In order to accelerate the image acquisition process, the set $J$ is typically constructed by reducing the number of radial lines. We denote by $\Au$ the radial Fourier encoding operator which samples a complex-valued 2D cine MR image $\XX$ at a radial grid given by the indices in the set $I = I_1 \cup \ldots \cup I_{N_t}$ with $I_t \subset J$ for all $t=1,\ldots,N_t.$ Thus, the considered reconstruction problem is given by 
\begin{equation}\label{reco_problem}
\Au \XX + \mathbf{e}= \Yu,
\end{equation}
where $\Yu =[\Yu^1, \ldots, \Yu^{n_c}]^\trans$ with $\Yu^c \in \mathbb{C}^{N_t \cdot m_{\mathrm{rad}}}$   denotes the  measured $k$-space data for the dynamic 2D cine MR image $\XX$. Multiple receiver coils are typically used in order to achieve $m_{\mathrm{rad}} > N_{\mathrm{rad}}$  such that problem  \eqref{reco_problem} is overdetermined and can be solved by considering the normal equations
\begin{equation}\label{normal_equations}
\Au^\herm \Au \XX = \Au^\herm \Yu.
\end{equation}
System \eqref{normal_equations} could in principle be solved by conjugate gradient (CG)-like methods yielding an approximation of the solution given by
\begin{equation}\label{pseudo_inv_solution}
\XX^{\ast} = (\Au^\herm \Au)^{-1}\Au^\herm \Yu,
\end{equation}
where the configuration of the receiver coils ensures that the operator $\Au^\herm \Au$ is invertible.  However, for non-Cartesian trajectories, problem \eqref{normal_equations} is ill-conditioned. 
Thus, methods used to approximate $\XX^{\ast}$ can exhibit a semi-convergence  behavior and lead to undesired noise-amplification \cite{qu2005convergence}. 
Hence, regularization techniques must be used to stabilize the inversion process.\\
In this work, we propose a reconstruction network based on a fixed-point iteration which has the form
\begin{equation}\label{iterative_CNN}
u_{\Theta}^M = \underbrace{\big( f_{\lambda}^{\mathrm{DC}} \circ u_{\Theta} \big)  \circ \ldots \circ \big( f_{\lambda}^{\mathrm{DC}} \circ u_{\Theta} \big)}_{M\,\, \mathrm{times}}= \big( f_{\lambda}^{\mathrm{DC}} \circ u_{\Theta} \big)^M,
\end{equation}
where $u_{\Theta}$ is a CNN-block which reduces undersampling artefacts and noise and $f^{\mathrm{DC}}$ is a module which increases data-consistency of the intermediate outputs of the CNN-blocks. Limits of \eqref{iterative_CNN} simultaneously  satisfy data-consistency induced by the DC module and regularity induced by the CNN-block.  

\begin{figure}
\includegraphics[width=\linewidth]{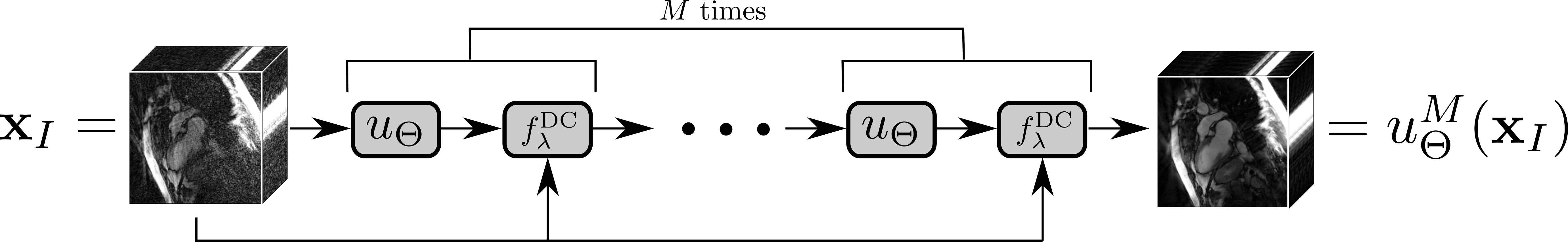}
\caption{The proposed reconstruction network alternates between the application of CNN-blocks $u_{\Theta}$ and and data-consistency blocks $f^{\mathrm{dc}}$ which increase the consistency of the outputs of the intermediate CNN-blocks by making use of the measured data which is implicitly given by the initial reconstruction $\XX_I$.}\label{proposed_cascade_fig}
\vspace{0.2cm}
\end{figure}

For example, in the simplest case where $f^{\rm DC}_{\lambda}$ is defined as the minimizer of a penalized least squares functional (see \eqref{CNN_reg_theta_fixed_problem}) and  where $u_{\Theta}$ defines a linear projection, the fixed-point iteration \eqref{iterative_CNN} can be shown to converge to the minimizer of the functional 
\begin{equation}\label{CNN_reg_problem}
F_{\lambda,\Yu}(\XX)= \| \Au \XX - \Yu \|_2^2 + \lambda\,\| \XX-u_{\Theta}(\XX)\|_2^2.
\end{equation}
Note that,  similar to \cite{aggarwal2018modl} but in contrast to \cite{schlemper2017deep}, \cite{qin2018convolutional}, \cite{hammernik2018learning}, the set of parameters $\Theta$ is the same for each CNN-block and therefore in general allows a repeated application of the iterative network $u_{\Theta}^M$. Further, the proposed CNN-block differs from the one in \cite{aggarwal2018modl}, as we shall discuss later. Figure \ref{proposed_cascade_fig} shows an illustration of the described reconstruction network.
In the following, we describe the CNN-block $u_{\Theta}$ and the DC-module $f_{\lambda}^{\mathrm{DC}}$ in more detail.

\subsection{Proposed CNN-Block}
\begin{figure}[t]
\centering
\includegraphics[width=\linewidth]{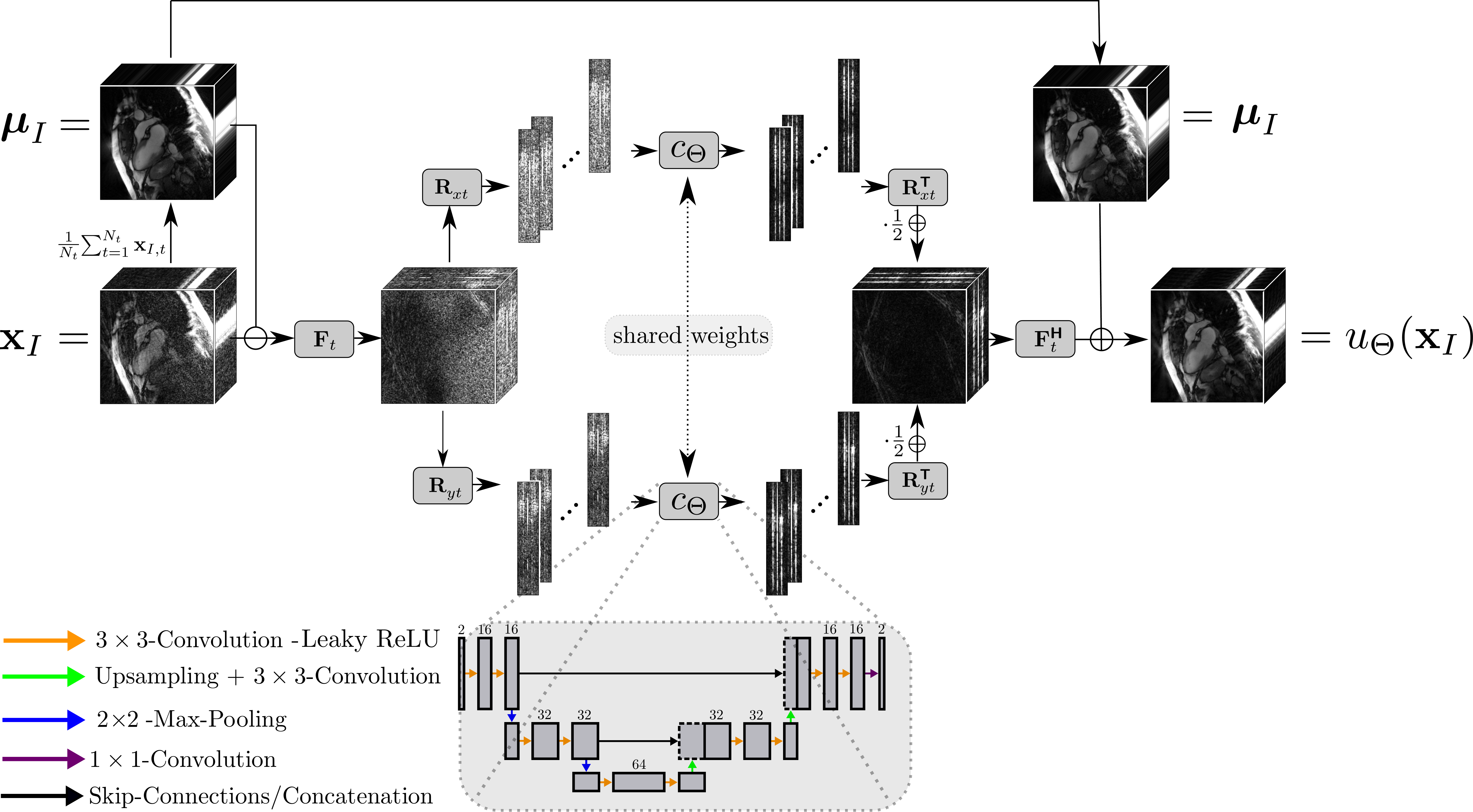}
\caption{The proposed CNN-block. First, the input image $\Xu$ is Fourier-transformed along the temporal axis, then the Fourier-transformed image is reshaped into the $xt$- and $yt$-domain. The same 2D U-Net $u_{\Theta}$ is applied to each of the 2D slices. Then, the estimate in the temporal Fourier-domain is calculated by reassembling the processed slices and the output is obtained by applying the inverse temporal Fourier-transform and summing it up to the input image. }\label{XTYTFFT_cnn_fig}
\vspace{1cm}
\end{figure}

In order to keep the notation as simple as possible, we neglect the indices referring to the iteration within the network. We illustrate the CNN-block for the first iteration of the network, i.e.\ when the input of the CNN-block is the initial non-uniform FFT (NUFFT)-reconstruction. The process described below is employed within every CNN block in the proposed network architecture.
Let $\mathbf{W}$ denote a diagonal operator which contains the entries of the density compensation function in $k$-space. By $\Xu\coloneqq \Au^{\sharp} \Yu\coloneqq \Au^\herm \mathbf{W} \Yu$ we denote the initial NUFFT-reconstruction which is obtained by re-gridding the measured   and density compensated $k$-space coefficients onto a Cartesian grid and applying the inverse FFT (IFFT). First,  we compute a temporal average of the input over all cardiac phases, i.e.\
\begin{equation}
\boldsymbol{\nu}_I = \frac{1}{N_t}\sum_{t=1}^{N_t} \XX_{I,t} \in \mathbb{C}^{N_x \times N_y}
\end{equation}
and stack it along the temporal axis, i.e.\ $\boldsymbol{\mu}_I= [\boldsymbol{\nu}_I,\ldots, \boldsymbol{\nu}_I] \in \mathbb{C}^{N_x \times N_y \times N_t}$. Then, we subtract $\boldsymbol{\mu}_I$ from the initial NUFFT-reconstruction $\XX_I$ and  apply a temporal Fourier-transform $\Fd_t$, i.e.\ we obtain $\ZZ= \Fd_t (\Xu-\boldsymbol{\mu}_I)$. 
Subtracting the temporal average from image frames is a well-known and established approach to sparsify  image sequences, for example in video-compression, see e.g.\ \cite{le1991mpeg}.
Further, applying a temporal FFT provides an even sparser representation and has been used for  example  in \cite{Tsao2003}, \cite{qin2019k} for dynamic MR image reconstruction. Thus, the CNN-block learns to reduce the present undersampling artefacts and noise in a sparse domain. We define the operators $\Rd^{xt}$ and $\Rd^{yt}$ which rotate $\ZZ$ by appropriately permuting the relevant axes of the images such that
\begin{eqnarray}\label{input_rl}
\ZZ^{xt} &=  \Rd^{xt} \ZZ \in \mathbb{C}^{N_y \times N_x \times N_t},\\
\ZZ^{yt} &=  \Rd^{yt} \ZZ \in \mathbb{C}^{N_x \times N_y \times N_t}.
\end{eqnarray}
At this point, $\ZZ^{xt}$ and $\ZZ^{yt}$ can be interpreted as $N_y$ complex-valued images of shape $N_x \times N_t$ and  $N_x$ complex-valued images of shape $N_y \times N_t$, respectively. Then, a simple 2D U-Net \cite{Ronneberger2015}, which we denote by $c_{\Theta}$ is applied both  to $\ZZ^{xt}$ and $\ZZ^{yt}$. Note that we use the same U-Net for both $xt$- and $yt$-branches, i.e.\ the weights are shared among the two. Also, note that because the U-Net only consists of convolutional and max-pooling layers and has no fully-connected layers, the approach can be used for the case $N_x \neq N_y$ as well. Then, we  obtain
\begin{eqnarray}\label{input_rl}
\ZZ_{\mathrm{CNN}}^{xt} &=  c_{\Theta}(\ZZ^{xt})\\
\ZZ_{\mathrm{CNN}}^{yt} &=  c_{\Theta}(\ZZ^{yt}).
\end{eqnarray}
and calculate the estimate $\ZZ_{\cnn}$ by reassembling the processed spatio temporal slices, i.e.\
\begin{equation}\label{zcnn}
\ZZ_{\cnn} =  \frac{1}{2} \big((\Rd^{xt})^\trans \ZZ_{\cnn}^{xt} + (\Rd^{yt})^\trans\ZZ_{\cnn}^{yt} \big).
\end{equation}
The factor $1/2$ in \eqref{zcnn} is needed because every pixel is used twice, once in the $xt$-branch and once in the $yt$-branch. Finally, the estimate $\Xcnn$ is given by applying $\Fd_t^\herm$ and adopting a residual connection, i.e.

\begin{equation}
\Xcnn = \Fd_t^\herm \ZZ_{\cnn} + \boldsymbol{\mu}_I.
\end{equation}
Figure \ref{XTYTFFT_cnn_fig} shows an illustraton of  the just described procedure. The 2D U-Net employed in our proposed network consists of three encoding stages which are interleaved by max-pooling layers. Each stage consists of a block of 2D convolutional layers with Leaky ReLU as activation function. The  number of initially extracted feature maps which is doubled after each max-pooling layer is $n_f=16$ and is on purpose chosen to be relatively small in order to keep the model's complexity as low as possible. The reason for that is that the 2D U-Net will have to be used together with a (computationally expensive) CG-block, which we describe in the next Subsection. In the decoding path, bilinear upsampling followed by a $3\times 3$ convolutional layer with no activation function is used to upsample the images. Our 2D U-Net also involves skip connections from the last to the first layers of the corresponding stages in the encoding and the decoding path. The residual connection is performed in image domain rather than in Fourier-domain.
\subsection{DC-Module}
We illustrate the DC-module for the first iteration of the network, i.e.\ where the incoming input of the CNN-block is given by the initial reconstruction. For later iterations, the construction is analogous. 
Let us fix $\Xcnn = u_{\Theta}(\Xu)$ and consider 
\begin{equation}\label{CNN_reg_theta_fixed_problem}
F_{\lambda,\Yu}(\XX)= \| \Au \XX - \Yu \|_2^2 + \lambda\, \|  \XX - \Xcnn\|_2^2.
\end{equation}
For the special case where the operator $\Ad$ is an isometry, i.e.\ $\| \Ad\|_Y = 1$, where $\| \cdot \|_Y$ denotes the operator-norm of $\Ad$, problem \eqref{CNN_reg_theta_fixed_problem} has a closed-form solution, see e.g.\ \cite{schlemper2017deep}, \cite{qin2018convolutional}. 
A simple example of  $\Ad$ being an isometry is given by a single-coil acquisition on a Cartesian grid using a simple FFT. In this case, the solution of \eqref{CNN_reg_theta_fixed_problem} can be obtained by performing a linear combination of the measured $k$-space data $\Yu$ and the one estimated by applying $\Au$ to $\Xcnn$. See e.g.\ \cite{schlemper2017deep}, for more details. However, in the more general case, a minimizer of \eqref{CNN_reg_problem} is obtained by  solving a system of linear equations.
By setting the derivative of \eqref{CNN_reg_theta_fixed_problem} with respect to $\XX$ to zero, it can be easily seen that the system one needs to solve is given by $\Hd \XX = \mathbf{b}$ with
\begin{eqnarray}\label{lin_system}
\Hd = \Au^\herm\Au + \lambda\, \Id, \nonumber \\
\mathbf{b} = \Au^\herm\Yu + \lambda\, \Xcnn.  
\end{eqnarray}
System \eqref{lin_system} can be solved by means of any iterative algorithm and, since the operator $\Hd$ is symmetric, an appropriate choice is the conjugate gradient (CG) method \cite{hestenes1952methods}. 
Note that functional \eqref{CNN_reg_theta_fixed_problem} is linear in $\XX$ and therefore, due to strong-convexity, solving \eqref{lin_system} leads to the unique minimizer of \eqref{CNN_reg_theta_fixed_problem}. In practice, as we discuss later in the implementation details, the DC-module is an implementation of a finite number of iterations of the CG-method. We denote the number of such iterations by $n_{\mathrm{CG}}$.
Note that in the CG method, the operator $\Hd = \Au^\herm \Au$ has to be applied at each iteration.
In addition, note that in general, the application of $\Au$ as well as $\Au^{\herm}$ can be computationally way more expensive than for a simple FFT, for example if the gridding of the $k$-space coefficients is part of the operators as in our case. Thus, it is desirable to have a CNN-block with as few trainable parameters as possible such that end-to-end training of the entire network is possible in a reasonable amount of time.\\

\subsection{Training Scheme} 
Because the solution of \eqref{lin_system} has to be approximated  using an iterative scheme which employs the (computionally expensive) application of the operator $\Hd$ at each iteration, end-to-end training of  the entire reconstruction network from scratch would be time consuming.
Thus, we circumvent this issue  by the following more efficient training strategy.\\
First, in a pre-training step, we only train a single CNN-block on image pairs as is typically done in Deep Learning-based post-processing methods. More precisely, we minimize the $L_2$-norm of the error between the output of the CNN-block which was predicted from the initial reconstruction $\Xu$ and its corresponding label. Then, in a second training-stage, we construct a network as in \eqref{iterative_CNN} and initialize each of the CNN-blocks by the previously obtained parameter set $\Theta$ and perform a further fine-tuning of the entire network by end-to-end training. 
Further, the regularization parameter $\lambda$ contained in each fo the CG-blocks $f_{\lambda}^{\mathrm{DC}}$ can be included in the set of trainable parameters as well and is trained to find the optimal strength of the contribution of the regularization term $\mathcal{R}$. This means that it implicitly learns to estimate the noise-level present in the measured $k$-space data for the whole dataset.\\

\subsection{Experiments with In-Vivo Data}\label{subsec:experiments}

To evaluate the efficacy of our proposed approach, we performed different experiments. First, we investigated the effect of our proposed training strategy. We also compared the proposed network with different configurations of hyper-parameters $M$ and $n_{\mathrm{CG}}$. Finally, we further compared our proposed approach to several other methods for non-Cartesian cardiac cine MR image reconstruction. In the following, we provide the reader with  information about the used dataset, the methods of comparison and some details on the implementation of the proposed method.

\subsection{Dataset}

We used a dataset of cine MR images of  $n=19$ subjects (15 healthy volunteers + 4 patients). For each healthy volunteer as well as for two patients, $N_z=12$ different orientations of cine MR images were acquired. For the resting two patients, only $N_z=6$ slices were acquired due to limited breathhold capabilities. Thus, we have a total of 216 complex-valued cine MR images of shape $N_x \times N_y \times N_t = 320 \times 320 \times 30$. We split the dataset in 144/36/36 images used for training, validation and testing, where for the test set, we use the 36 cine MR images of the patients in order to be able to qualitativey assess the images with respect to clinically relevant features.
All images were acquired using a bSSFP sequence on a 1.5\,T MR scanner during a single breathhold of 10 seconds. The images used as ground-truth data for the retrospective undersampling were reconstructed from the $k$-space data which was sampled along $N_{\theta}=3400$ radial spokes with $kt$-SENSE \cite{feng_mrm_2012}. 
From these images, we retrospectively simulated $k$-space data by sampling $N_{\theta}=560$ and $N_{\theta}=1130$ radial spokes. Since sampling along $N_{\theta}=3400$   already corresponds to an acceleration factor of approximately $\sim 3$ (with respect to the Nyquist-limit) which was needed to perform the scan during one breathhold, $N_{\theta}=560$  and $N_{\theta}=1130$ correspond to acceleration factors of $\sim 9$ and $\sim 18$, respectively.\\
Further, the $k$-space data was corrupted by normally distributed noise with standard variation $\sigma=0.02$ which was added to each the real and imaginary parts of $\Yu^c$ for $c=1,\ldots,12$ after having centred them.
The calculation of the density compensation function for a fixed set of trajectories was based on partial Voronoi diagrams \cite{malik2005gridding}. The coil sensitivity maps were estimated from the fully sampled central $k$-space region.

\subsection{Quantitative Measures}

We evaluated the performance of our proposed network architecture in  terms of different error- and image-similarity-based measures. The peak signal-to-noise ratio (PSNR) and the normalized root mean squared error (NRMSE) are used as error-based measures. Further, we report a variety of different similarity-based measures:  the structural similarity index measure (SSIM) \cite{wang2004image}, the multi-scale SSIM (MS-SSIM) \cite{wang2003multiscale}, the universal image quality index (UQI) \cite{wang2002universal}, the visual information quality measure (VIQM) \cite{sheikh2006image} and the Haar wavelet-based similarity index measure \cite{reisenhofer2018haar}.\\
We calculate all measures by comparing the 2D complex-valued images in the $xy$-plane for each time point.
This means that our test set consists of 1080 2D images. For the similarity measures, the real and imaginary part of the images are treated as channels and the measures are averaged over the two channels. The statistics were calculated over a region of interest of $160 \times 160$ pixels in order to discard background noise. Further, we segmented the patients in the images of the test set in order for the statistics to reflect the achieved performance on regions of interest.

\subsection{Implementation Details}\label{subsec:implementation_details}
The architecture was implemented in \texttt{PyTorch}. Complex-valued images were stored as two-channel- images. The forward and the adjoint operator $\Au$ and $\Au^\herm$ were implemented using the publicly  available library \texttt{Torch KBNUFFT} \cite{Muckley2019}, \cite{Muckley2020} which also allows to perform back-propagation across the forward and adjoint operators. During training, the $k$-space trajectories, the coil-sensitivity maps as well as the density compensation functions were stored as tensors for the implementation of $\Au$ and $\Au^\herm$.
In \cite{schlemper2017deep}, \cite{qin2018convolutional}, \cite{kofler2018u}, for example, where closed-formulas are given for the DC-block, the measured $k$-space data as well as the masks are used as inputs to the DC-blocks. 
Note that we make the regularization parameter $\lambda$ trainable such that a trade-off between the measured $k$-space and the output of the CNN-blocks is learned. In order to constrain  $\lambda$ to be strictly positive, during the fine-tuning of the model, we apply a Softplus activation-function, i.e.\ we set $\lambda:=\mathrm{SoftPlus}(\tilde{\lambda})= \frac{1}{\beta}(\mathrm{log}(1 + \mathrm{exp}(\beta \tilde{\lambda}))$, which maps $\tilde{\lambda}$ to the interval $(0,\infty)$. We used the  default parameter $\beta=1$. The implementation of the encoding operators $\Au$, $\Au^\herm$ as well as the proposed CNN-block $u_{\Theta}$ and the entire reconstruction network $u_{\Theta}^M$ will be made available on \url{https://github.com/koflera/DynamicRadCineMRI} after peer-review. 
Note that a CG scheme is usually stopped after a certain stopping-criterion is met. Typically, a commonly used stopping criterion is if the norm of the newly calculated residual $\RR_k$ is small enough, i.e.\ $\| \RR_k\||_2 \leq \texttt{TOL} \|\mathbf{b}\|_2$ for a tolerance $\texttt{TOL}$ chosen by the user. During fine-tuning the iterative network, we fix the number of CG iterations $n_{\mathrm{CG}}$ but when testing the network on unseen data, we can choose to use the number of iterations the CNN was fine-tuned with or set an own stopping-criterion. All experiments were performed on an NVIDIA GeForce RTX 2080 with 11 GB memory.
\subsection{Comparison to Other Methods}
We compared our proposed approach to the following methods which employ recently published learned and well-established non-learned regularization methods. As well-established reconstruction methods, we applied iterative SENSE \cite{pruessmann2001advances}, a Total Variation (TV)-minimization method \cite{block2007} and $kt$-SENSE \cite{Tsao2003}. Further, we compared our proposed method to a method based on dictionary learning (DL) and sparse coding (SC) \cite{wang2014compressed},\cite{caballero2014dictionary} using adaptive DL and adaptive SC \cite{pali2020} and a method which employs CNN-based regularization in the form of previously obtained image priors \cite{hyun2018deep},\cite{kofler2020neural}, where we used a previously trained 3D U-Net \cite{Hauptmann2019} for obtaining the prior.\\
Note that on purpose we did not compare our proposed  approach to other CNN-based methods involving generative adversarial networks (GANs). The reason is that we are mainly interested in the performance of the combination of the proposed CNN-block  in terms of artefacts-reduction as well as the trade-off between employing a CG-module or not. 
In addition, note that if the hardware allows it, it is always possible to add different components based on GANs to regularize the output of the CNN-blocks.
Further, note that although there exist several other state-of-the-art methods using cascaded/iterative networks for dynamic cine MRI, see e.g.\ \cite{schlemper2017deep}, \cite{qin2018convolutional}, \cite{qin2019k}, the underlying reconstruction problem is a different one (single-coil and Cartesian vs. multi-coil and radial) and thus the methods are not directly applicable as  originally published. \\

\section{Results}\label{sec:results}

In the following, we report the obtained results concerning the training behavior and the reconstruction performance of our proposed method.

\subsection{Computational Complexity of the Forward and Adjoint Operators}
Here, we evaluated the computational complexity of the proposed network architecture in terms of required GPU memory as well as training times which can be estimated for the end-to-end training stage.
Here, we fixed the number of iterations of the network to be $M=1$ and the CNN-block  was fixed to be the identity $c_{\Theta}=\Id$, i.e.\ it contains no trainable parameters. Thus, the allocated memory can be mainly attributed to the tensors needed for the radial trajectories, the density compensation, the coil-sensitivity maps which define the operator and the considered input image.\\
\begin{figure*}
\centering
\includegraphics[width=0.35\linewidth]{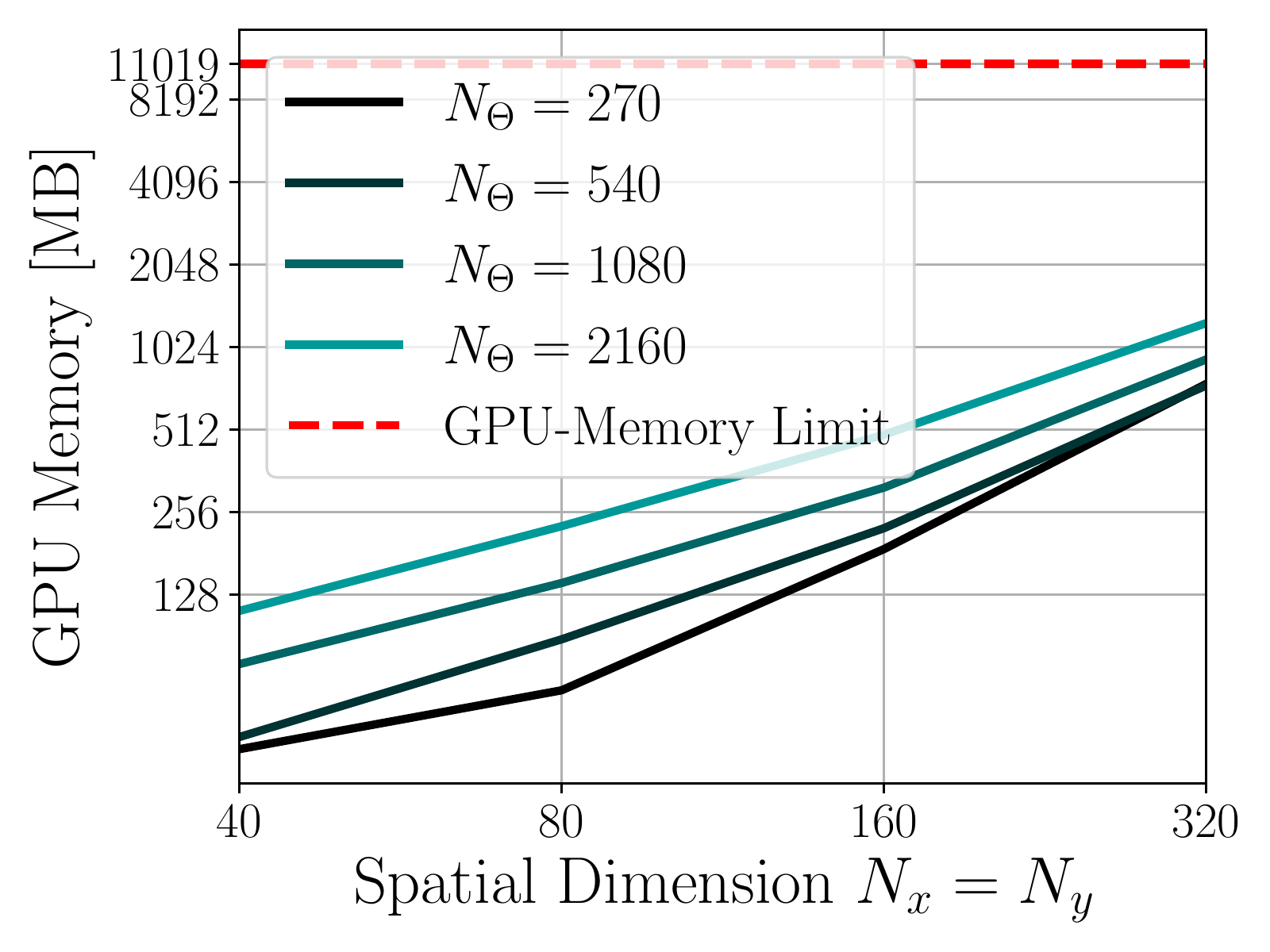}
\includegraphics[width=0.35\linewidth]{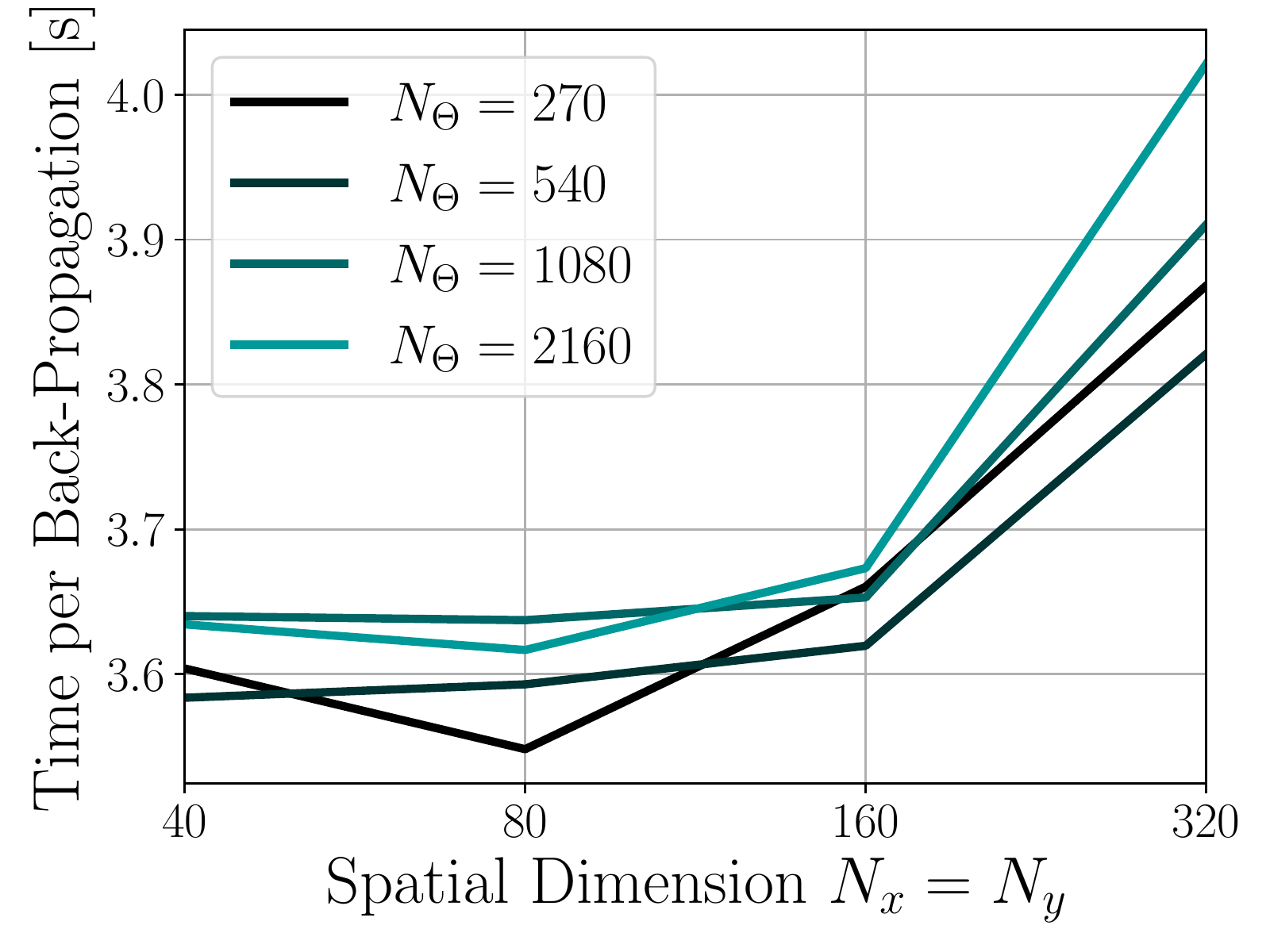}\\
\includegraphics[width=0.35\linewidth]{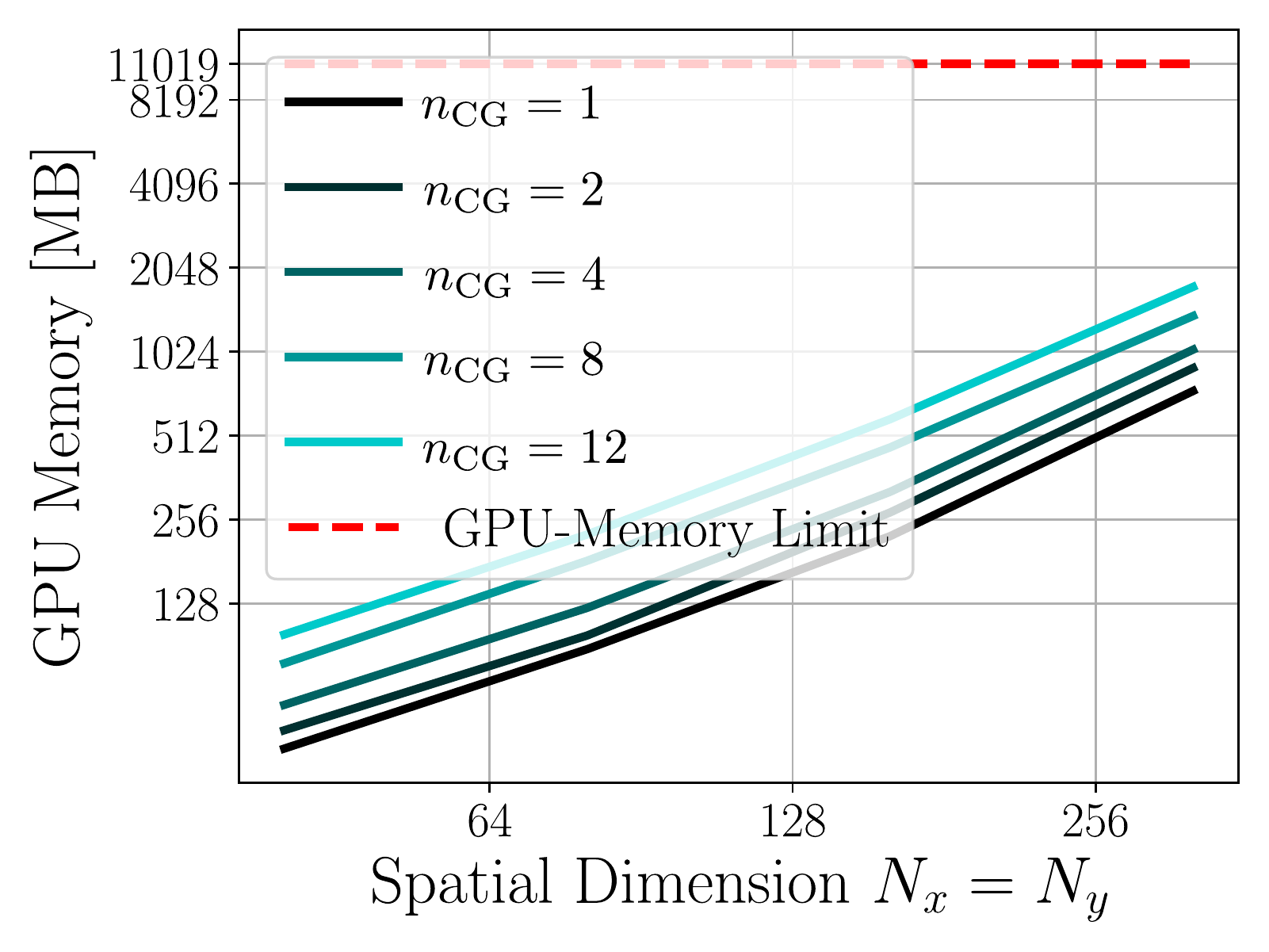}
\includegraphics[width=0.35\linewidth]{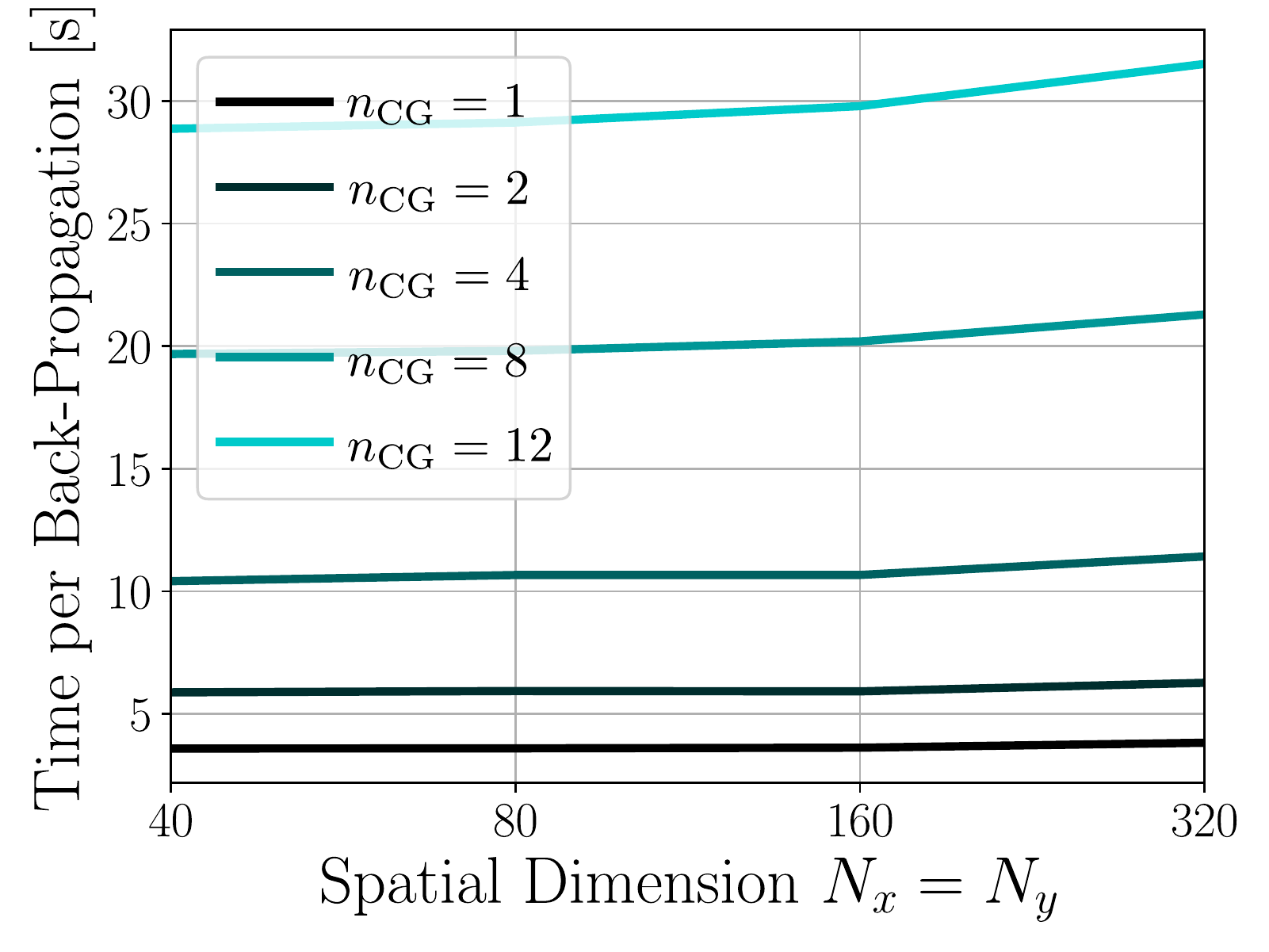}
\caption{Computational complexity and the time required for one back-ward pass for the operator $\Hd + \lambda \, \Id$ which is employed in the DC-module for different spatial image sizes $N_x \times N_y$. The number of temporal points was set to $N_t=30$ and the number of coils was $N_c=12$.}\vspace{0.4cm}\label{computational_times}
\end{figure*}
Figure \ref{computational_times} shows the allocated GPU memory as well as the required time to perform one weights-update by performing a forward- and a backward-pass through the entire reconstruction network. By $N_{\theta}$ we denote the total number of radial spokes which are acquired for each of the  $N_t$ cardiac phases. 
The first row of Figure \ref{computational_times} shows the average allocated GPU-memory as well the required time to perform one weights-update depending on the spatial image size. This is shown for $n_{\mathrm{CG}}=1$ and different numbers of radial spokes $N_{\theta}$. It can be seen that already for $n_\mathrm{cg}=1$, the required GPU memory amounts to approximately 512 GB - 1024 GB and performing one step of back-propagation is in the range of 4 seconds.
The second row of Figure \ref{computational_times} shows the same quantities for fixed $N_{\theta}=560$ and different $n_{\mathrm{CG}}$. Here, we also see that employing a relatively high number of CG iterations, say $n_{\mathrm{CG}}=12$, almost requires 2 GB of GPU memory and, more importantly, requires more than 30 seconds. Training the entire network in this configuration for 100 epochs would for example already require 5 days. By that, one can estimate that training the entire network from scratch in an end-to-end manner could easily amount to weeks or months.
Further, the required time does not vary much for different image sizes, meaning that  even for relatively small reconstruction problems, the application of $\Au$ and $\Au^\herm$ is inherently computationally demanding.\\
These computational aspects highlight the importance of the efficacy of the CNN-block in terms of having a small number of parameters to ensure fast convergence during training and at the same time a good performance in terms of undersampling artefacts-reduction.

\subsection{Efficacy of the Training Scheme}\label{subsec:training_behaviour}
\begin{figure}[t]
\centering
\begin{overpic}[height=7cm]{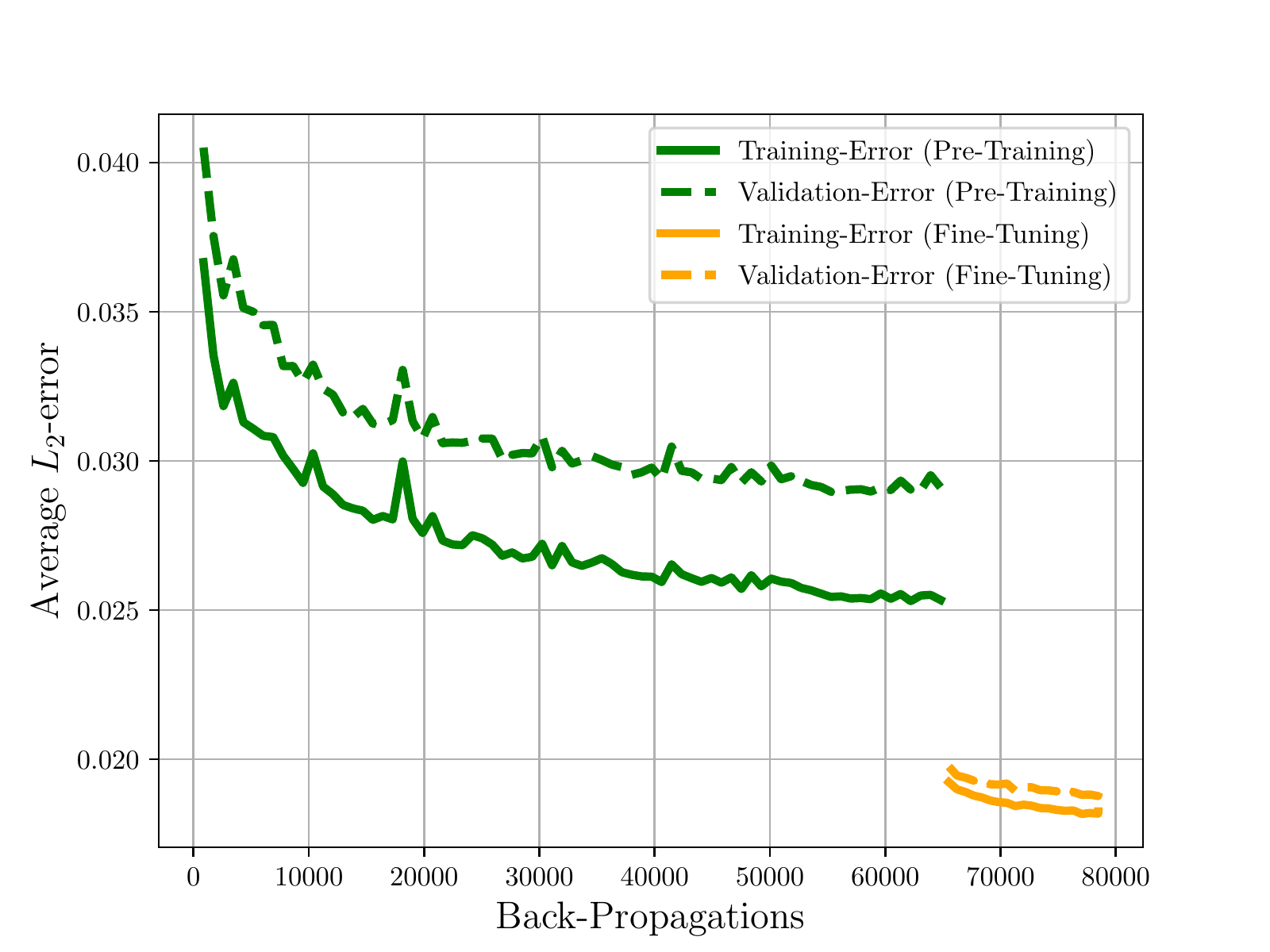}
\end{overpic}
\caption{Training behaviour of the proposed network architecture. 
The lines indicate the $L_2$-error on the training data (solid lines) as well as on the validation data (dashed lines) during pre-training of the CNN-block (green lines) and fine-tuning of the entire network (orange lines). As can be seen, in the fine-tuning stage, where the physical models $\Au$ and $\Au^\herm$ are included in the network architecture, the error is further reduced on both the training and the validation data. Further, the gap between training- and validation-errors becomes smaller. 
}\label{training_process_fig}\vspace{0.5cm}
\end{figure}

Here, we show the impact of including the forward  and the adjoint operators $\Au$ and $\Au^\herm$ in the network architecture during the learning process.  Figure \ref{training_process_fig} shows  the training- and validation-error of the  proposed CNN-module during the pre-training stage.
In the first pre-training stage, only one single CNN-block was trained to minimize the $L_2$-error between the output estimated by the CNN-block $u_{\Theta}$ and its corresponding label. In the fine-tuning stage, a CG-module with $n_{\mathrm{CG}}=8$ iterations was attached to the CNN-block. As can be seen, in the pre-training stage, after about $65\,000$ weight updates (which corresponds to approximately 450 epochs using a mini-batch size of one), training- and validation-error start to stagnate between 0.025 and 0.30, respectively. After pre-training, the parameter set $\Theta$ was stored and the fine-tuning of the entire  network was carried out by initializing the set of parameters $\Theta$ as the one obtained after pre-training. As can be seen from the orange curves, including the GC-module which employs the encoding operators in the CNN architecture allowed to further reduce the training- as well as the validation error and therefore to obtain a better more suitable parameter set $\Theta$.\\
Note that since the fine-tuning stage is much more computationally demanding than the pre-training stage, we only trained for 150 epochs and tested on the whole training and validation dataset only 12 times instead of 75 times as in the pre-training stage. The time for pre-training the CNN-module amounted to approximately 6 hours, while fine-tuning the entire network took approximately 5 days.
\begin{figure*}[t]
\begin{minipage}{\linewidth}
\begin{minipage}{\linewidth}
\begin{minipage}{0.49\linewidth}
\centering \quad After Pre-Training 
\end{minipage}
\begin{minipage}{0.49\linewidth}
\centering  After Fine-Tuning 
\end{minipage}
\end{minipage}
\begin{minipage}{\linewidth}
\begin{minipage}{0.015\linewidth}
\rotatebox{90}{\quad \quad \quad \quad \quad \quad \quad \quad  Final Reco  \quad \quad CNN-prior}
\end{minipage}
\begin{minipage}{0.47\linewidth}
\resizebox{\linewidth}{!}{
\includegraphics[height=3.cm]{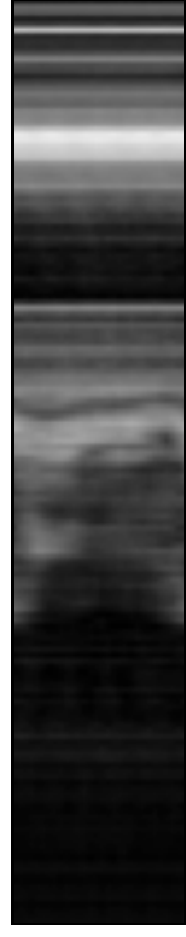}\hspace{-0.2cm}
\begin{overpic}[height=3.cm,tics=10]{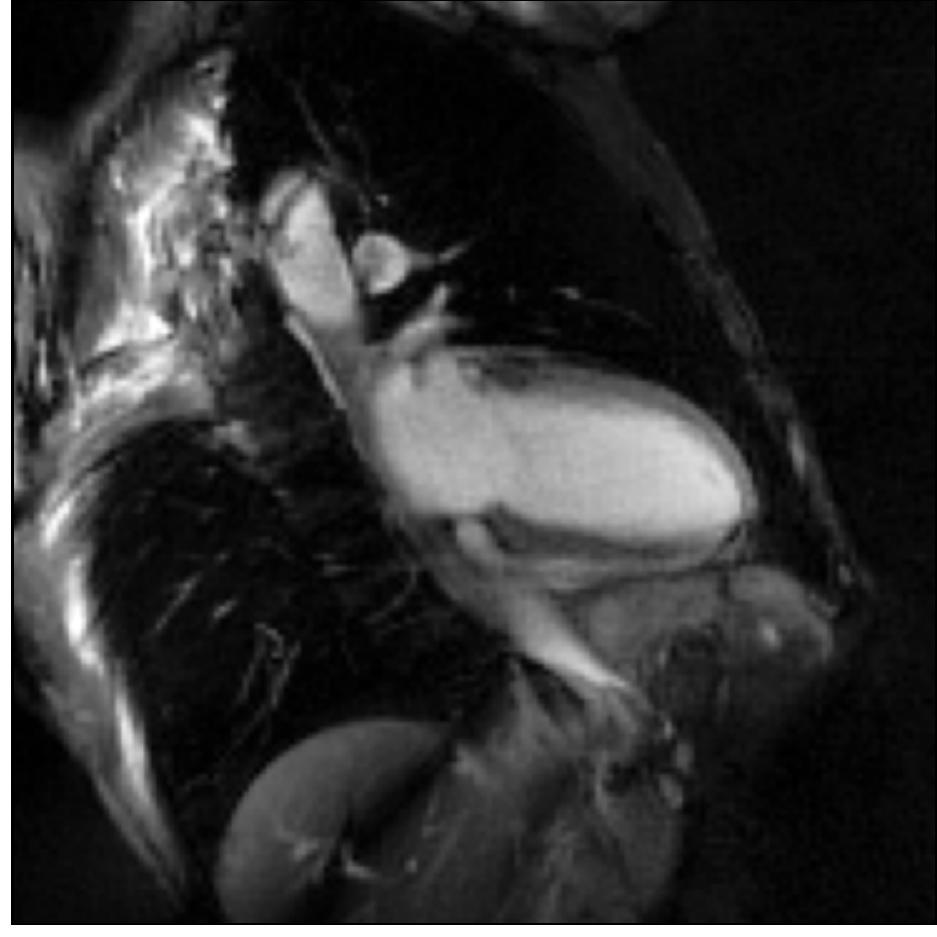}
\put (72,82) {\normalsize\textcolor{white}{(A)}}
\end{overpic} \hspace{-0.2cm}

\includegraphics[height=3.cm]{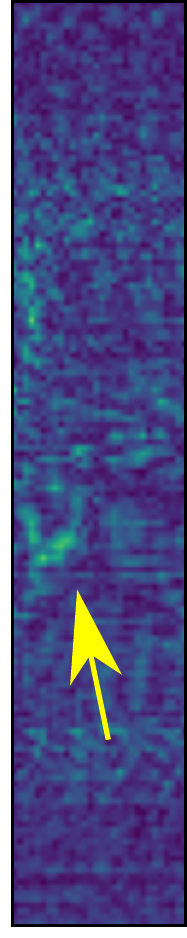}\hspace{-0.2cm}
\begin{overpic}[height=3.cm,tics=10]{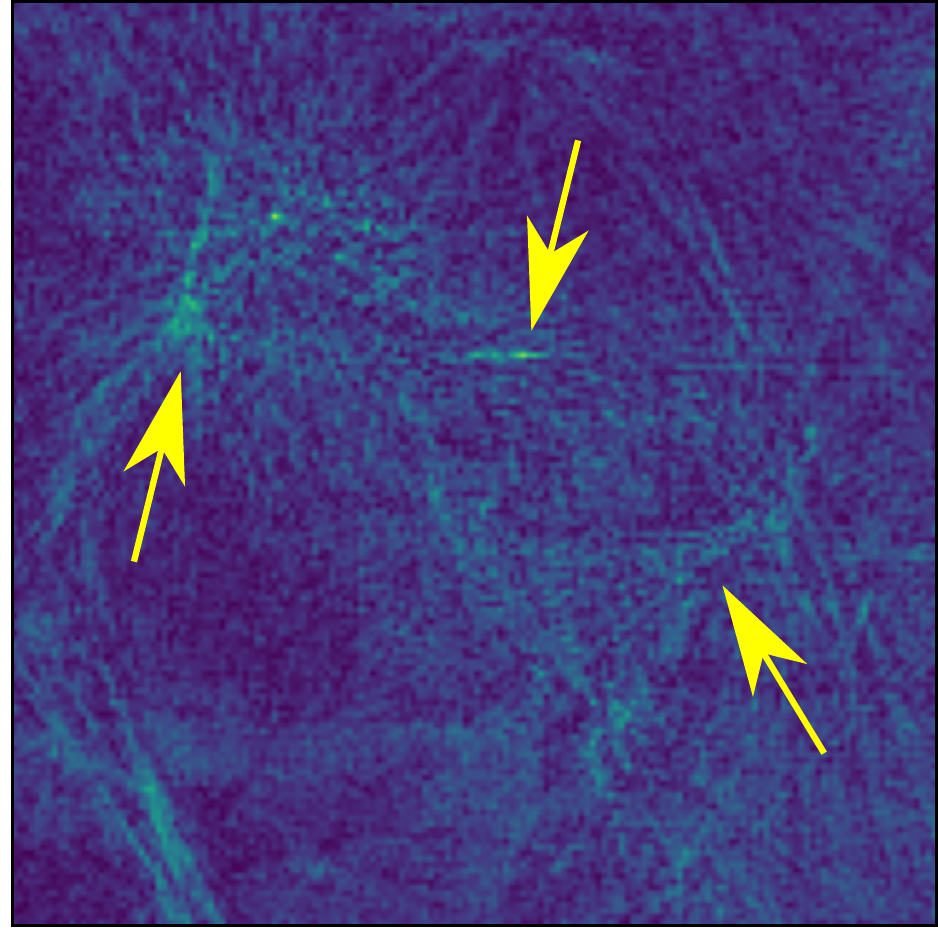}

\end{overpic} \hspace{-0.2cm}
}
\resizebox{\linewidth}{!}{
\includegraphics[height=3.cm]{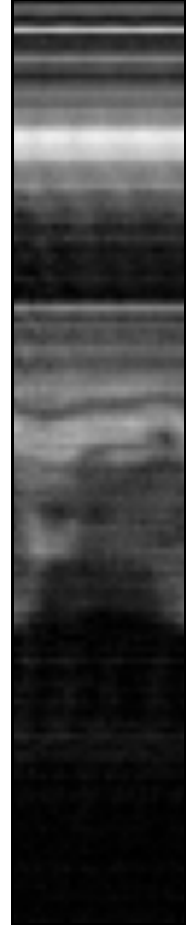}\hspace{-0.2cm}
\begin{overpic}[height=3.cm,tics=10]{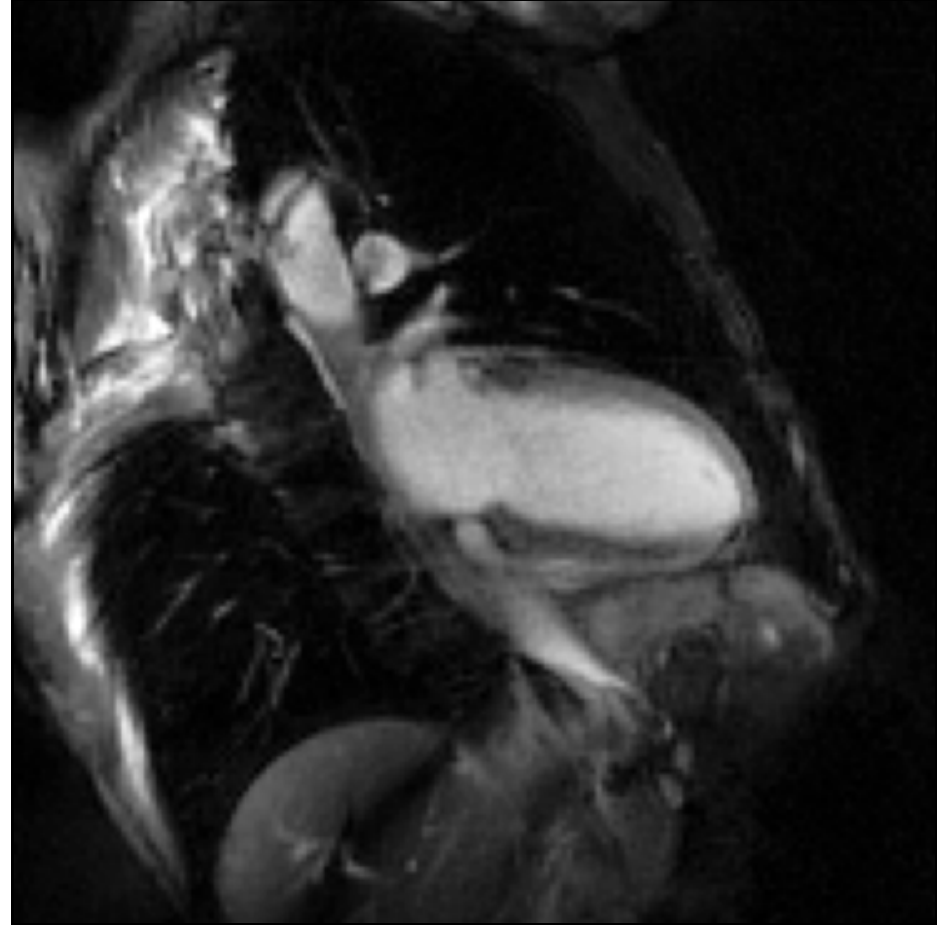}
\put (72,82) {\normalsize\textcolor{white}{(B)}}
\end{overpic} \hspace{-0.2cm}
\includegraphics[height=3.cm]{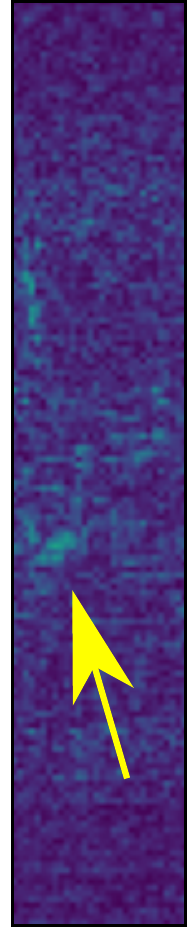}\hspace{-0.2cm}
\begin{overpic}[height=3.cm,tics=10]{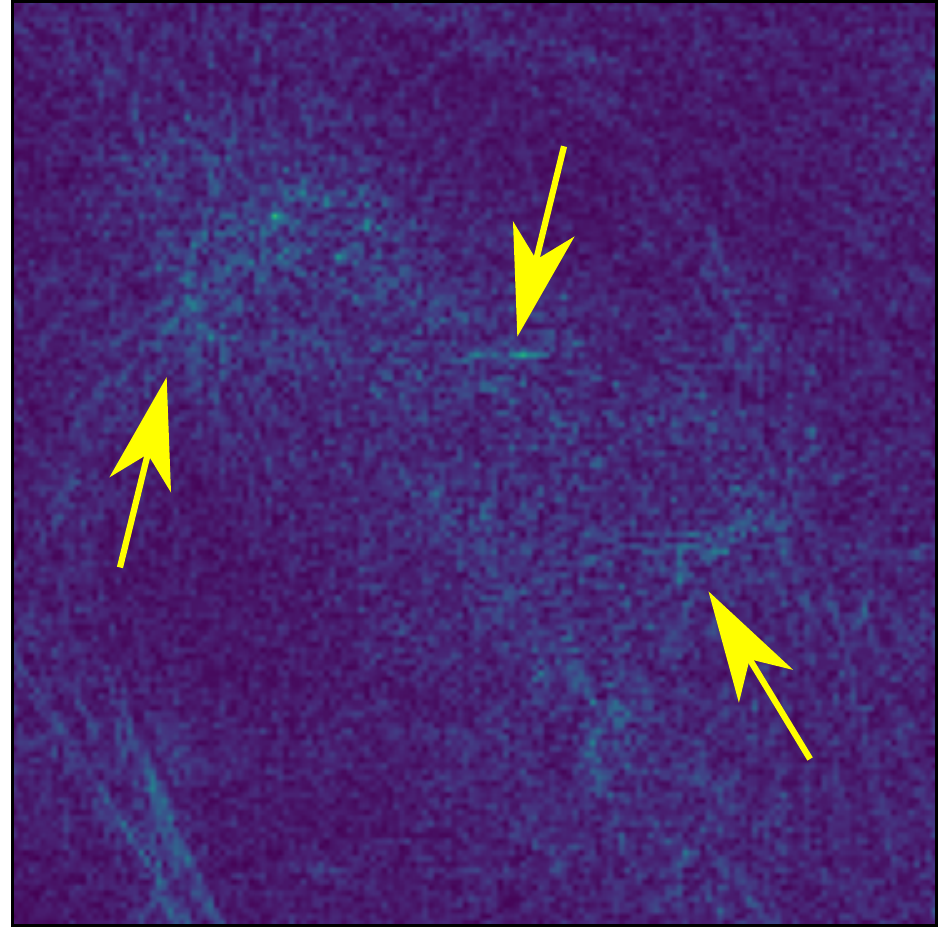}

\end{overpic} \hspace{-0.2cm}
}\vspace{0.2cm}
\resizebox{\linewidth}{!}{
\includegraphics[height=3.cm]{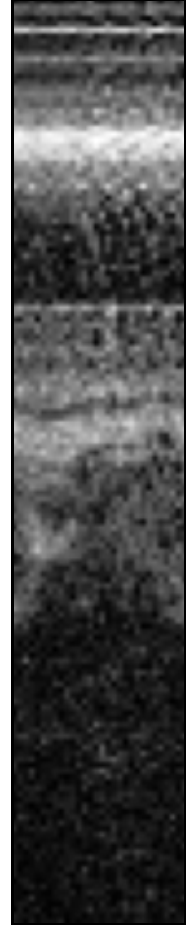}\hspace{-0.2cm}
\begin{overpic}[height=3.cm,tics=10]{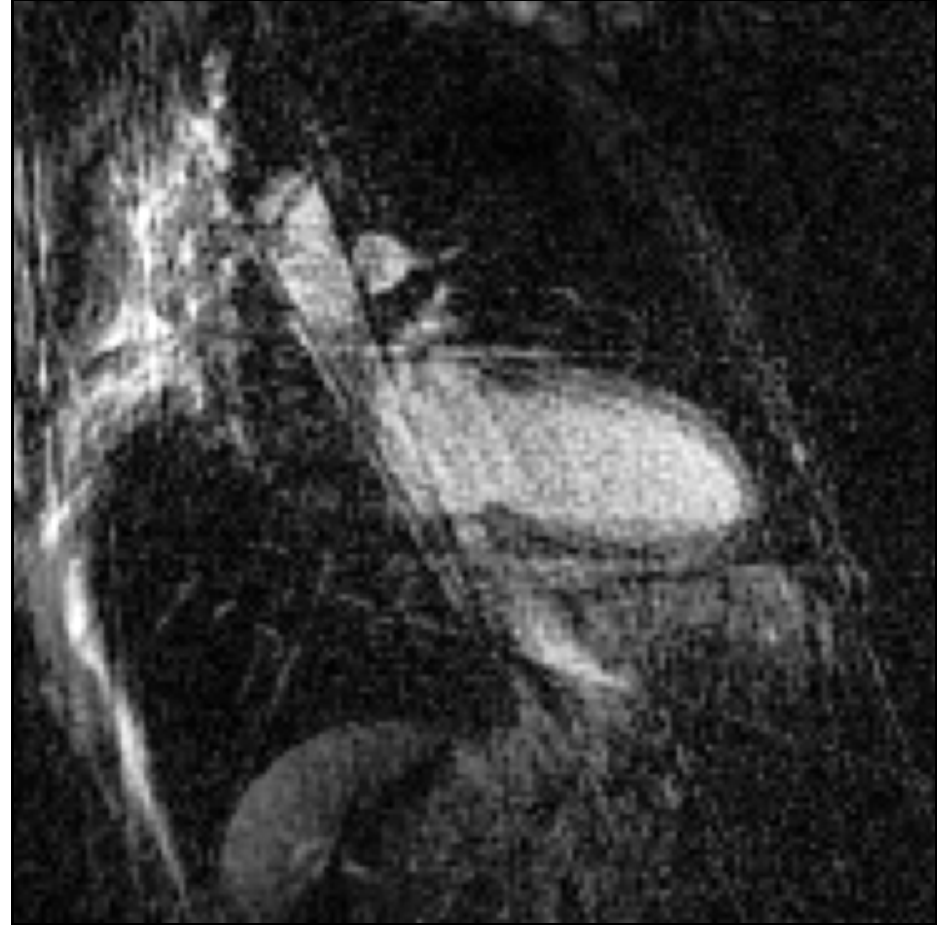}
\put (72,82) {\normalsize\textcolor{white}{(E)}}
\end{overpic} \hspace{-0.2cm}

\includegraphics[height=3.cm]{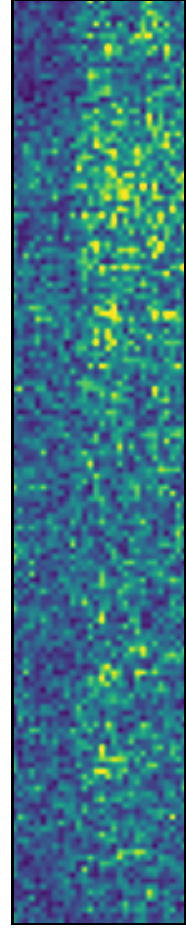}\hspace{-0.2cm}
\begin{overpic}[height=3.cm,tics=10]{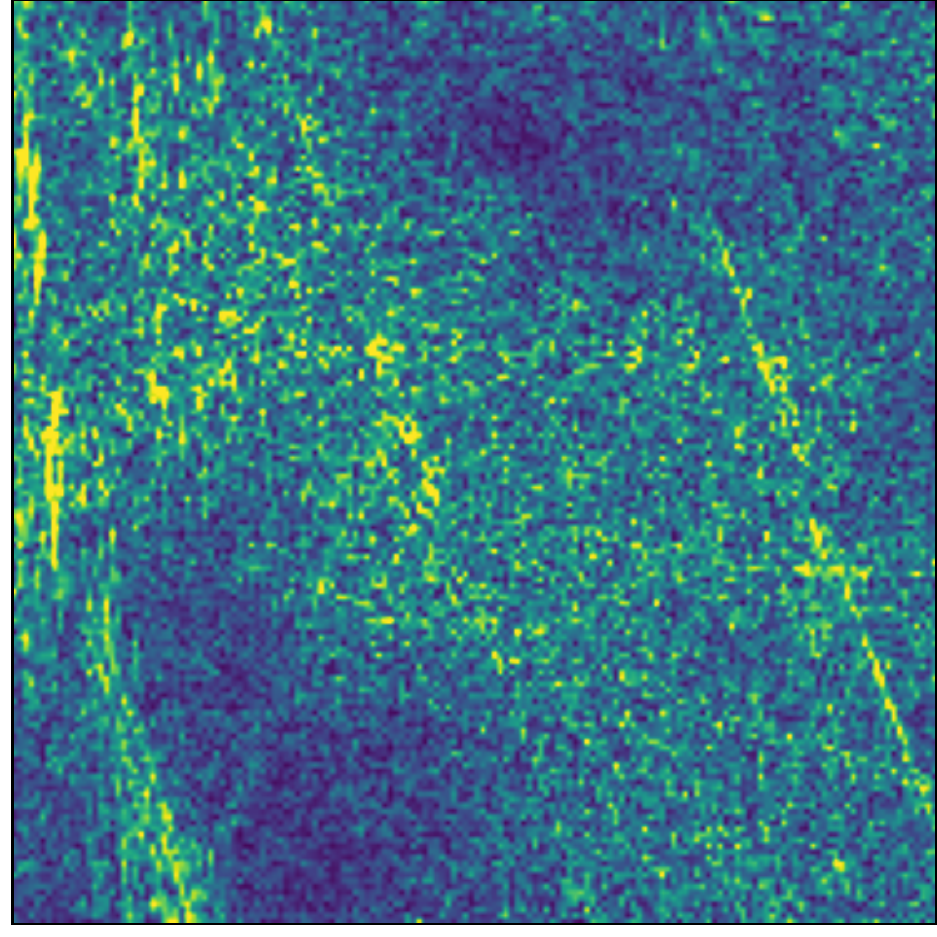}

\end{overpic} \hspace{-0.2cm}
}
\end{minipage}
\begin{minipage}{0.47\linewidth}
\resizebox{\linewidth}{!}{
\includegraphics[height=3.cm]{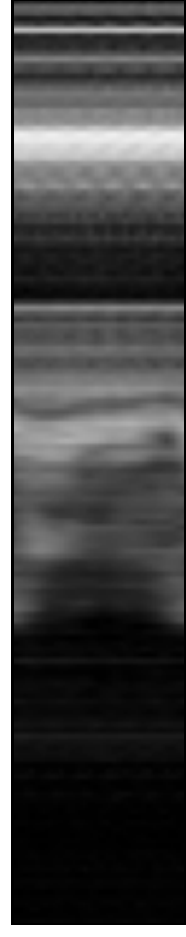}\hspace{-0.2cm}
\begin{overpic}[height=3.cm,tics=10]{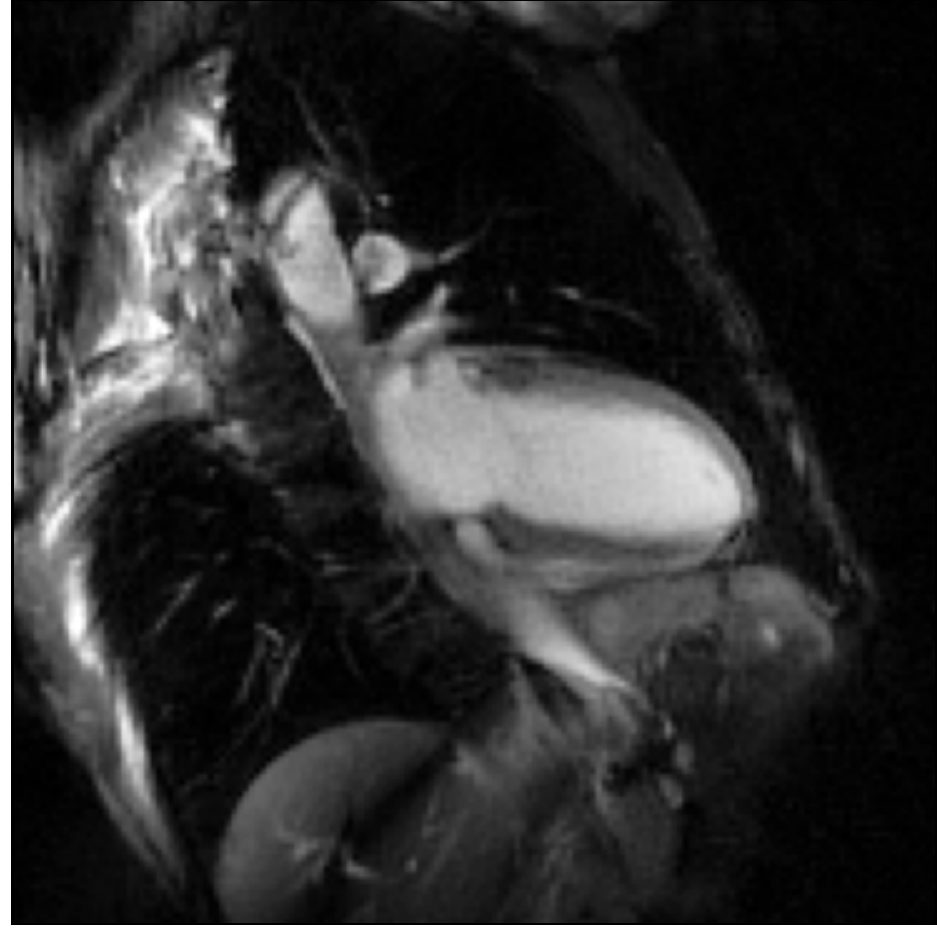}
\put (72,82) {\normalsize\textcolor{white}{(C)}}
\end{overpic} \hspace{-0.2cm}

\includegraphics[height=3.cm]{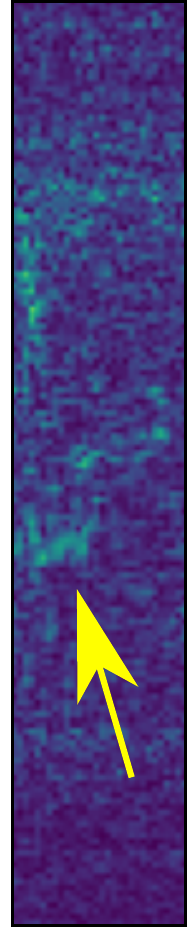}\hspace{-0.2cm}
\begin{overpic}[height=3.cm,tics=10]{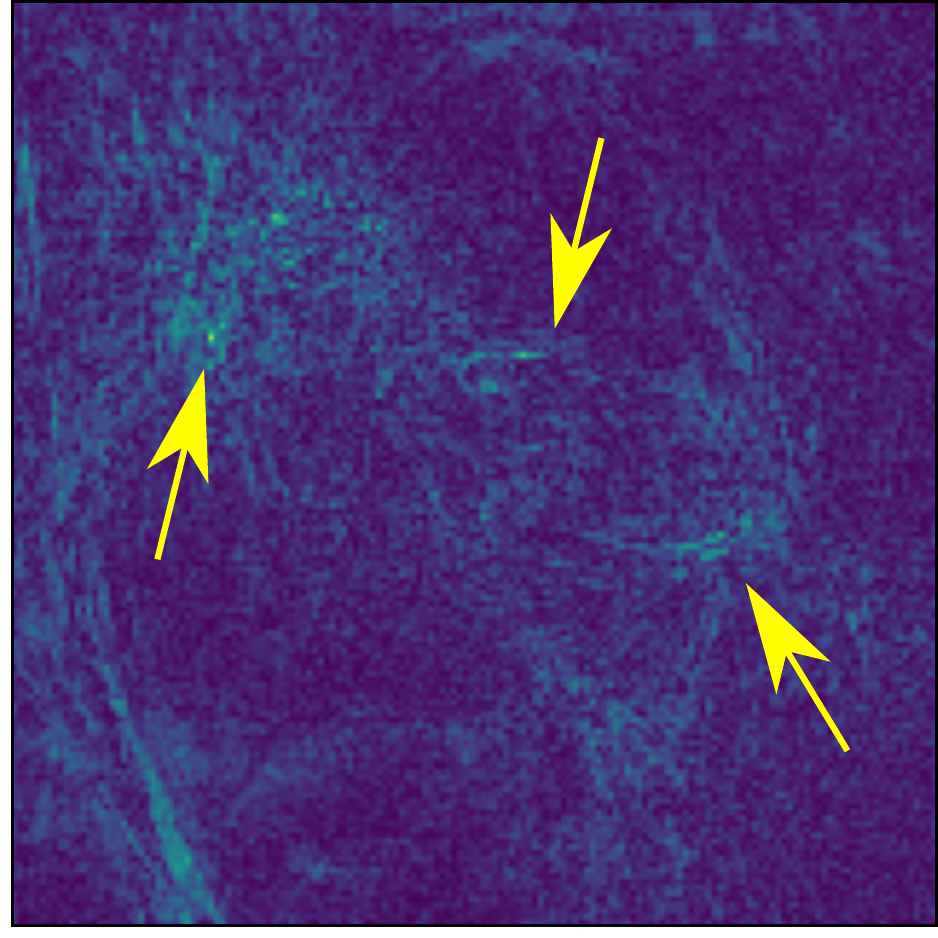}

\end{overpic} \hspace{-0.2cm}
}
\resizebox{\linewidth}{!}{
\includegraphics[height=3.cm]{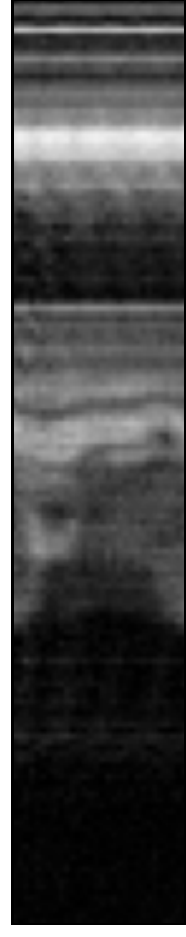}\hspace{-0.2cm}
\begin{overpic}[height=3.cm,tics=10]{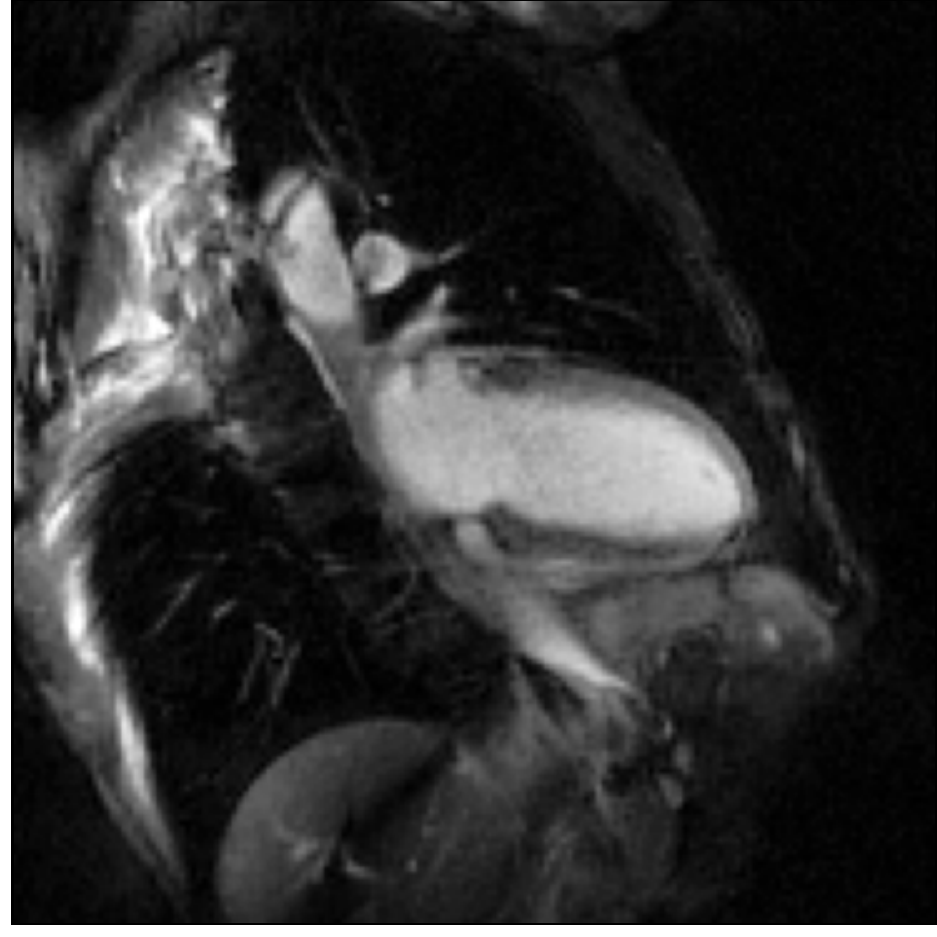}
\put (72,82) {\normalsize\textcolor{white}{(D)}}
\end{overpic} \hspace{-0.2cm}

\includegraphics[height=3.cm]{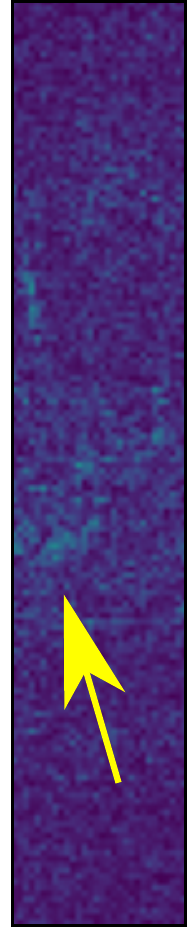}\hspace{-0.2cm}
\begin{overpic}[height=3.cm,tics=10]{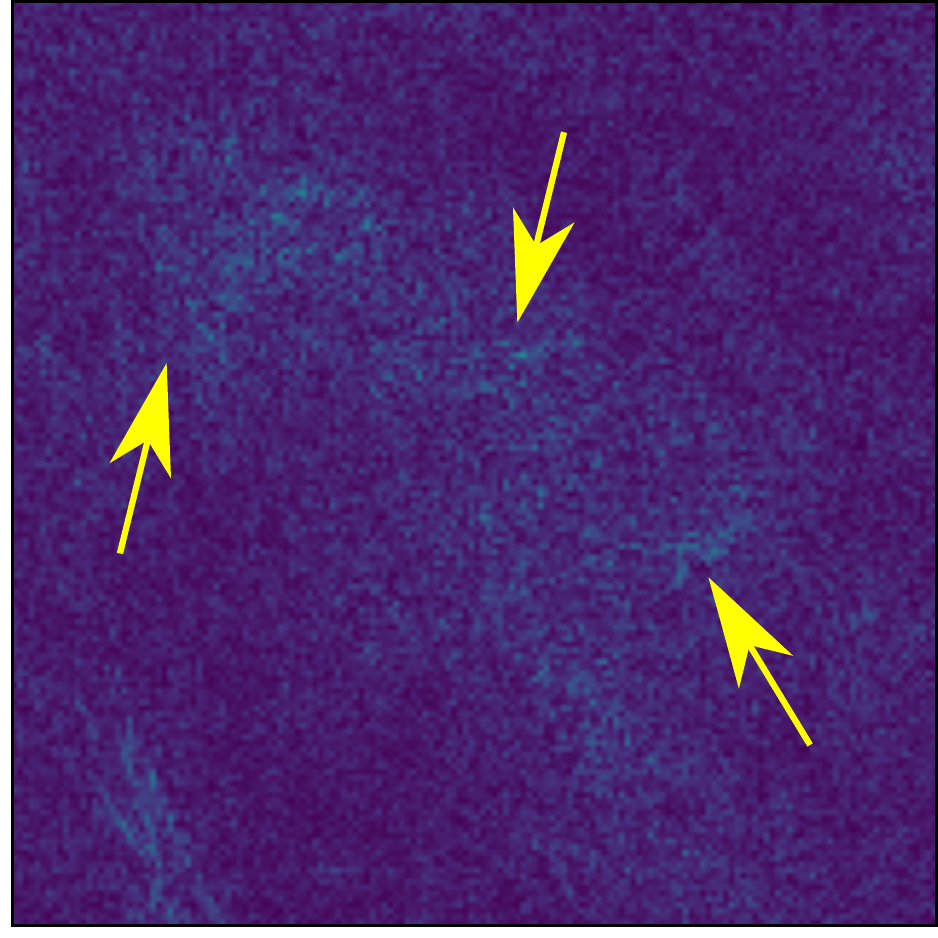}

\end{overpic} \hspace{-0.2cm}
}\vspace{0.2cm}
\resizebox{\linewidth}{!}{
\includegraphics[height=3.cm]{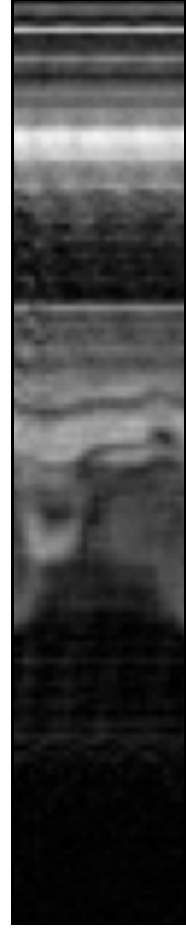}\hspace{-0.2cm}
\begin{overpic}[height=3.cm,tics=10]{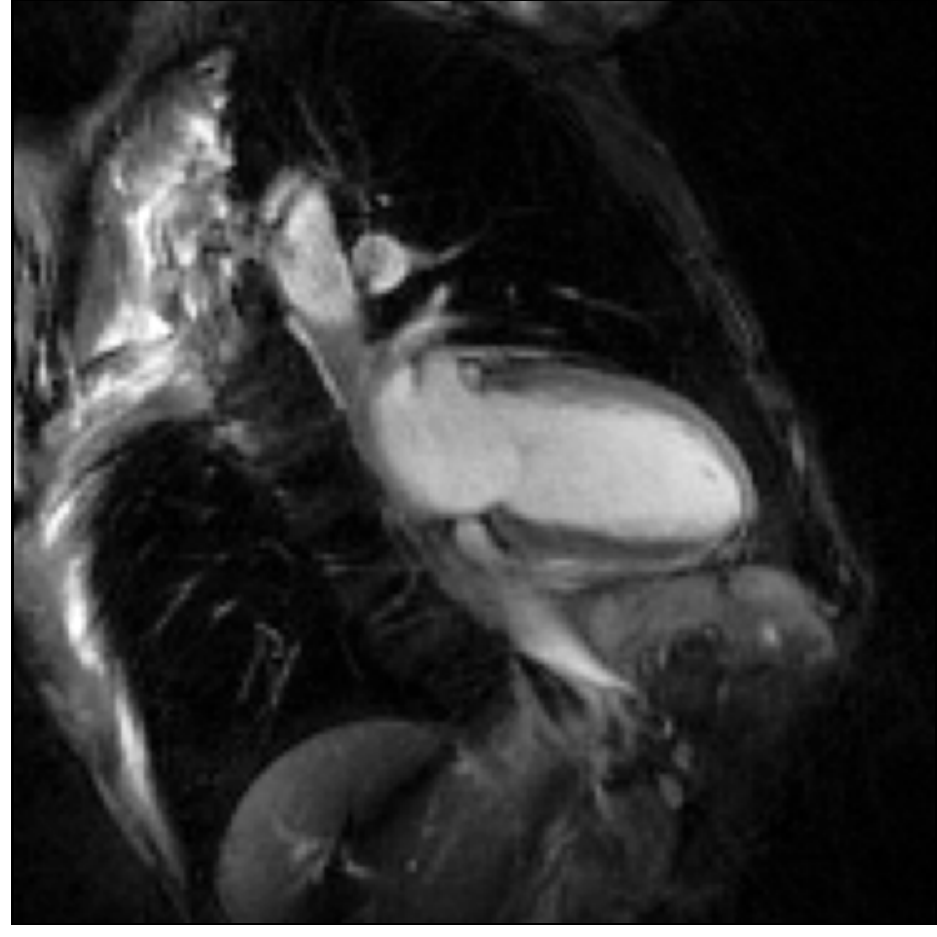}
\put (72,82) {\normalsize\textcolor{white}{(F)}}
\end{overpic} \hspace{-0.2cm}

\includegraphics[height=3cm]{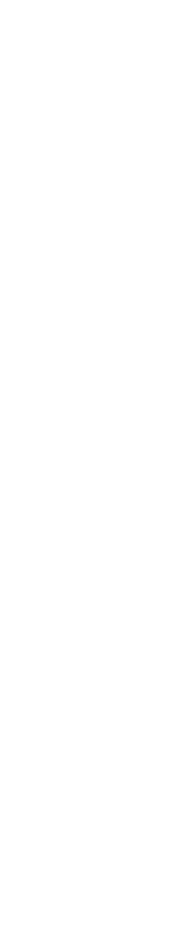}\hspace{-0.2cm}
\begin{overpic}[height=3cm,tics=10]{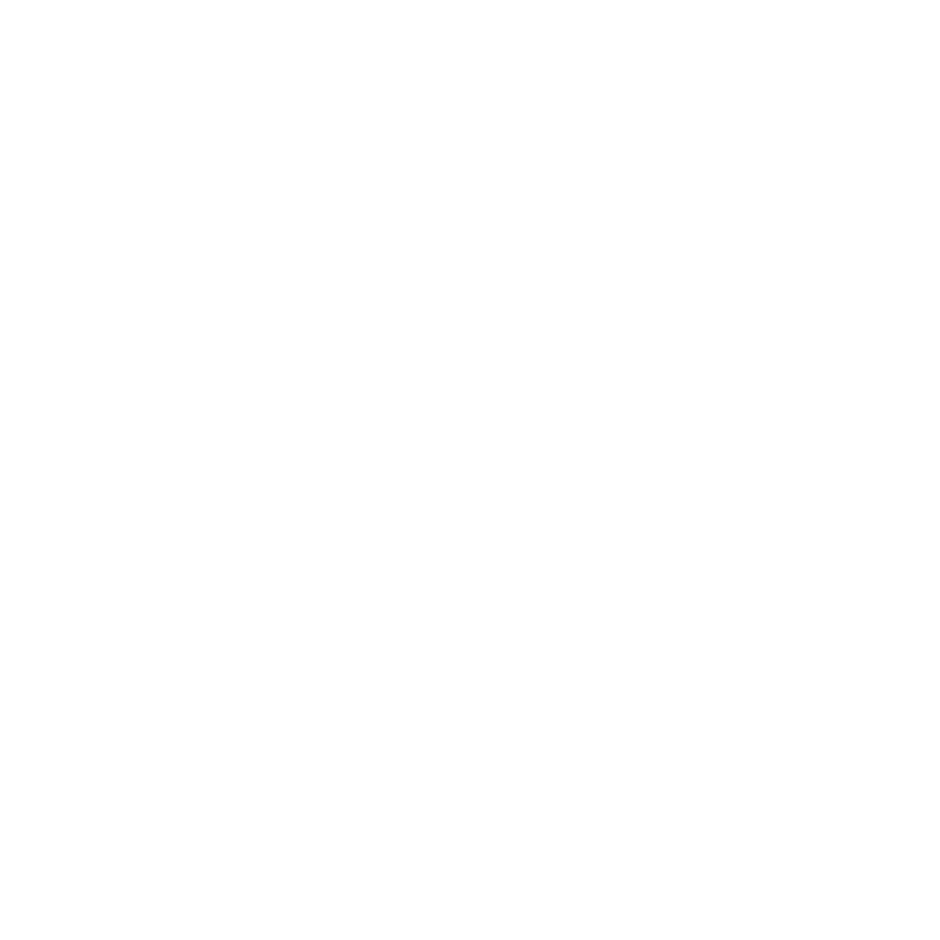}

\end{overpic} \hspace{-0.2cm}
}

\end{minipage}
\end{minipage}
\end{minipage}
\caption{Intermediate results and point-wise error images of our reconstruction method after pre-training and after fine-tuning our proposed method with $M=1$ and $n_{\mathrm{CG}}=8$. The output of the CNN-block after pre-training (A), the final reconstruction (B) after pre-training where (A) was the output of the CNN-block, the output of the CNN-block after fine-tuning the entire cascade in an end-to-end manner (C) and the final reconstruction after fine-tuning (D) where (C) was the output of the CNN-block, the initial NUFFT-reconstrution $\Xu$ obtained from $N_{\theta}=560$ radial spokes (E) and  the corresponding ground-truth image obtained by $kt$-SENSE with $N_{\theta}=3400$ radial spokes (F). We see that after having fine-tuned the entire network, the point-wise error of the final reconstruction is the smallest. Further, the cardiac motion is much better preserved as is pointed out by the yellow arrows.}\label{different_stages_results_fig}
\end{figure*}
\begin{table*}[t]
	\centering
	\renewcommand{\arraystretch}{1.3}
	\footnotesize{
		\caption{Intermediate and final reconstruction results after pre-training of the CNN-block and after fine-tuning of the entire network architecture with $M=1$ and $n_{\mathrm{CG}}=8$. The results are shown for   $N_{\theta}=560$ and $N_{\theta}=1130$ radial spokes.} \label{table_intermediate_results}
		\vspace{0.2cm}
		\centering
		\begin{tabular}{ |l|
				c
				c|
				c
				c|
			}
	\bottomrule 
	& \multicolumn{2}{c|}{\textbf{After Pre-Training}} & \multicolumn{2}{c|}{\textbf{After Fine-Tuning}}\\
	\midrule
	& \textbf{CNN-Prior} & \textbf{Final Reconstruction}  &  \textbf{CNN-Prior} & \textbf{Final Reconstruction}  \\
			\midrule
& \multicolumn{4}{c|}{\textbf{Number of Radial Spokes:} $N_{\theta} = 560$}\\
\midrule	 
\textbf{PSNR} 		&42.4541 &45.4826 &44.484 & \bf{46.0036}  \\
\textbf{NRMSE} 		& 0.1252 & 0.0874 & 0.0963 & \bf{0.0810 } \\
\textbf{SSIM} 		& 0.9576 & 0.9796 & 0.9833 & \bf{0.9872}  \\
\textbf{MS-SSIM} 	& 0.9916 & 0.9966 & 0.9965 & \bf{0.9977}  \\
\textbf{UQI} 		& 0.8027 & 0.8776 & 0.9064 & \bf{0.9275}  \\
\textbf{VIQP}		& 0.9431 & 0.9506 & 0.9429 & \bf{0.9434}  \\
\textbf{HPSI} 		& 0.9889 & 0.9941 & 0.9901 & \bf{0.9943}  \\
\midrule

& \multicolumn{4}{c|}{\textbf{Number of Radial Spokes:} $N_{\theta} = 1130$}\\
\midrule	
\textbf{PSNR} 		&43.9893 &47.0087 &46.2383 & \bf{47.7545}  \\
\textbf{NRMSE} 		& 0.1044 & 0.0735 & 0.0808 & \bf{0.0673}  \\
\textbf{SSIM} 		& 0.9698 & 0.9854 & 0.9873 & \bf{0.9903}  \\
\textbf{MS-SSIM} 	& 0.9945 & 0.9977 & 0.9974 & \bf{0.9984}  \\
\textbf{UQI} 		& 0.8324 & 0.8998 & 0.9209 & \bf{0.9408}  \\
\textbf{VIQP}		& 0.9574 & 0.9642 & 0.9604 &\bf{ 0.9626}  \\
\textbf{HPSI} 		& 0.9916 & 0.9961 & 0.9930 & \bf{0.9962}  \\

			\bottomrule 
		\end{tabular}
		
	}
	
	\vspace{0.3cm}
\end{table*}

Figure \ref{different_stages_results_fig} shows the effect of the proposed training strategy. Figure \ref{different_stages_results_fig}(E) shows the initial NUFFT-reconstruction which is directly obtained from the undersampled $k$-space data. After having pre-trained the single block $u_{\Theta}$, the image in Figure \ref{different_stages_results_fig}(A) is obtained by passing the input to the CNN-block. Further, proceeding in the reconstruction network with the CG-bock yields the image shown in Figure \ref{different_stages_results_fig}(B) for which the point-wise error is lower than for (A). 
Figure \ref{different_stages_results_fig}(C) shows the output of the CNN-block after having fine-tuned the entire network architecture, i.e.\ a CNN-block with attached CG-block,  by end-to-end training. By comparing (C) to (A), we see that the quality of the output of the CNN-block has clearly increased as the point-wise error has been further reduced. Further, performing proceeding with the CG-module in the network further reduces the point-wise errors, see (D). Again, by comparing the final reconstructions (D) and (B) we see that the quality of the reconstruction has further increased as the point-wise error has decreased. Note that by taking in consideration the stagnation of the training- and validation error shown in Figure \ref{training_process_fig} we can indeed attribute the increase of performance of the reconstruction method to the fact that we included the forward and adjoint operators in the training process and most probably not to the additionally performed weight updates. Further, we have experimentally confirmed the efficacy of the proposed training procedure.\\
Table \ref{table_intermediate_results} lists the quantitative measures obtained on the test set for the intermediate output of the CNN-block as well as for the final reconstruction before and after fine-tuning the entire network for two different acceleration factors. The table well reflects the visual results in the sense that the intermediate output of the CNN as well as the final reconstruction after the fine-tuning stage surpass their corresponding image estimates after the sole pre-traning in terms of all reported measures.

\subsection{Variation of the Hyper-Parameters $M$ and $n_{\mathrm{CG}}$}\label{subsec:MnCG_variation}

As already mentioned, at test time, one does not necessarily need to stick to the configuration of the network in terms of length $M$ and number of CG-iterations $n_{\mathrm{CG}}$ which were used for fine-tuning the network. In particular, for the fine-tuning stage, the choice of $M$ and $n_{\mathrm{CG}}$ is mainly driven by factors as training times and hardware constraints which do not play a role at test time.\\
Thus, we performed a parameter study where at test time, we varied the length of the network $M$ as well as the number of CG-iterations $n_{\mathrm{CG}}$. We repeated these experiment for two different configurations of our proposed network. We fine-tuned one network with $M=3$ and $n_{\mathrm{CG}}=2$ and another one with $M=1$ and $n_{\mathrm{CG}}=8$. Due to hardware constraints, the networks also differ in terms of number of trainable parameters of the CNN-block.\\
At test time, we fixed $n_{\mathrm{CG}}=4$ and varied $M$ from $M=2,4,\ldots,12$. With this configuration, the noise and the artefacts are gradually reduced and data-consistency of the solution is increased several times during the whole reconstruction process. In the supplementary material,  one can see that increasing $M$ consistently further improved the results for both networks. Further, the comparison of the two networks shows that the network containing more trainable parameters (i.e.\ the one which was fine-tuned with $M=1$ and $n_{\mathrm{CG}}=8$) surpasses the other one with respect to all measures which motivates the choice of the final reconstruction network used for comparison with other methods in the next Subsection.

\subsection{Reconstruction Results}

\begin{figure}
\begin{minipage}{0.49\linewidth}
\resizebox{\linewidth}{!}{
\includegraphics[height=2.9cm]{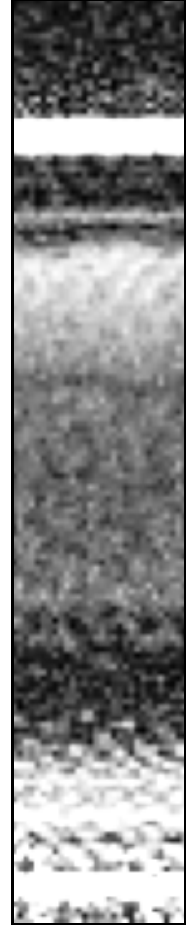}\hspace{-0.2cm}
\includegraphics[height=2.9cm]{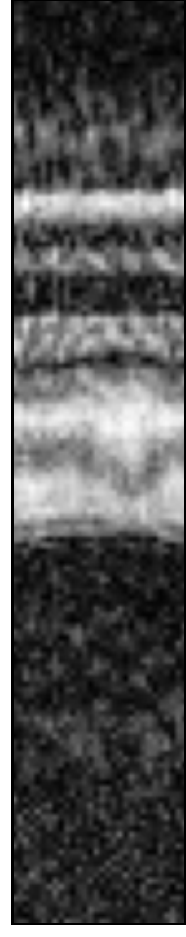}\hspace{-0.2cm}
\begin{overpic}[height=2.9cm,tics=10]{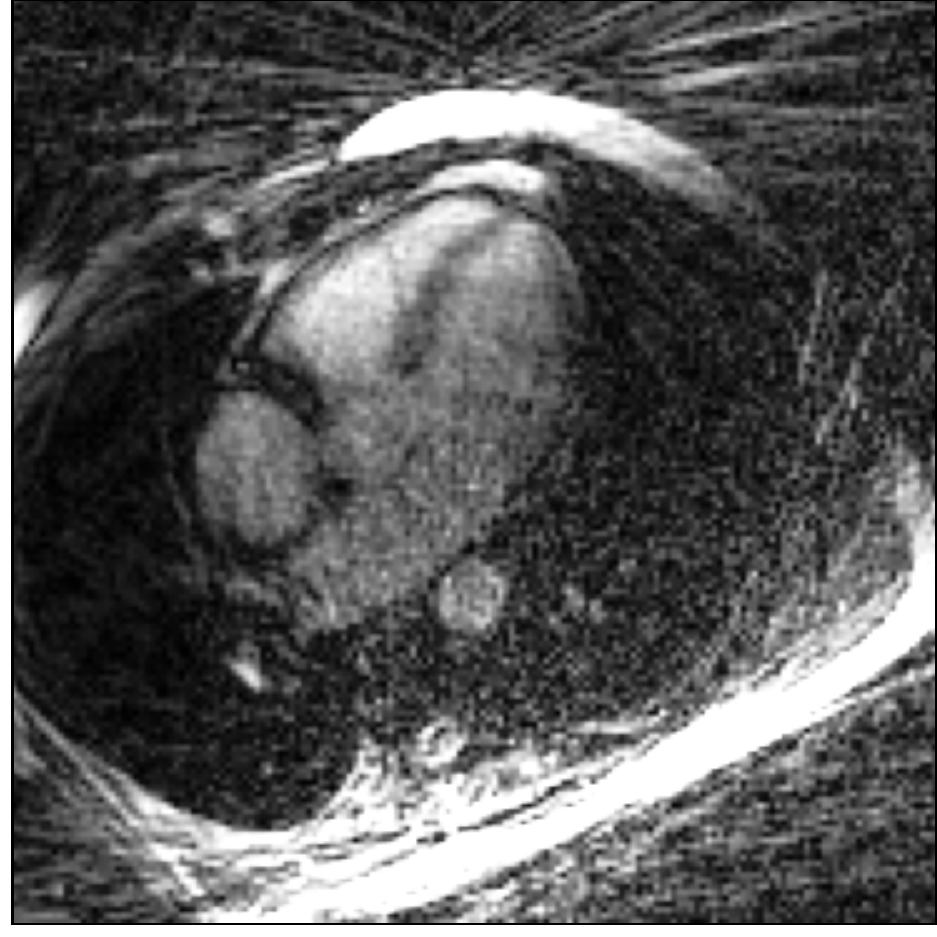}
\end{overpic} \hspace{-0.2cm}
\includegraphics[height=2.9cm]{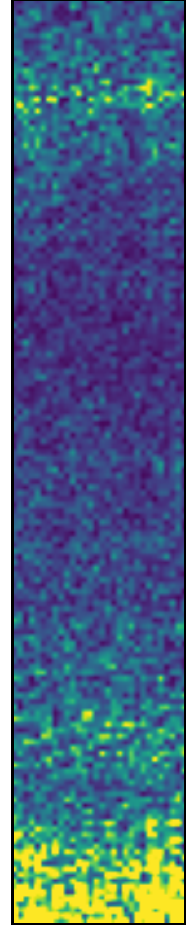}\hspace{-0.2cm}
\includegraphics[height=2.9cm]{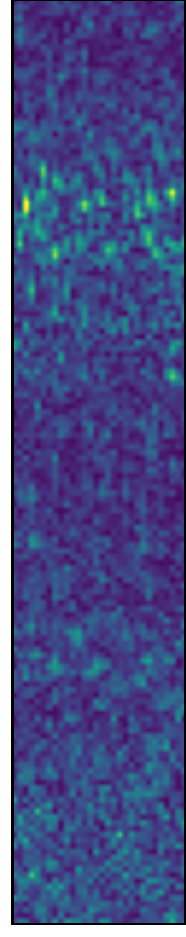}\hspace{-0.2cm}
\begin{overpic}[height=2.9cm,tics=10]{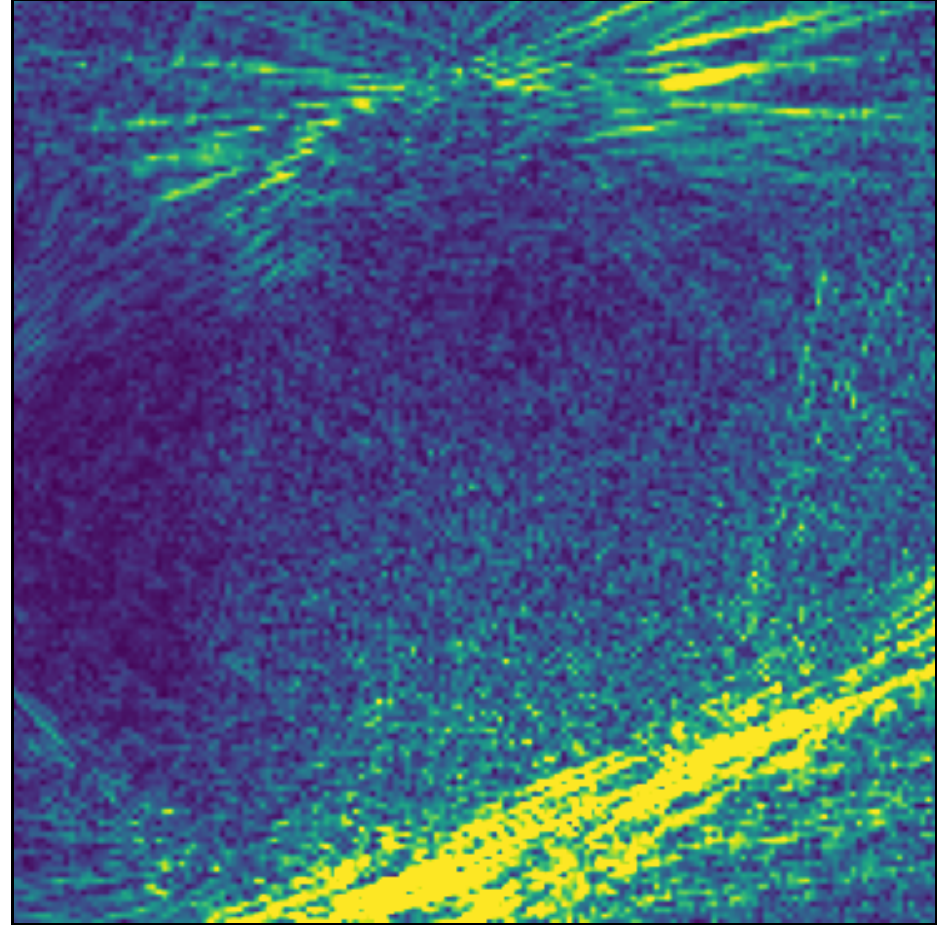}
 \put (59,87) {\tiny\textcolor{white}{\textbf{NUFFT}}}
\end{overpic} \hspace{-0.2cm}
}
\resizebox{\linewidth}{!}{
\includegraphics[height=2.9cm]{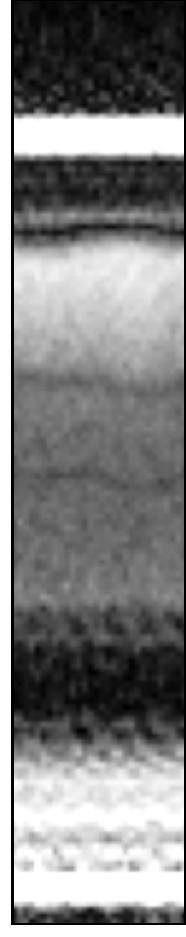}\hspace{-0.2cm}
\includegraphics[height=2.9cm]{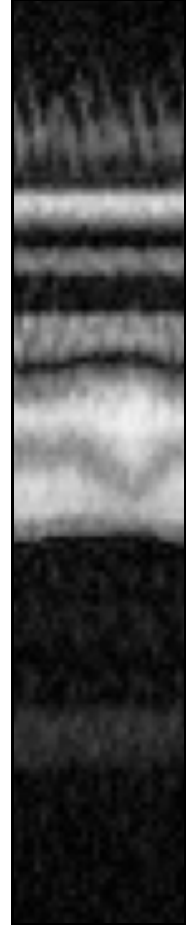}\hspace{-0.2cm}
\begin{overpic}[height=2.9cm,tics=10]{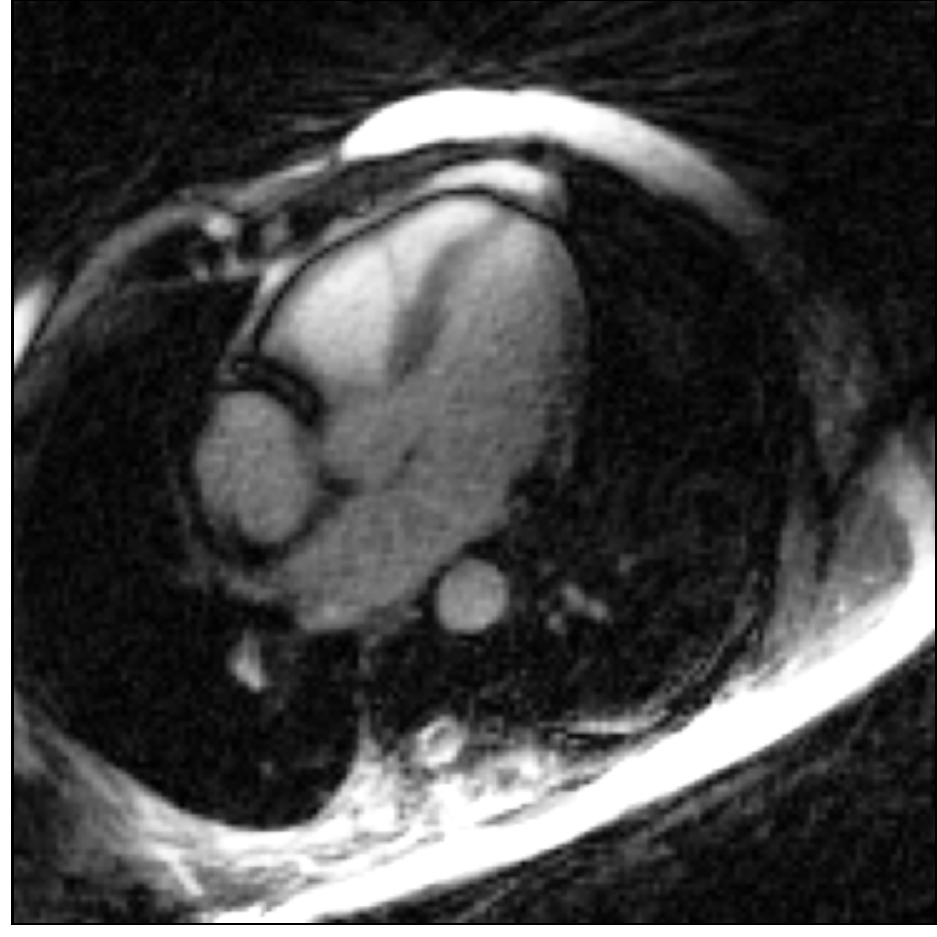}
\end{overpic} \hspace{-0.2cm}
\includegraphics[height=2.9cm]{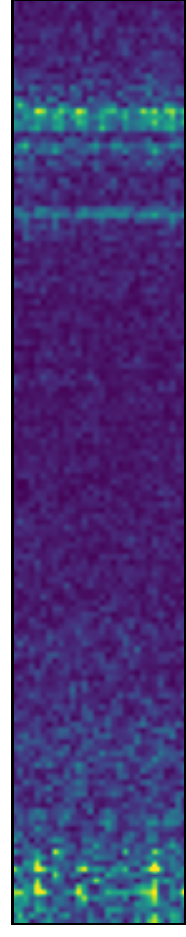}\hspace{-0.2cm}
\includegraphics[height=2.9cm]{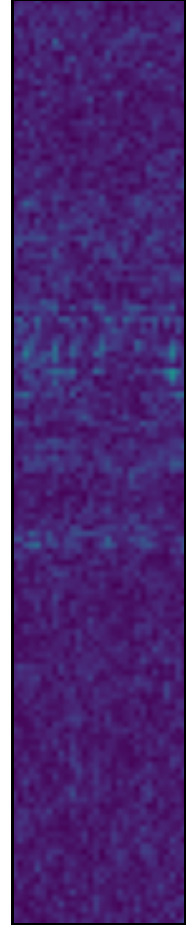}\hspace{-0.2cm}
\begin{overpic}[height=2.9cm,tics=10]{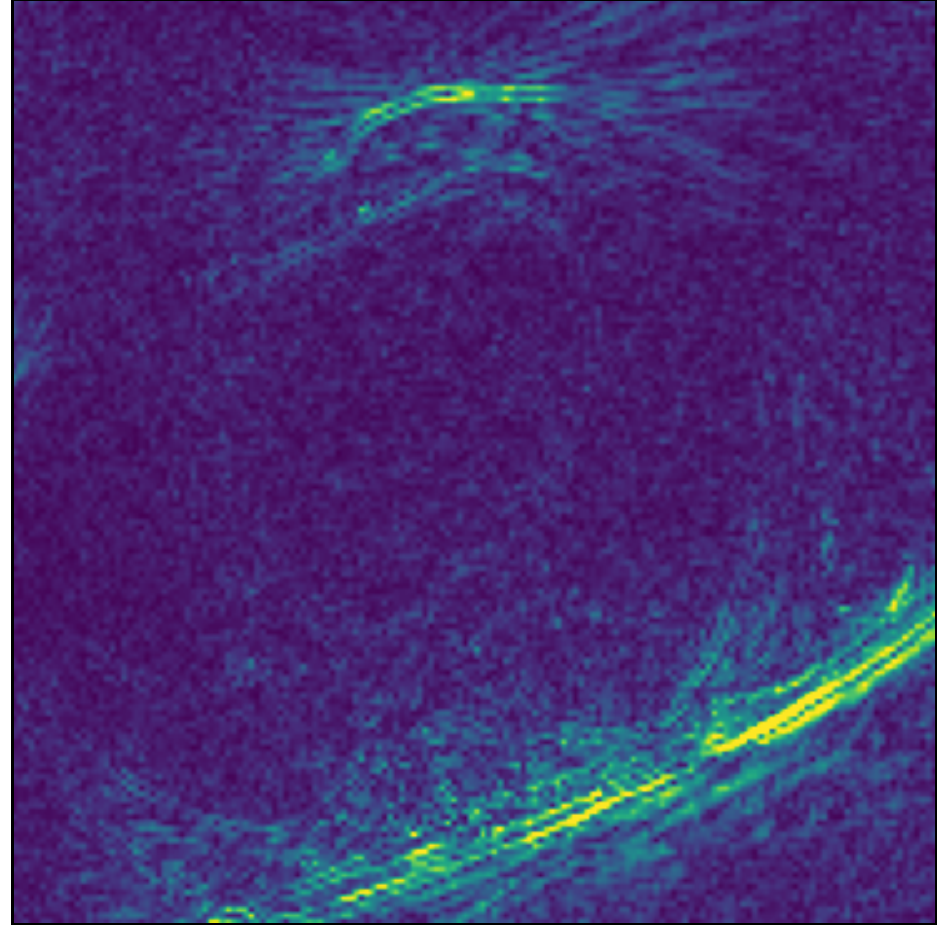}
 \put (46,87) {\tiny\textcolor{white}{\textbf{It. SENSE}}}
\end{overpic} \hspace{-0.2cm}
}
\resizebox{\linewidth}{!}{
\includegraphics[height=2.9cm]{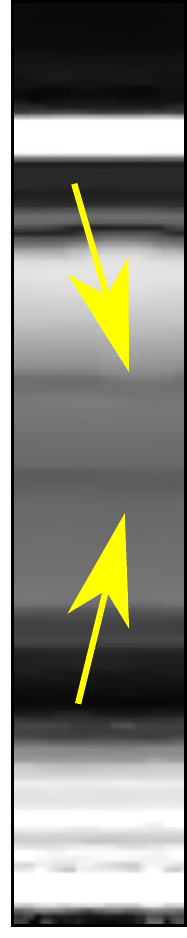}\hspace{-0.2cm}
\includegraphics[height=2.9cm]{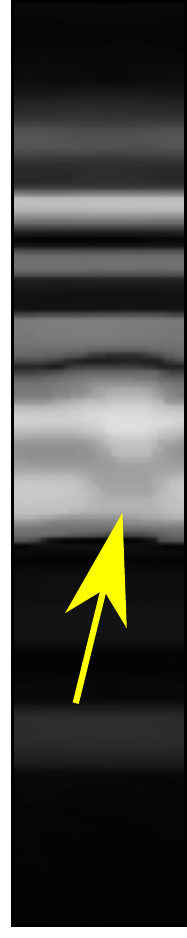}\hspace{-0.2cm}
\begin{overpic}[height=2.9cm,tics=10]{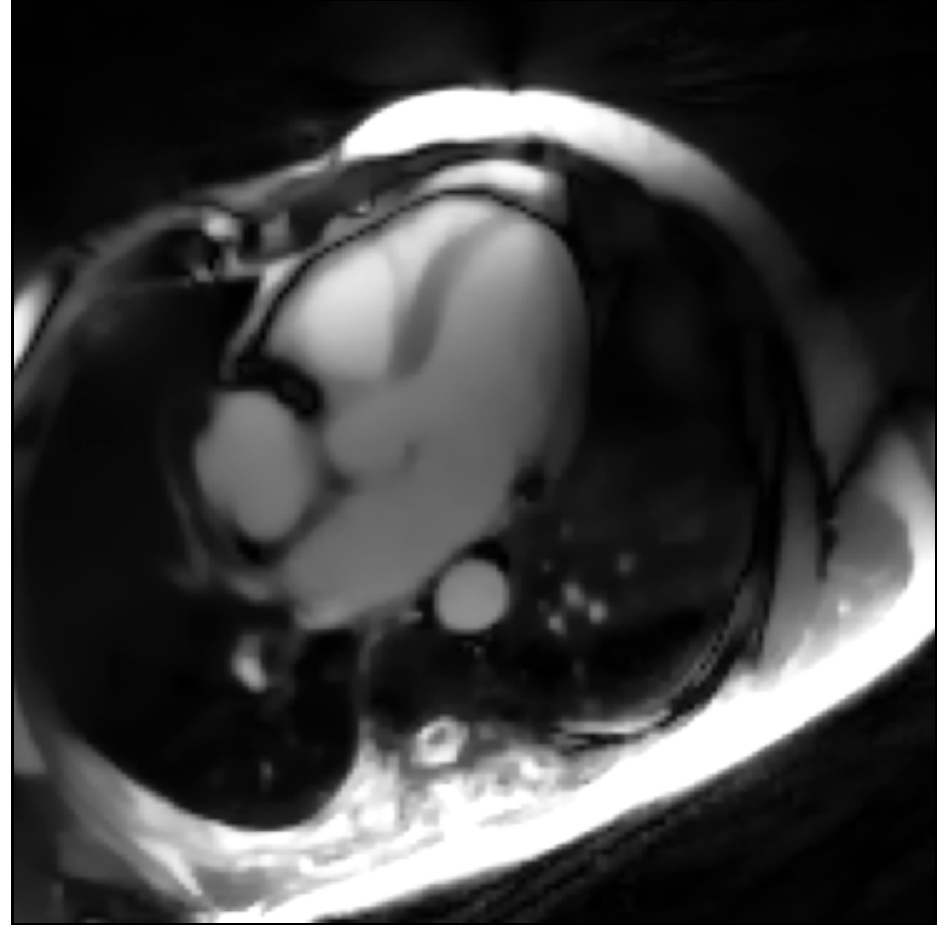}
\end{overpic} \hspace{-0.2cm}
\includegraphics[height=2.9cm]{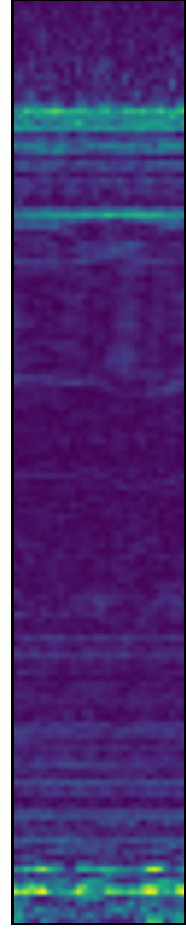}\hspace{-0.2cm}
\includegraphics[height=2.9cm]{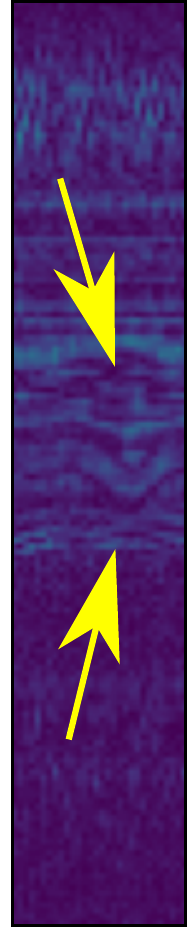}\hspace{-0.2cm}
\begin{overpic}[height=2.9cm,tics=10]{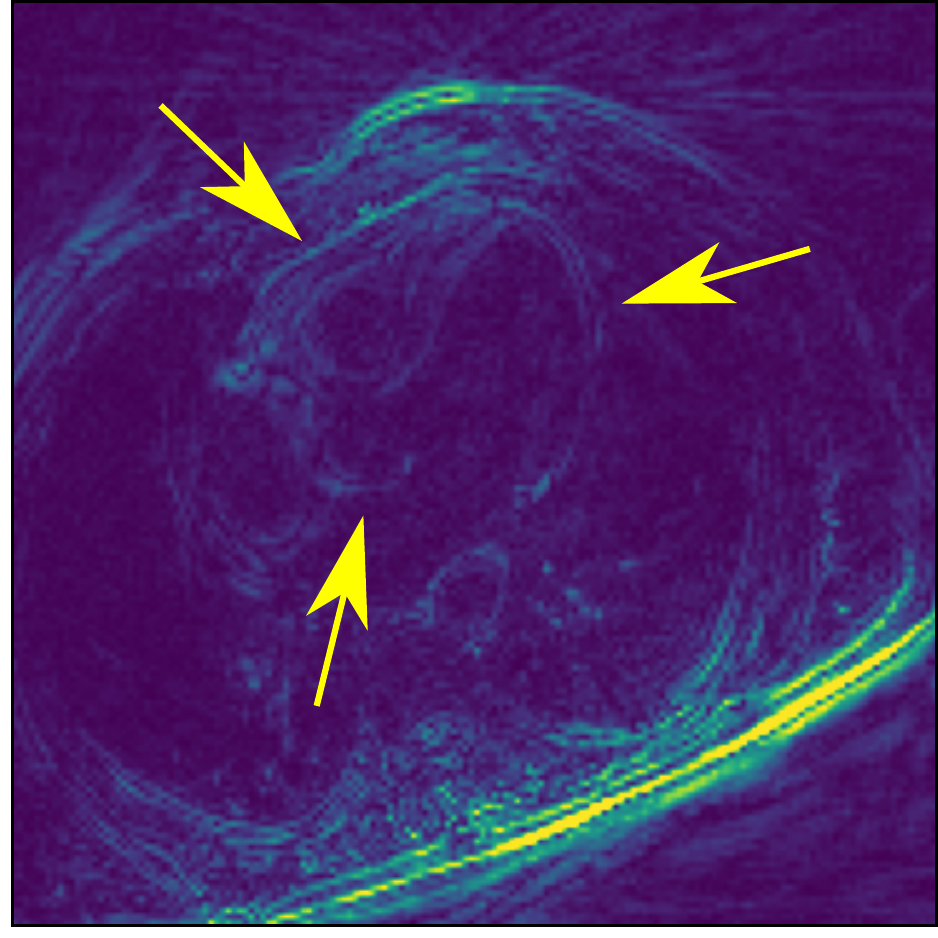}
 \put (75,87) {\tiny\textcolor{white}{\textbf{TV}}}
\end{overpic} \hspace{-0.2cm}
}
\resizebox{\linewidth}{!}{
\includegraphics[height=2.9cm]{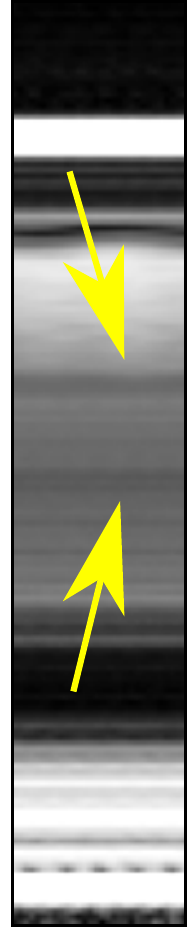}\hspace{-0.2cm}
\includegraphics[height=2.9cm]{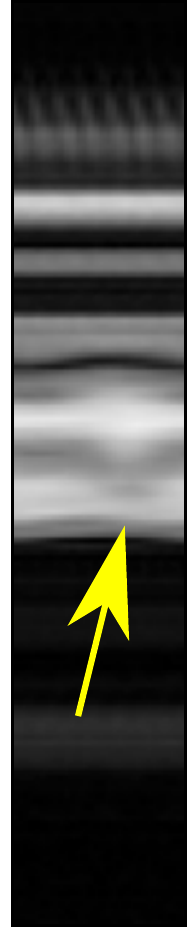}\hspace{-0.2cm}
\begin{overpic}[height=2.9cm,tics=10]{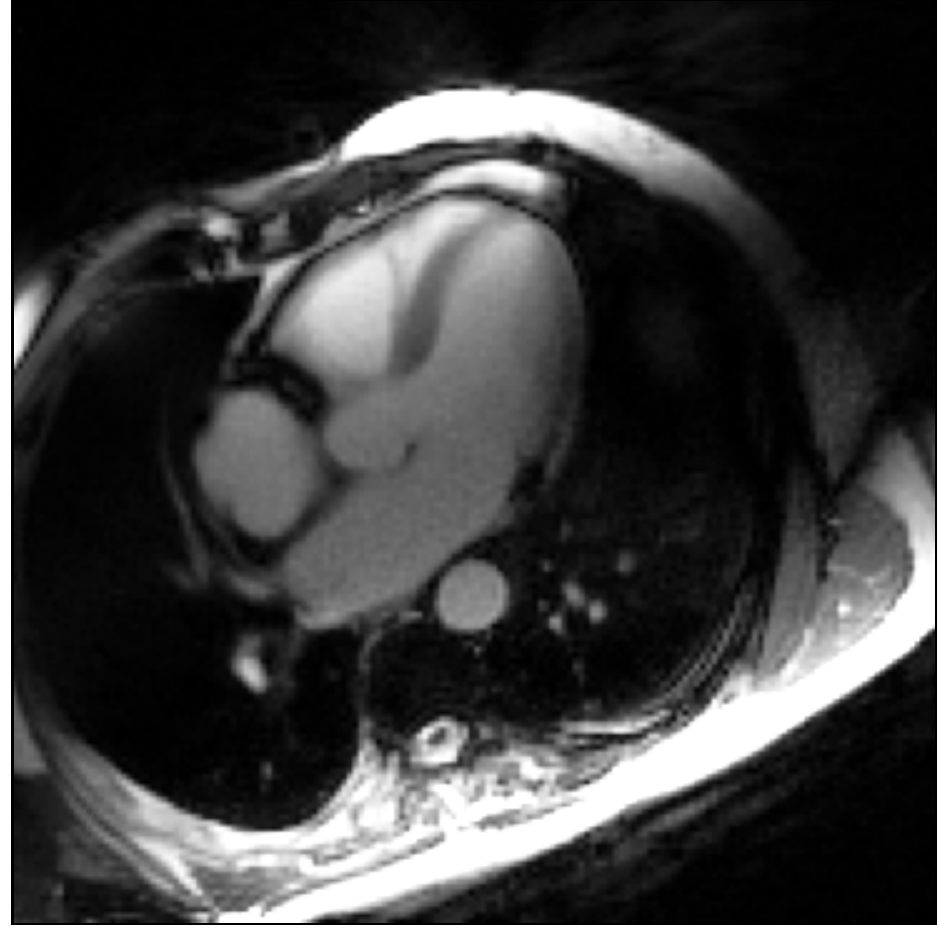}
\end{overpic} \hspace{-0.2cm}
\includegraphics[height=2.9cm]{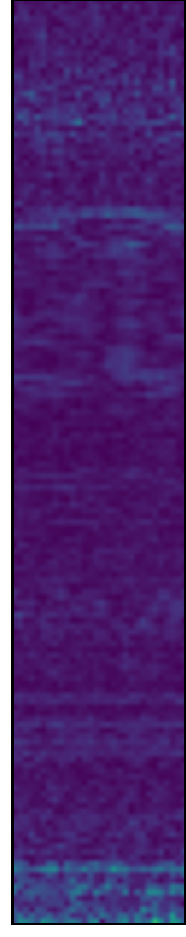}\hspace{-0.2cm}
\includegraphics[height=2.9cm]{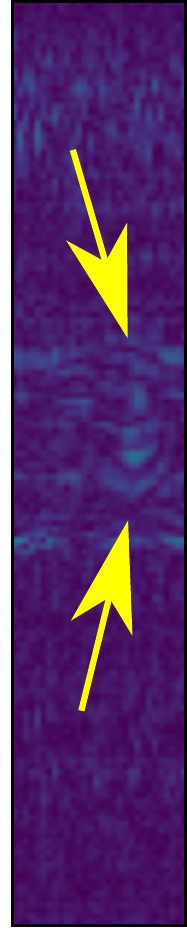}\hspace{-0.2cm}
\begin{overpic}[height=2.9cm,tics=10]{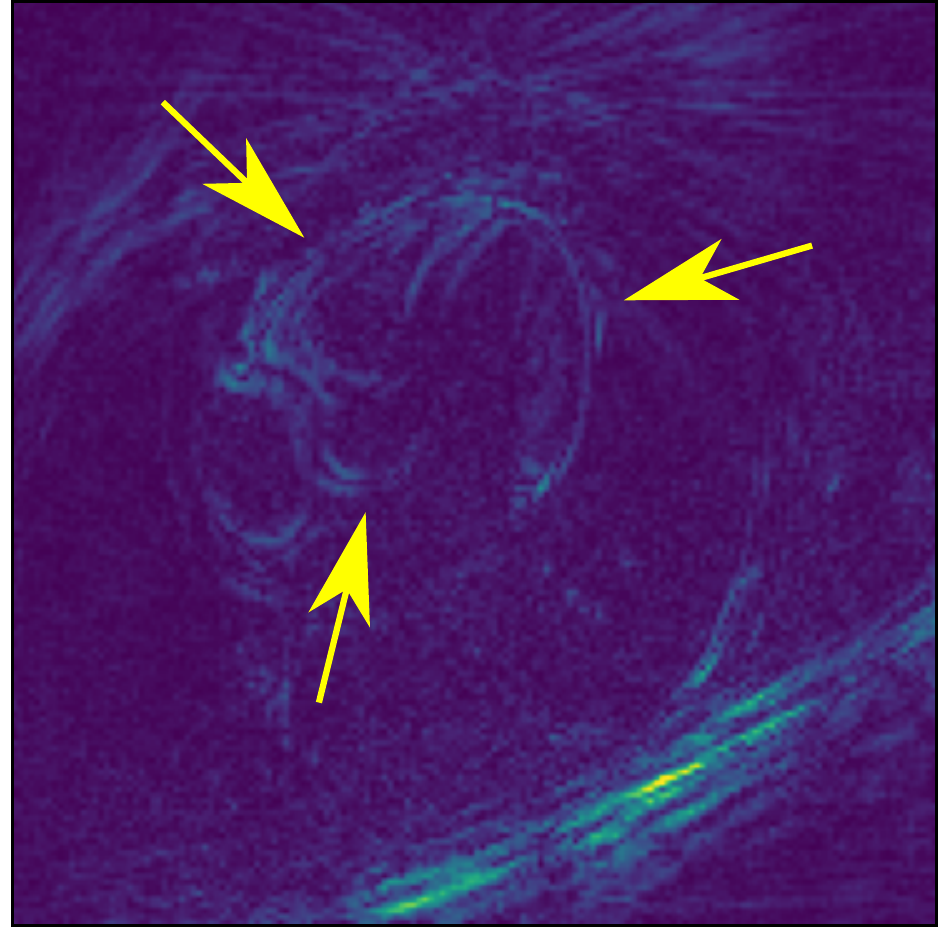}
 \put (46,87) {\tiny\textcolor{white}{\textbf{$kt$-SENSE}}}
\end{overpic} \hspace{-0.2cm}
}
\end{minipage}
\begin{minipage}{0.49\linewidth}
\resizebox{\linewidth}{!}{
\includegraphics[height=2.9cm]{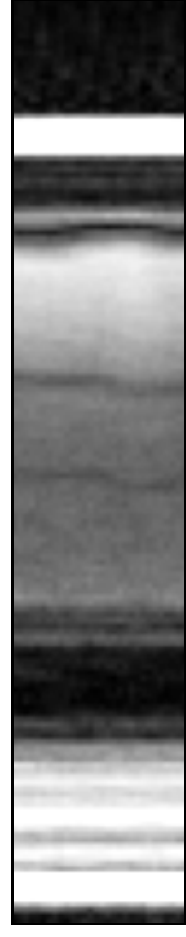}\hspace{-0.2cm}
\includegraphics[height=2.9cm]{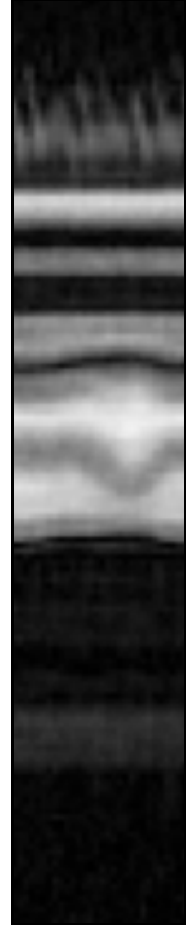}\hspace{-0.2cm}
\begin{overpic}[height=2.9cm,tics=10]{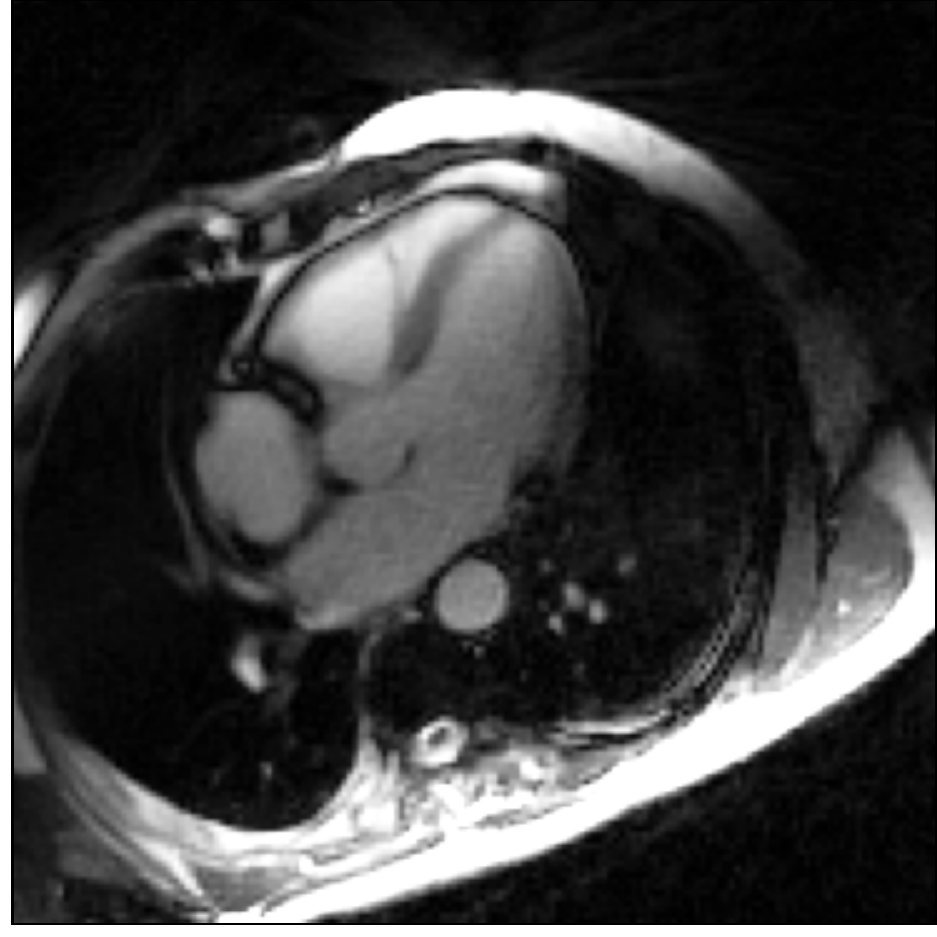}
\end{overpic} \hspace{-0.2cm}
\includegraphics[height=2.9cm]{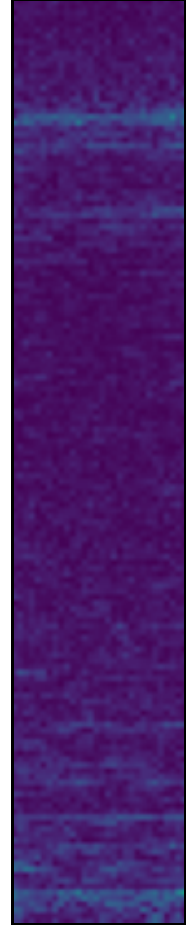}\hspace{-0.2cm}
\includegraphics[height=2.9cm]{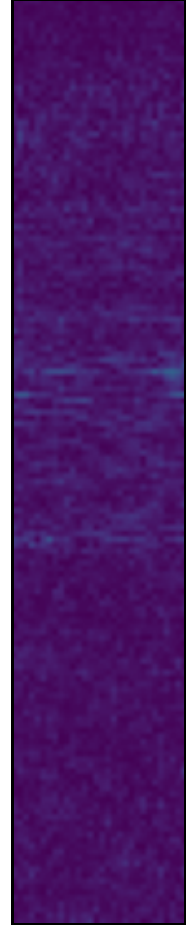}\hspace{-0.2cm}
\begin{overpic}[height=2.9cm,tics=10]{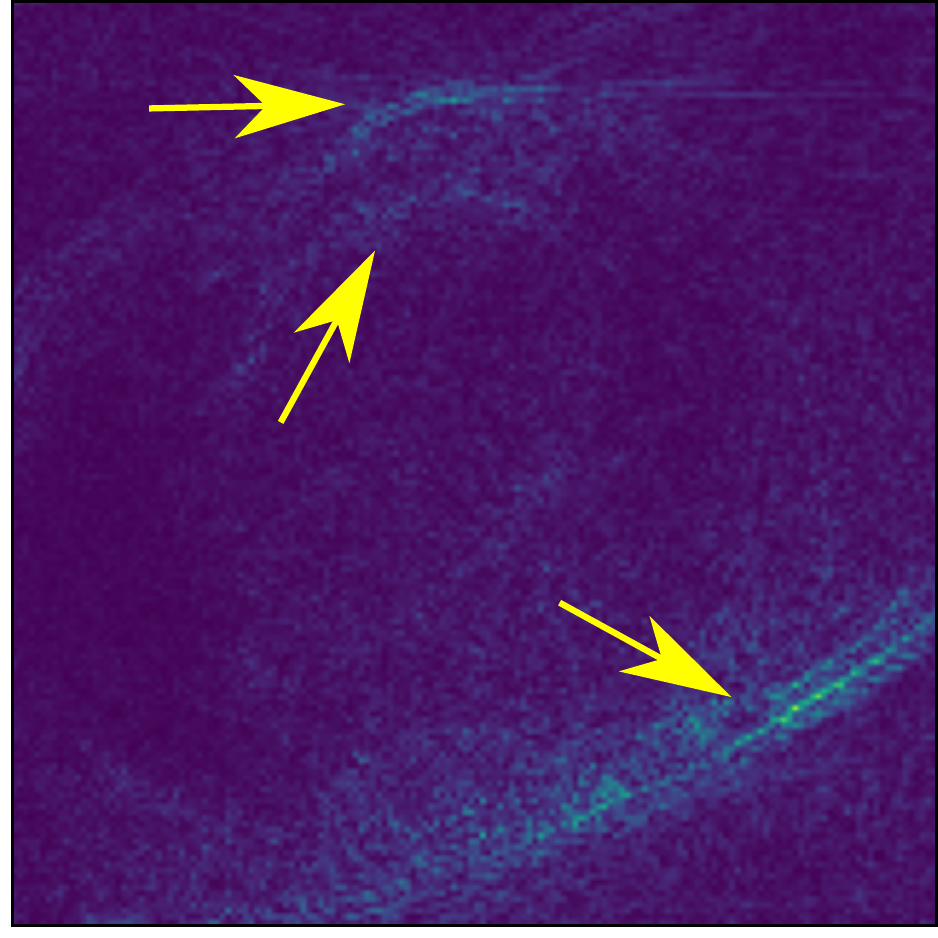}
 \put (72,87) {\tiny\textcolor{white}{\textbf{DIC}}}
\end{overpic} \hspace{-0.2cm}
}
\resizebox{\linewidth}{!}{
\includegraphics[height=2.9cm]{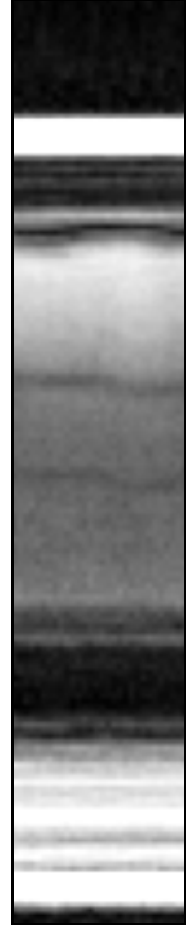}\hspace{-0.2cm}
\includegraphics[height=2.9cm]{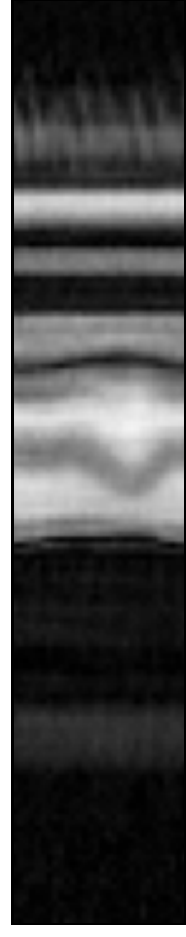}\hspace{-0.2cm}
\begin{overpic}[height=2.9cm,tics=10]{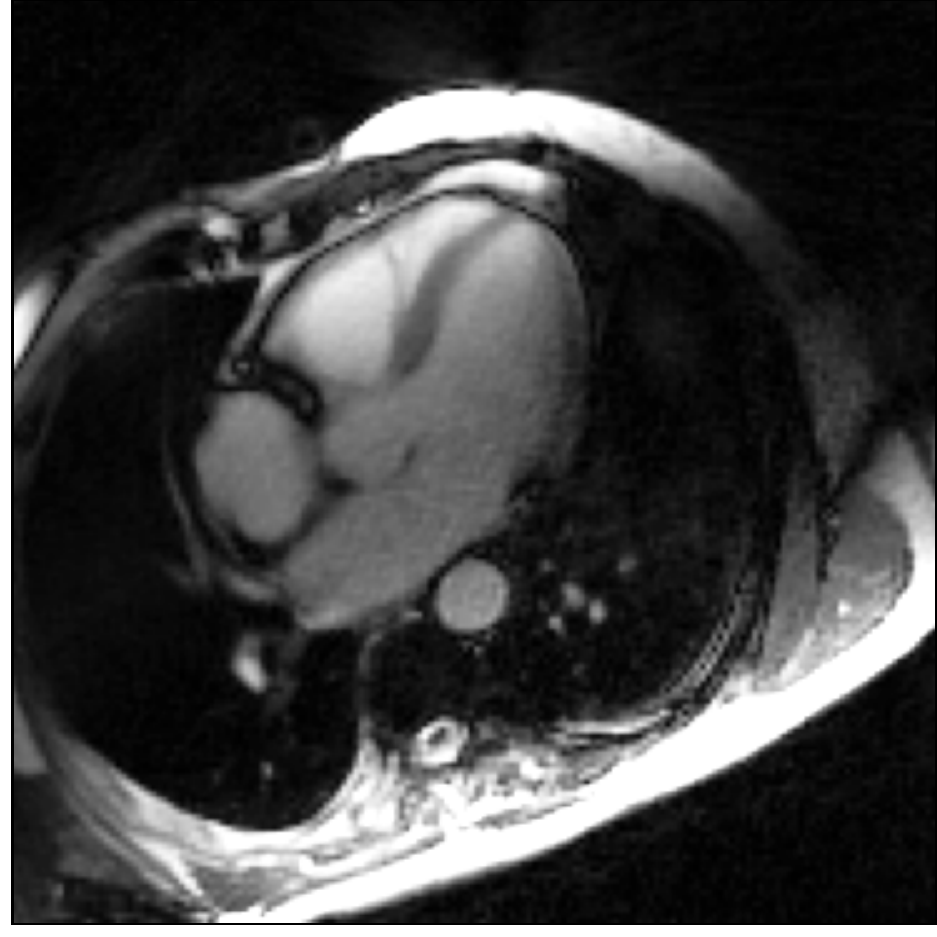}
\end{overpic} \hspace{-0.2cm}
\includegraphics[height=2.9cm]{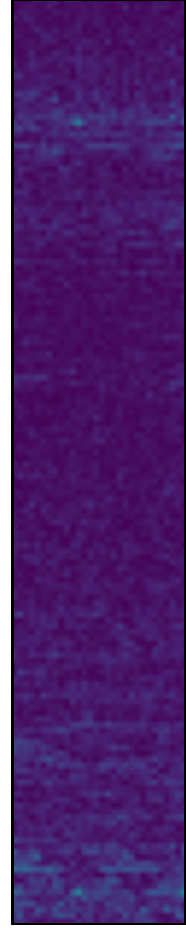}\hspace{-0.2cm}
\includegraphics[height=2.9cm]{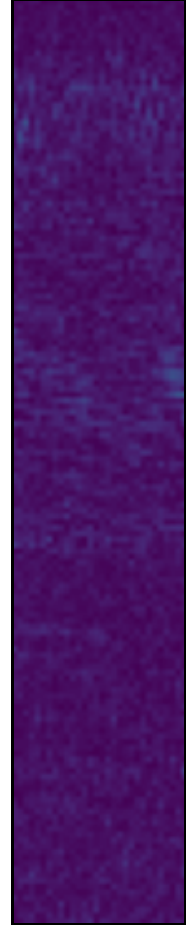}\hspace{-0.2cm}
\begin{overpic}[height=2.9cm,tics=10]{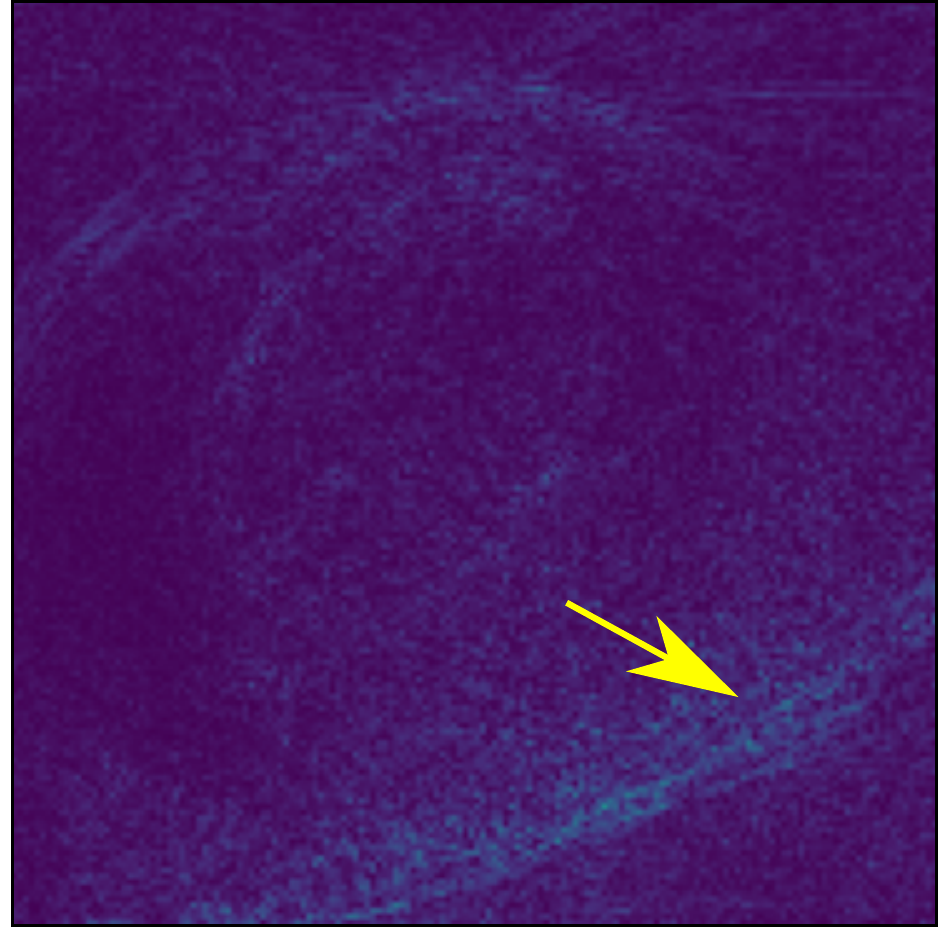}
 \put (26,87) {\tiny\textcolor{white}{\textbf{CNN-based IR}}}
\end{overpic} \hspace{-0.2cm}
}
\resizebox{\linewidth}{!}{
\includegraphics[height=2.9cm]{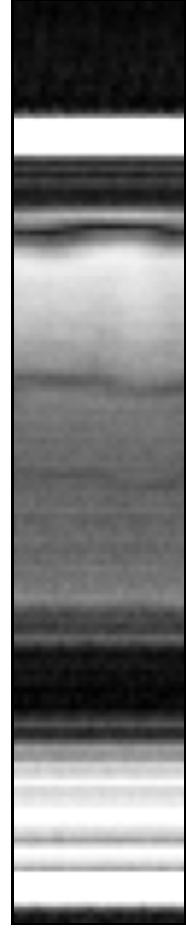}\hspace{-0.2cm}
\includegraphics[height=2.9cm]{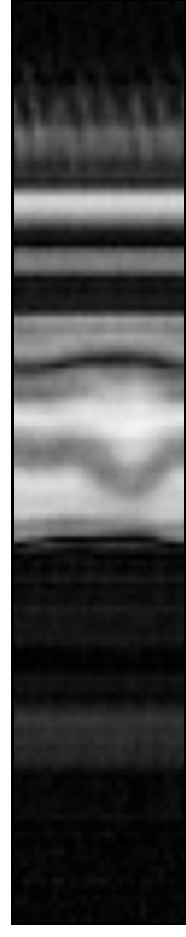}\hspace{-0.2cm}
\begin{overpic}[height=2.9cm,tics=10]{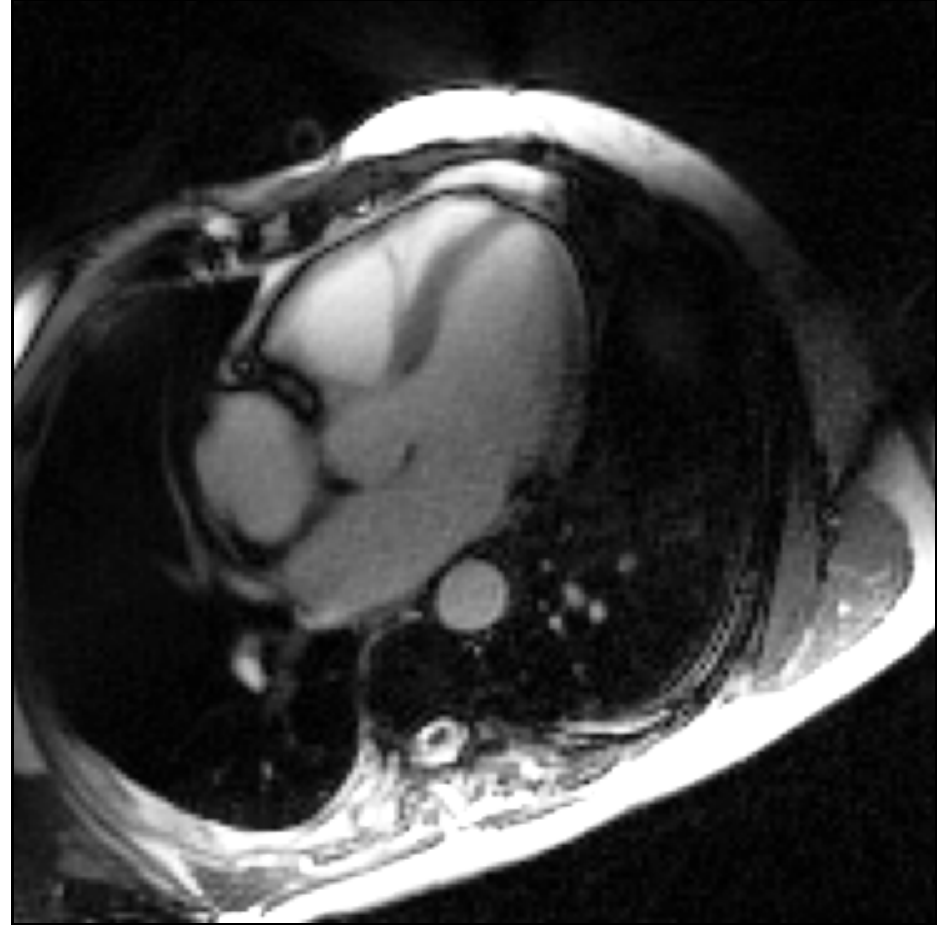}
\end{overpic} \hspace{-0.2cm}
\includegraphics[height=2.9cm]{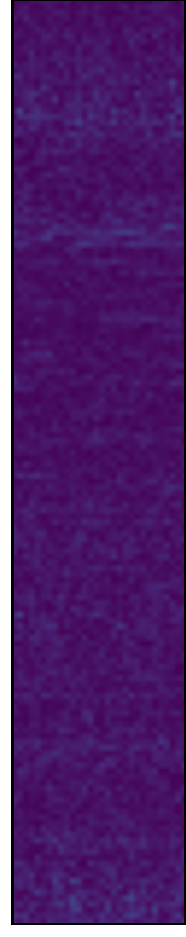}\hspace{-0.2cm}
\includegraphics[height=2.9cm]{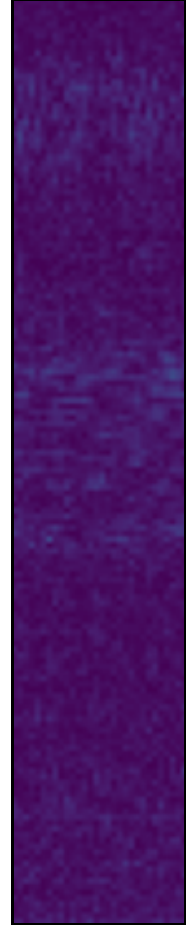}\hspace{-0.2cm}
\begin{overpic}[height=2.9cm,tics=10]{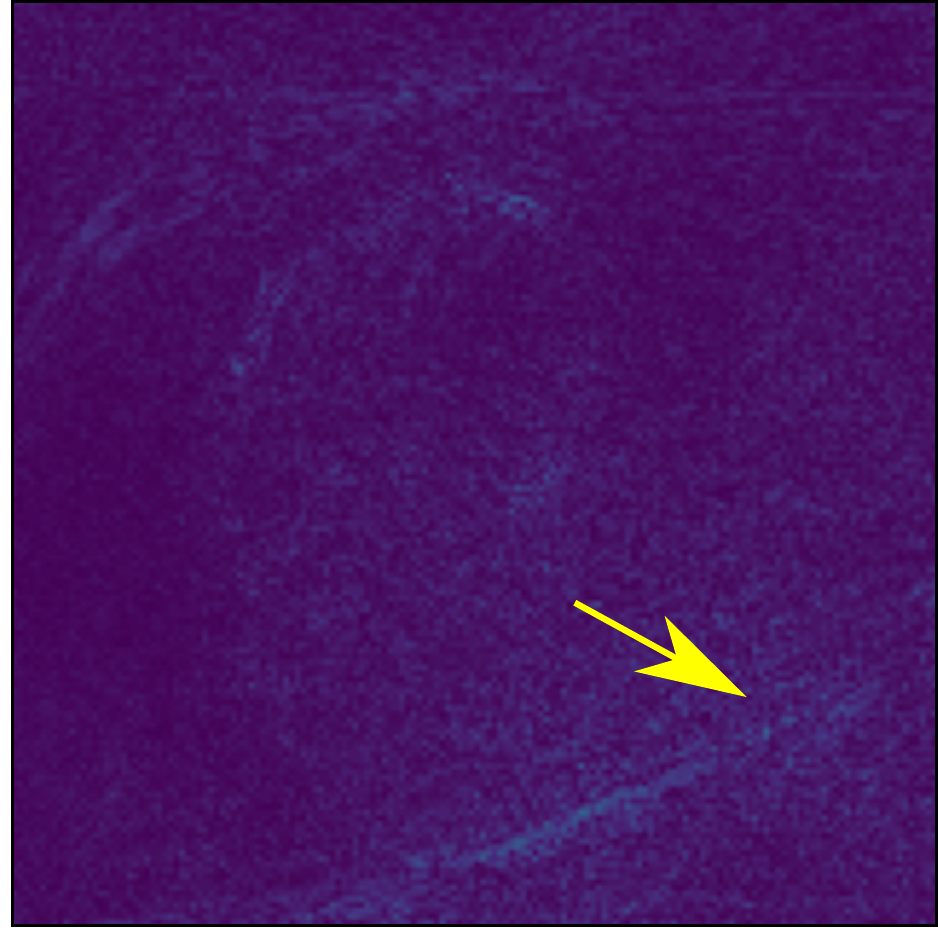}
 \put (53,87) {\tiny\textcolor{white}{\textbf{Proposed}}}
\end{overpic} \hspace{-0.2cm}
}
\resizebox{\linewidth}{!}{
\includegraphics[height=2.9cm]{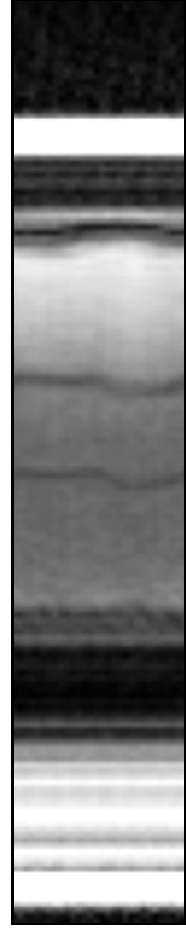}\hspace{-0.2cm}
\includegraphics[height=2.9cm]{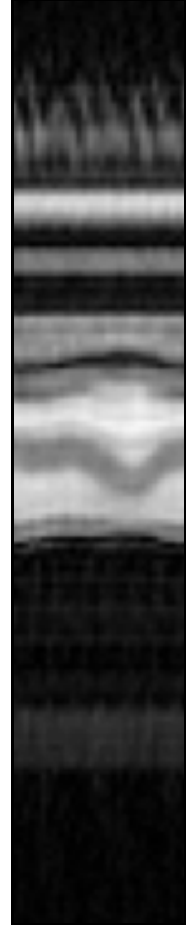}\hspace{-0.2cm}
\begin{overpic}[height=2.9cm,tics=10]{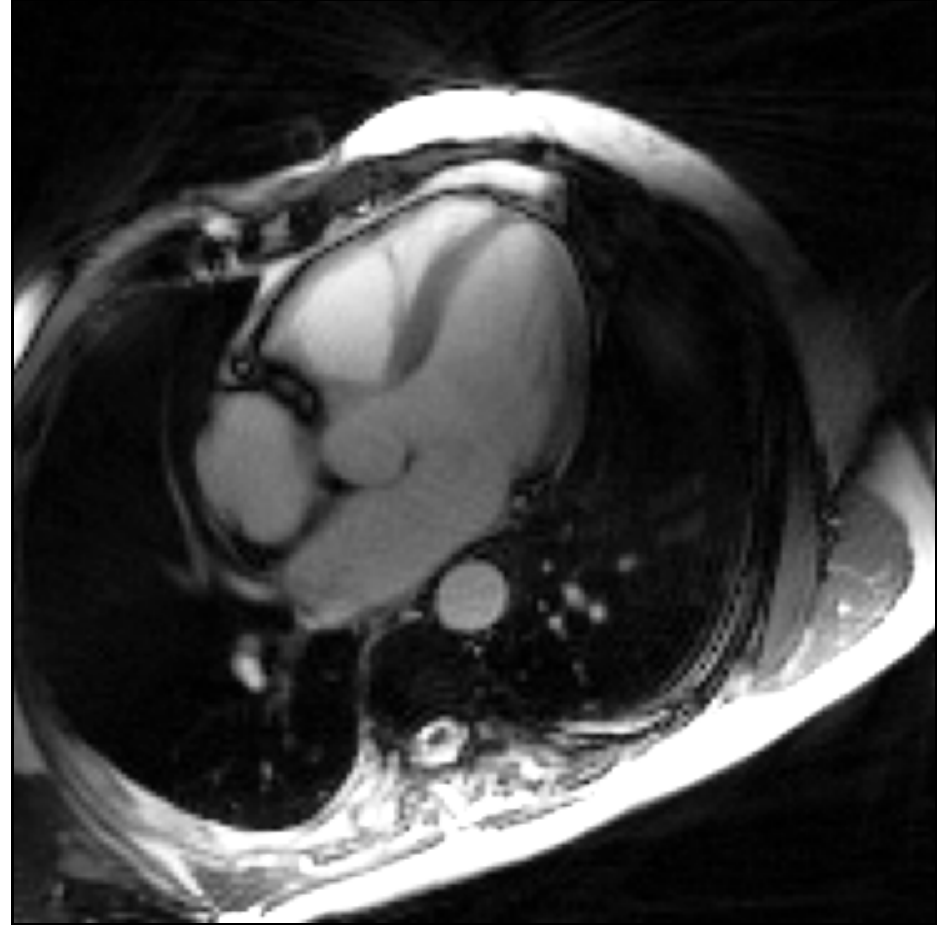}
\end{overpic} \hspace{-0.2cm}
\includegraphics[height=2.9cm]{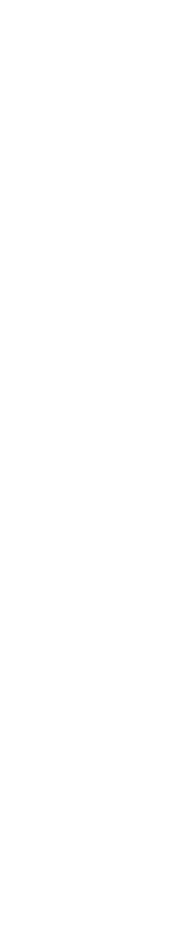}\hspace{-0.2cm}
\includegraphics[height=2.9cm]{images/results/white_images/white_xt_yt.pdf}\hspace{-0.2cm}
\begin{overpic}[height=2.9cm,tics=10]{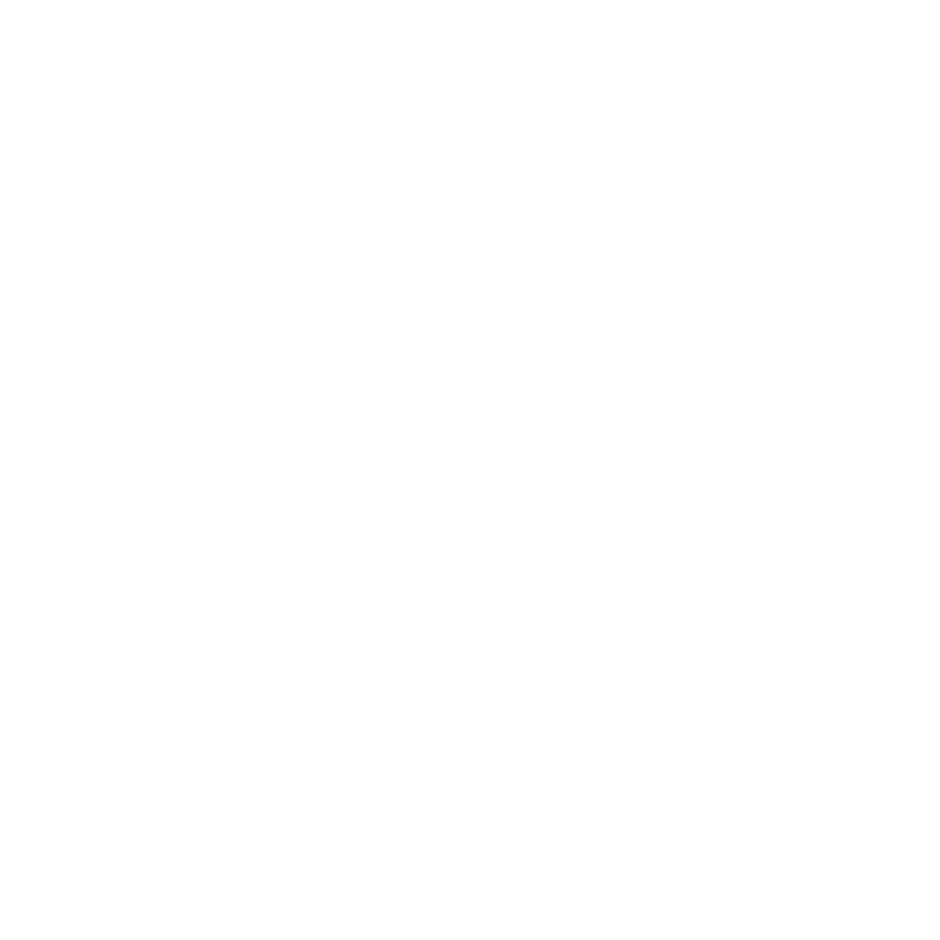}
\end{overpic} \hspace{-0.2cm}
}
\end{minipage}
\caption{Comparison of images and point-wise errors for different reconstruction methods with learned and non-learned regularization for $N_{\theta}=560$. Left column: Classical iterative reconstruction methods with - from top to bottom - direct NUFFT-reconstructon, iterative SENSE \cite{pruessmann2001advances}, total variation-minimization (TV) \cite{block2007} , $kt$-SENSE \cite{Tsao2003}. Right column: Learning-based regularization methods with - from top to bottom - adaptive dictionary learning and sparse coding (DIC) \cite{pali2020}, CNN-based iterative reconstruction (IR) \cite{kofler2020neural} using a 3D U-Net \cite{Hauptmann2019} and  our proposed approach. }\label{comparison_results_fig}
\end{figure}

Here, we compare our reconstruction method to the image reconstruction methods previously introduced. 
As discussed in Subsection \ref{subsec:MnCG_variation} we chose to use $M=12$ and $n_{\mathrm{CG}}=4$ although the network was trained with $M=1$ and $n_{\mathrm{CG}}=8$. 
Note that, since the same strategy seemed not to be consistently useful for the 3D U-Net which was trained without the integration of the encoding operators, we report here the values for $M=1$ and $n_{\mathrm{CG}}=12$ which are also the ones used in \cite{kofler2020neural}. In the supplementary material, the results for the variation of $M$ and $n_{\mathrm{CG}}$ can be found as well. The regularization parameter for the  dictionary learning method \cite{pali2020} and the CNN-based regularization method \cite{kofler2020neural} was chosen as $\lambda=1$. \\
Figure \ref{comparison_results_fig} shows an example of the results obtained with the previously described methods of comparison and our proposed approach. As can be seen, the total variation-minimization method as well as $kt$-SENSE successfully removed the undersampling artefacts but also led to a loss of image details as is indicated by the yellow arrows. In contrast, all learning-based method yielded a good reconstruction performance in terms of preservation of the cardiac motion. We see that our proposed method is in addition the one which best reduced residual image noise in the images.
Table \ref{table_reco_results} shows the results achieved in terms of the chosen measures. The best achieved results are highlighted as bold numbers. Again, the experiments were repeated for two different undersampling (i.e.\ acceleration) factors, given by sampling the $k$-space along  $N_{\theta}=560$ and $N_{\theta}=1130$ radial spokes, respectively.\\
The numbers well-reflect the observations from Figure \ref{comparison_results_fig} and we see that all methods based on regularization methods employing learning-based methods consistently  outperform the methods using hand-crafted regularization methods with respect to all measures. 
All three reported methods using machine learning-based regularization achieve competitive results, where we see that our proposed method  yields substantially better results compared to the dictionary learning (DL)-based method and the other CNN-based method in terms of error-based measures. 
In terms of image similarity-based measures, the difference between the dictionary learning-based method and ours becomes less prominent except for VIQP.  Interestingly, the 3D-U-Net-based method is consistently surpassed either by the dictionary learning-based method or our proposed one. All observations are consistent among both acceleration factors with $N_{\theta}=560$ and $N_{\theta}=1130$.
In addition, note that while the obtained results for DL, the 3D U-Net based iterative reconstruction and our proposed approach are similar, the average reconstruction time using DL amounts to approximately 2000 seconds, where the most computationally intensive part is the repeated sparse coding of the image patches at each iteration. In contrast, for the CNN-based methods, the reconstruction time amounts to approximately 12 and 48 seconds which are mainly coming from the CG-module. However, the  implementation of the dictionary learning and sparse coding algorithms aITKrM and aOMP is currently only available for running on the CPU and thus, a further acceleration could be expected. Further, our proposed method quite consistently surpasses the 3D U-Net-based reconstruction method even if it only contains around $9.1\%$ of the trainable parameters.

\begin{table*}[t]
	\centering
	\renewcommand{\arraystretch}{1.3}
	\scriptsize{
		\caption{Quantitative results for $N_{\theta}=560$ $N_{\theta}=1130$ radial spokes which corresponds to an acceleration factor of $\sim 9$  and $\sim 18$, respectively}. \label{table_reco_results}
		\vspace{0.2cm}
		\centering
		\begin{tabular}{ |l|
				c
				c
				c
				c|
				c
				c
				c|
			}
	\bottomrule 
	& \multicolumn{4}{c|}{\textbf{Non-Learned Regularization}} & \multicolumn{3}{c|}{\textbf{Learned Regularization}}\\
	\midrule
	& \textbf{NUFFT} & \textbf{It SENSE}  &  \textbf{TV} & \textbf{$kt$-SENSE} & \textbf{DL}  & \textbf{3D U-Net + IR} &  \textbf{Proposed} \\
			\midrule
& \multicolumn{7}{c|}{\textbf{Number of Radial Spokes:} $N_{\theta} = 560$}\\
\midrule	 
\textbf{PSNR} 		&34.9542 &40.5610 &41.8778 &44.2316 &45.6085 &46.0845 & \bf{47.4396} \\
\textbf{NRMSE} 		& 0.2913 & 0.1535 & 0.1313 & 0.0986 & 0.0844 & 0.0800 & \bf{0.0697} \\
\textbf{SSIM} 		& 0.8843 & 0.9686 & 0.9786 & 0.9820 & 0.9885 & 0.9880 & \bf{0.9895}  \\
\textbf{MS-SSIM} 	& 0.9762 & 0.9935 & 0.9950 & 0.9958 & 0.9980 & 0.9979 &  \bf{0.9982}  \\
\textbf{UQI} 		& 0.7574 & 0.8827 & 0.8889 & 0.8887 & 0.9347 & 0.9288 & \bf{0.9388}  \\
\textbf{VIQP}		& 0.8339 & 0.8379 & 0.8234 & 0.8931 & 0.9256 & 0.9340 & \bf{0.9620}\\
\textbf{HPSI} 		& 0.9456 & 0.9811 & 0.9849 & 0.9904 & 0.9950 & 0.9949 & \bf{0.9961} \\

\midrule
& \multicolumn{7}{c|}{\textbf{Number of Radial Spokes:} $N_{\theta} = 1130$}\\

\textbf{PSNR} 		&39.0289 &43.8077 &44.1841 &46.0096 &47.4373 &47.5094 & \bf{48.6761}\\
\textbf{NRMSE} 		& 0.1864 & 0.1067 & 0.0998 & 0.0817 & 0.0682 & 0.0681 & \bf{0.0606}\\
\textbf{SSIM} 		& 0.9378 & 0.9778 & 0.9841 & 0.9875 & 0.9913 & 0.9906 &  \bf{0.9916}\\
\textbf{MS-SSIM} 	& 0.9892 & 0.9965 & 0.9965 & 0.9976 & \bf{0.9986} & 0.9985 & \bf{0.9986}\\
\textbf{UQI} 		& 0.8363 & 0.9156 & 0.9028 & 0.9224 & \bf{0.9484} & 0.9420 & 0.9480\\
\textbf{VIQP}		& 0.9228 & 0.9513 & 0.8832 & 0.9498 & 0.9521 & 0.9547 & \bf{0.9695} \\
\textbf{HPSI} 		& 0.9775 & 0.9923 & 0.9903 & 0.9943 & \bf{0.9972} & 0.9966 &  \bf{0.9972}\\
\midrule      
\textbf{Parameters} & - &  - & - & -	& 	\bf{9\,664}  & 1\,024\,224  & 93\,617 \\ 
\textbf{Backend} & CPU & CPU & CPU & CPU & CPU/GPU & GPU & GPU \\
\midrule

		\end{tabular}	
	}
	  
	\vspace{0.3cm}
\end{table*}

\section{Discussion}\label{sec:discussion}

In this work, we have proposed the first end-to-end trainable iterative reconstruction network for dynamic multi-coil MR image reconstruction using non-uniform sampling schemes. As we have seen, the proposed end-to-end trainable reconstruction network provides a competitive method for image reconstruction for 2D cine MR image reconstruction with non-Cartesian multi-coil encoding operators. In the following, we highlight several advantages and limitations of our proposed approach and put it in relation to other works.

\subsection{End-To-End Trainability}
Since the considered forward model is computationally demanding, methods as \cite{hyun2018deep}, \cite{kofler2020neural} can be used as an alternative, where the generation of the CNN-output and the step of increasing data-consistency are decoupled from each other. However, the major advantage of our proposed network over methods similar to \cite{hyun2018deep} or \cite{kofler2020neural},  is that the reconstruction network is trained in and-end-to-end manner. As can be seen in Figure \ref{training_process_fig} in Subsection \ref{subsec:training_behaviour}, including the DC-module given by the CG-method in the network architecture is highly beneficial since it further reduces the training- and validation-error and also reduces the gap between the both, leading to a better generalization power. This result experimentally confirms the theory on the achievable performance of the reconstruction network derived in \cite{maier2019learning}.
Further, we demonstrated that the proposed training strategy is a viable option for training the entire network in an end-to-end manner in a relatively short amount of time while achieving good convergence properties of the network parameters.\\

\subsection{The Choice of the CNN-Block}

As we have seen in Figure \ref{training_process_fig}, the proposed CNN-block's trainable parameters converge relatively fast (approximately 6 hours) to a set of suitable trainable parameters in the pre-training stage. Training the 3D U-Net on the other hand took approximately 3 days. Because the proposed CNN-block is trained relatively fast, fine-tuning the entire network is possible by investing approximately 5 days of further training.\\ 
Note that our proposed approach transfers the learning of the artefact-reduction mapping to from 3D to 2D by the application of 2D convolutional layers in the temporally Fourier-transformed spatio-temporal domain. Thus, it inherits all benefits method presented in \cite{kofler2019spatio}. The most important property is that due to the change of perspective on the data, for each single cine MR image, $N_x + N_y$ samples are actually considered to train the CNN-block which is essential for training. Since only a relatively small amount of training data in terms of number of subjects  already suffices for a successful training, the end-to-end-training, which is particularly computationally expensive during the fine-tuning stage, can be carried out in a reasonable amount of time without the need to additionally augment the dataset in order to prevent overfitting.\\
Note that, of course, if enough training data is available and the hardware constraints allow it, training the network with computationally heavier CNN-blocks is possible as well. Thus, the proposed method can also be seen as a general method for training an end-to-end reconstruction network for large-scale image reconstruction problems with computationally demanding forward operators.

\subsection{The Trade-Off Between the Hyper-Parameters $M$ and $n_{\mathrm{CG}}$}

From the formulation of the network architecture in \eqref{iterative_CNN}, we can identify two main hyper-parameters which can be varied and determine the nature of the proposed reconstruction algorithm.
The overall number of iterations $M$ defines the length of the un-rolled iterative scheme. Further, because in general, minimizing functional \eqref{CNN_reg_theta_fixed_problem} requires solving a linear system using an iterative solver, the number of iterations to approximate the solution of \eqref{CNN_reg_theta_fixed_problem}, here named $n_{\mathrm{CG}}$, has to be chosen as well.
By setting $M=1$ and $n_{\mathrm{CG}}$ relatively "high", say $n_{\mathrm{CG}}=12$, one aims at constructing an end-to-end trainable network which conceptually resembles methods like \cite{hyun2018deep} and \cite{kofler2020neural}. There, the CNN-block is only applied once to obtain a CNN-based image-prior and functional \eqref{CNN_reg_theta_fixed_problem} is minimized until convergence of the iteration. On the other hand, one can also set $M>1$, but because of hardware constraints, one necessarily also has to either lower  $n_{\mathrm{CG}}$ in each CG-block,  reduce the complexity of the CNN-blocks  or in the worst case, both. Lowering the number of CG-iterations $n_{\mathrm{CG}}$ causes the solution of \eqref{CNN_reg_theta_fixed_problem} most probably to only be poorly approximated and lowering the CNN-blocks complexity can be expected to deliver poorer intermediate outputs of the different CNN-blocks in terms of artefacts-reduction.\\
Interestingly, we have observed that even fine-tuning the entire network with $M=1$ and $n_{\mathrm{CG}}=8$ allowed to change the hyper-parameters at test time by further obtaining a boost in performance, see the supplementary material for more details.
We also trained an iterative network architecture with  our proposed CNN-block for $M=3$ and $n_{\mathrm{CG}}=2$. Due to hardware constraints, we employed smaller 2D U-Nets as CNN-blocks, where, different to before, we set the initial number of applied filters to $n_f=4$. The so-constructed network consisted of only 5908 trainable parameters, i.e.\ only about 0.57$\% $ of the 3D U-Net.\\
In the supplementary material, one can see a comparison of our method fine-tuned with $M=1$ and $n_{\mathrm{CG}}=8$ against $M=3$ and  $n_{\mathrm{CG}}=2$ which were evaluated with different configurations of $M$ and $n_{\mathrm{CG}}$ at test time. The network which was fine-tuned with $M=1$ and $n_{\mathrm{CG}}=8$ consistently outperforms the other with respect to all measures. Altough the comparison is not entirely fair since the number of trainable parameters highly differs from one CNN-block to the other, this result is important because of the following reason. It suggests that sacrificing expressiveness of the CNN-block in terms of trainable parameters in order be able to fine-tune with $M>1$ seems not to be necessary since also for the network fine-tuned with $M=1$ and $n_{\mathrm{CG}}=8$, different $M$ and $n_{\mathrm{CG}}$ can be used at test time and further increase the performance. Interestingly, we were not able to observe this phenomenon for the 3D U-Net. Thus, we attribute this property to the fine-tuning stage in which the encoding operator is included in the network architecture which again highlights the importance of employing a CNN-block which allows end-to-end training of the entire network in a reasonable amount of time.

\subsection{Limitations}

For the proposed method, training times tend to be quite long, amounting to several days. This makes the algorithmic development challenging in terms of hyper-parameter tuning and limits the  possibility to draw general conclusions because repeating experiments for different parameter configurations is prohibitive. Nevertheless, the proposed training scheme shows that it is easily possible to outperform post-processing methods as in \cite{Hauptmann2019}, \cite{kofler2019spatio} or the approaches in \cite{hyun2018deep} and \cite{kofler2020neural}, where training of the CNN-block is decoupled from the subsequent minimization of a CNN-regularized functional.\\
Further, although we have seen that it is possible to increase $M$ at test time and observe an increase of performance in terms of reconstruction results, from a theoretical point of view, it remains somewhat unclear why this is exactly possible. We believe that the reason for this lies the end-to-end training the entire network but leave a rigorous theoretical investigation and a convergence analysis as future work.

\subsection{Differences and Similarities to Other Works}

Our presented approach shares similarities across different works. First, for the case $M=1$ it can be seen as an extension of the approaches presented in \cite{hyun2018deep}, \cite{kofler2019spatio}, \cite{schwab2018deep} in the sense that we only generate only one CNN-based image-prior which is used in a Tikhonov functional. However, because of the proposed end-to-end training strategy, the obtained CNN-based image-prior tends to be much a better estimate of the ground-truth image compared to the cases when the training of the CNN-block is decoupled. Further, for $M>1$, the structure of the network is similar to \cite{schlemper2017deep}, \cite{aggarwal2018modl}, with the difference that first of all, the considered inverse problem is different (radial multi-coil instead of Cartesian single-coil) and thus the DC-module is a CG-block instead of the implementation of a closed-form solution to \eqref{CNN_reg_theta_fixed_problem}. Second, our CNN-block consists of a spatio-temporal 2D U-Net which is  applied in the Fourier-transformed spatio-temporal domain.\\
In our work, the 2D U-Nets are applied in the temporally Fourier-transformed spatio-temporal domain and thus use the same change of perspective on the data as in  \cite{kofler2019spatio}. However, in \cite{kofler2019spatio}, the slices extraction and re-assembling process is not part of the network architecture and thus does not allow end-to-end training. Further, we apply the U-net after having performed a temporal FFT, similar to \cite{qin2019k}.\\
In \cite{biswas2019dynamic}, where a non-Cartesian multi-coil dynamic acquisition is considered,  the measured $k$-space data is first interpolated onto a Cartesian grid. After this interpolation, a simple FFT can be used as forward operator and thus facilitates the construction of an iterative network. However, because the $k$-space data-interpolation is decoupled from the network, the network cannot learn to compensate for the interpolation errors. Thus, in order to really utilize the measured $k$-space data, applying the actual encoding operator (which involves the gridding-step)  which is associated to the considered image reconstruction problem is unavoidable.\\
In contrast, in our approach, the gridding of the measured $k$-space is part of the network architecture. This important difference is the main motivation of the work since due to the computational difficulties linked with the integration of the non-uniform FFT in the network architecture, a computationally light CNN-block  has to be used.\\
Although the method shares similar features to other works, this is - to the best our knowledge - the first work to combine several components to construct a network architecture which is trainable in an end-to-end manner for a dynamic non-uniform multi-coil MR image reconstruction problem by actually using the radially acquired data and not the one interpolated onto a Cartesian grid.
In fact, we believe that the large-scale of the considered problem is the reason for the lack of end-to-end trainable reconstruction networks for dynamic non-Cartesian data-acquisition protocols using multiple receiver coils.

\section{Conclusion}\label{sec:conclusion}

In this work, we have proposed a new end-to-end trainable data-consistent reconstruction network for accelerated 2D dynamic MR image reconstruction with non-uniformly sampled data using multiple receiver coils. Further, since end-to-end training is computationally expensive because of the forward and the adjoint operators are included in the network, we have proposed and investigated an efficient training strategy to circumvent this issue. In addition, we have compared our method to other well-established iterative reconstruction methods as well as several methods based on learned regularization. Our  proposed method surpassed all methods using non-learned regularization methods and achieved competitive results compared to a dictionary learning-based method and a method based on CNN-based image-priors.
Although the method was presented for 2D radial cine MRI, we expect the training strategy  to be applicable to general image reconstruction problems with computationally expensive operators as well. Further, we expect the proposed CNN-block to be applicable to arbitrary reconstruction problems with a time component and temporal correlation.

\newpage

\cen{\sf {\Large {\bfseries An End-To-End-Trainable Iterative Network Architecture for Accelerated Radial Multi-Coil 2D Cine MR Image Reconstruction} \\  
\vspace{0.5cm}
{\bfseries Supplementary Material} \\ 
		\vspace*{10mm}
		}
		}

\pagenumbering{roman}
\pagestyle{plain}

\setcounter{section}{0}
\setcounter{table}{0}
\setcounter{figure}{0}

 \section{Variation of the Hyper-Parameters $M$ and $n_{\mathrm{CG}}$}
 
Table \ref{shallow_table} shows a comparison of the proposed network which was fine-tuned with $M=3$ and $n_{\mathrm{CG}}=2$ for different $M$ and $n_{\mathrm{CG}}$ which were chosen at test time. The CNN-block is a U-Net with only $5\,908$ trainable parameters.\\
As can be seen, increasing $M$ tends to yield better results, especially in terms of the error-based measures PSNR and NRMSE, while for the image-similarity-based measures, the performance tends to only slightly increase.

\begin{table*}[h]
	\centering
	\renewcommand{\arraystretch}{1.3}
	\footnotesize{
		\caption{Quantitative results for our proposed method fine-tuned with $M=3$ and $n_{\mathrm{CG}}=2$  for a different number of outer iterations $M$ with fixed $n_{\mathrm{CG}}=4$.} \label{shallow_table}
		\vspace{0.2cm}
		\centering
		\begin{tabular}{ |l|
				c
				c
				c
				c
				c
				c|
			}
	\bottomrule 

	& \textbf{$M=2$} &\textbf{$M=4$}  &  \textbf{$M=6$} & \textbf{$M=8$} & \textbf{$M=10$}  & \textbf{$M=12$} \\
			\midrule
& \multicolumn{6}{c|}{\textbf{Number of Radial Spokes:} $N_{\theta} = 560$}\\
\midrule	 
\textbf{PSNR} 		&45.8219 &46.3317 &46.5028 &46.5664 &46.5899 &\bf{46.5972}\\
\textbf{NRMSE} 		& 0.0830 & 0.0785 & 0.0772 & 0.0767 & \bf{0.0766} & \bf{0.0766}\\
\textbf{SSIM} 		& 0.9866 & 0.9876 & 0.9878 & \bf{0.9879} & \bf{0.9879} & 0.9878\\
\textbf{MS-SSIM} 	& 0.9974 & 0.9976 & \bf{0.9977} & \bf{0.9977} & \bf{0.9977} & \bf{0.9977}\\
\textbf{UQI} 		& 0.9258 & 0.9303 & 0.9312 & \bf{0.9314} & 0.9313 & 0.9312\\
\textbf{VIQP}		& 0.9418 & 0.9470 & 0.9506 & 0.9527 & 0.9539 & \bf{0.9547}\\
\textbf{HPSI} 		& 0.9939 & 0.9947 & 0.9949 & \bf{0.9950} & \bf{0.9950} & \bf{0.9950}\\

\midrule
& \multicolumn{6}{c|}{\textbf{Number of Radial Spokes:} $N_{\theta} = 1130$}\\
\textbf{PSNR} 		&47.3795 &47.7015 &47.7920 &47.8245 &47.8383 &\bf{47.8448}\\
\textbf{NRMSE} 		& 0.0700 & 0.0676 & 0.0669 & 0.0667 & \bf{0.0666} & \bf{0.0666}\\
\textbf{SSIM} 		& 0.9898 & 0.9902 & 0.9903 & 0.9904 & \bf{0.9904} & \bf{0.9904}\\
\textbf{MS-SSIM} 	& 0.9981 & \bf{0.9982} & \bf{0.9982} & \bf{0.9982} & \bf{0.9982} & \bf{0.9982}\\
\textbf{UQI} 		& 0.9377 & 0.9404 & 0.9409 & \bf{0.9411} & \bf{0.9411} & \bf{0.9411}\\
\textbf{VIQP}		& 0.9569 & 0.9591 & 0.9609 & 0.9620 & 0.9626 & \bf{0.9630}\\
\textbf{HPSI} 		& 0.9959 & 0.9963 & 0.9963 & \bf{0.9964} &\bf{ 0.9964} & \bf{0.9964}\\

			\bottomrule 
		\end{tabular}	
	}
	   
	\vspace{0.3cm}
\end{table*}

Table \ref{deep_table} shows an analogous comparison for our proposed method with the difference that it was trained with $M=1$ and $n_{\mathrm{CG}}=8$ and the CNN-block consists of  a U-Net with $93\,617$ trainable parameters. Also for this hyper-parameter configuration, increasing $M$ yields in general better results. This result is important as having $M=1$ is in general computationally more manageable since the CG-block only has to be applied once. Thus, the results shown here suggest that it is possible to change the configuration of $M$ and $n_{\mathrm{CG}}$ at test time without the need of having $M>1$ during the fine-tuning stage.
Note that in contrast, for the 3D U-Net which was simply trained on input-output pairs without the integration of the physical models, increasing $M$ at test time seems not to be particularly beneficial.

\begin{table*}[t]
	\centering
	\renewcommand{\arraystretch}{1.3}
	\footnotesize{
		\caption{Quantitative results for our proposed method fine-tuned with $M=1$ and $n_{\mathrm{CG}}=8$  for a different number of outer iterations $M$ with fixed $n_{\mathrm{CG}}=4$.} \label{deep_table}
		\vspace{0.2cm}
		\centering
		\begin{tabular}{ |l|
				c
				c
				c
				c
				c
				c
				c
				c|
			}
	\bottomrule 

	& \textbf{$M=1$} & \textbf{$M=2$} & \textbf{$M=3$} &\textbf{$M=4$}  &  \textbf{$M=6$} & \textbf{$M=8$} & \textbf{$M=10$}  & \textbf{$M=12$} \\
	& \textbf{$n_{\mathrm{CG}}=12$} & \textbf{$n_{\mathrm{CG}}=6$} & \textbf{$n_{\mathrm{CG}}=4$} &\textbf{$n_{\mathrm{CG}}=4$}  &  \textbf{$n_{\mathrm{CG}}=4$} & \textbf{$n_{\mathrm{CG}}=4$} & \textbf{$n_{\mathrm{CG}}=4$}  & \textbf{$n_{\mathrm{CG}}=4$} \\
			\midrule
& \multicolumn{8}{c|}{\textbf{Number of Radial Spokes:} $N_{\theta} = 560$}\\
\midrule	 
\textbf{PSNR} 		&46.0099 &46.9338 &47.1193 &47.2919 &47.4071 &47.4196 &47.4277 & \bf{47.4396}   \\
\textbf{NRMSE} 		& 0.0809 & 0.0731 & 0.0718 & 0.0706 & 0.0700 & 0.0697 & 0.0698 & \bf{0.0697}   \\
\textbf{SSIM} 		& 0.9873 & 0.9890 & 0.9892 & 0.9894 & 0.9893 & 0.9895 & 0.9894 & \bf{0.9895}   \\ 
\textbf{MS-SSIM} 	& 0.9977 & 0.9981 & \bf{0.9982} & \bf{0.9982} & \bf{0.9982} & \bf{0.9982} & \bf{0.9982} & \bf{0.9982}   \\
\textbf{UQI} 		& 0.9275 & 0.9360 & 0.9368 & 0.9380 & 0.9386 & 0.9387 & 0.9387 & \bf{0.9388}   \\ 
\textbf{VIQP}		& 0.9436 & 0.9480 & 0.9500 & 0.9535 & 0.9628 & 0.9581 & 0.9606 & \bf{0.9620}   \\
\textbf{HPSI} 		& 0.9941 & 0.9956 & 0.9958 & 0.9960 & 0.9960 & \bf{0.9961}  & 0.9960 & \bf{0.9961}   \\
\midrule

& \multicolumn{8}{c|}{\textbf{Number of Radial Spokes:} $N_{\theta} = 1130$}\\
\midrule	
\textbf{PSNR} 		&47.7514 &48.3989 &48.5259 &48.6130 &48.6691 &48.6666 &48.6724 &\bf{48.6761}  \\
\textbf{NRMSE} 		& 0.0673 & 0.0623 & 0.0615 & 0.0609 & 0.0606 & 0.0607 & 0.0607 & \bf{0.0606} \\
\textbf{SSIM} 		& 0.9903 & 0.9914 & 0.9915 & 0.9916 & \bf{0.9916} & \bf{0.9916} & \bf{0.9916} & \bf{0.9916}  \\
\textbf{MS-SSIM} 	& 0.9984 & 0.9986 & \bf{0.9986} & \bf{0.9986} & \bf{0.9986} & \bf{0.9986} & \bf{0.9986}& \bf{0.9986}  \\
\textbf{UQI} 		& 0.9408 & 0.9460 & 0.9469 & 0.9476 & 0.9480 & 0.9479 & 0.9480 & \bf{0.9480}   \\
\textbf{VIQP}		& 0.9626 & 0.9635 & 0.9643 & 0.9659 & 0.9683 & \bf{0.9705} & 0.9702 & 0.9695   \\
\textbf{HPSI} 		& 0.9962 & 0.9970 & 0.9971 & \bf{0.9972} &\bf{0.9972} & \bf{0.9972} & \bf{0.9972} &  \bf{0.9972}    \\

			\bottomrule 
		\end{tabular}
		
	}

	\vspace{0.3cm}
\end{table*}

\begin{figure}
\begin{minipage}{\linewidth}
\resizebox{\linewidth}{!}{
\includegraphics[height=2.9cm]{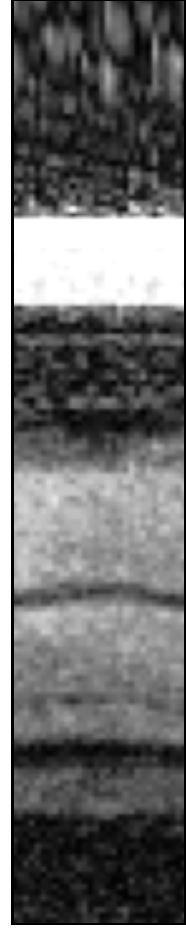}\hspace{-0.2cm}
\includegraphics[height=2.9cm]{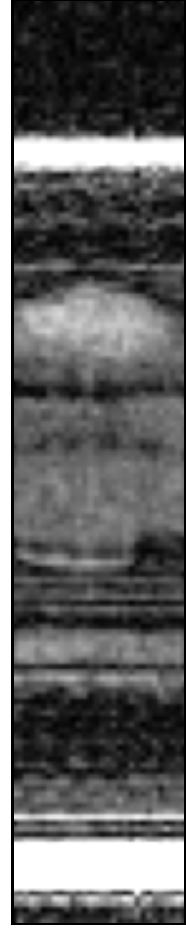}\hspace{-0.2cm}
\begin{overpic}[height=2.9cm,tics=10]{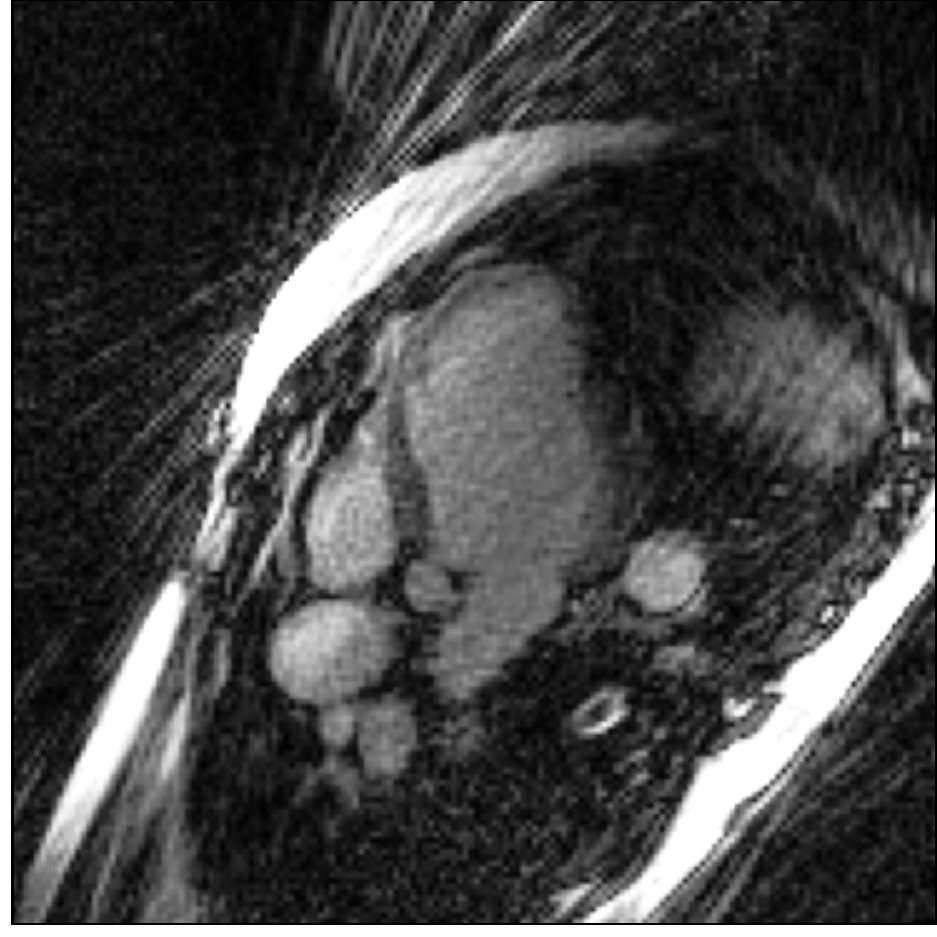}
\end{overpic} \hspace{-0.2cm}
\includegraphics[height=2.9cm]{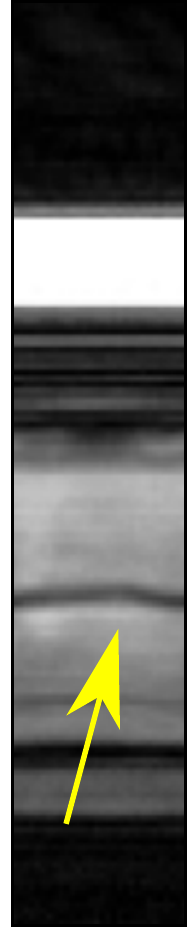}\hspace{-0.2cm}
\includegraphics[height=2.9cm]{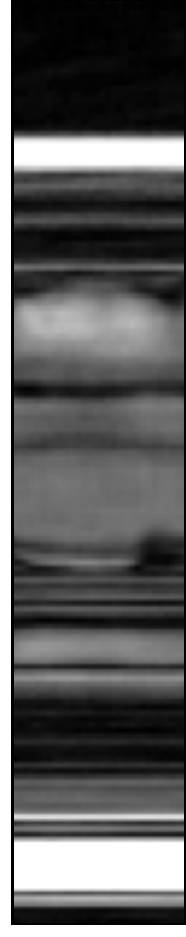}\hspace{-0.2cm}
\begin{overpic}[height=2.9cm,tics=10]{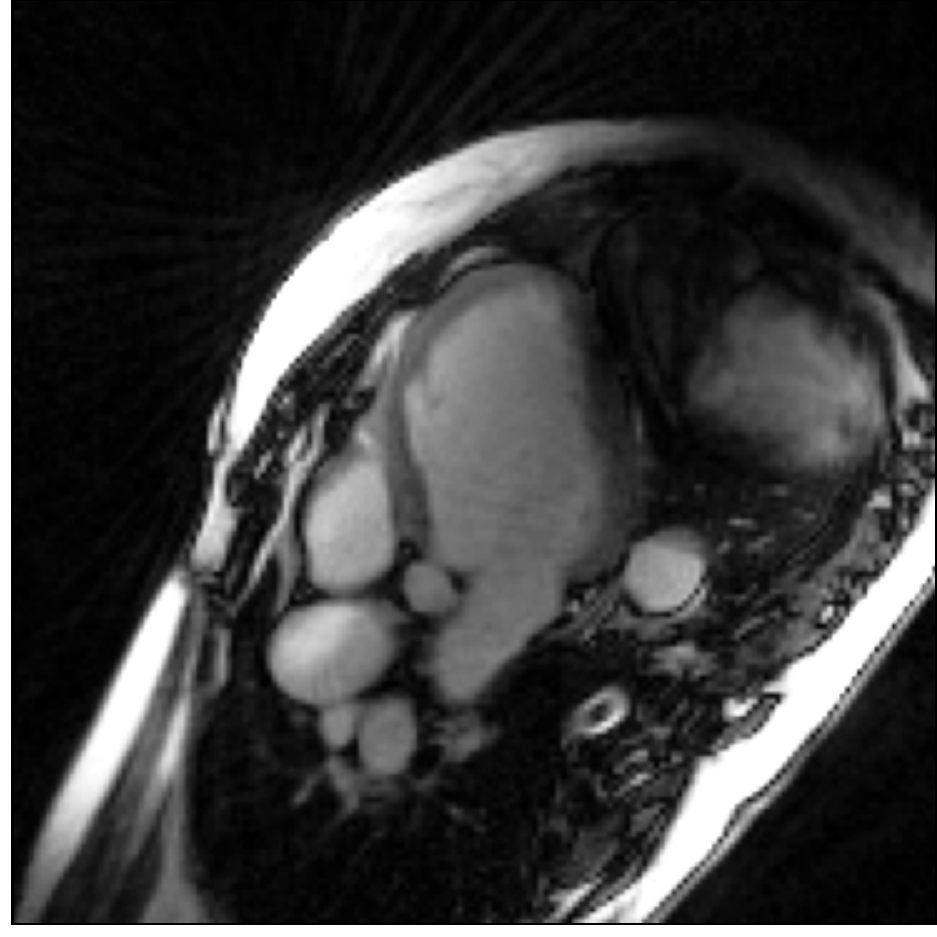}
\end{overpic} \hspace{-0.2cm}
\includegraphics[height=2.9cm]{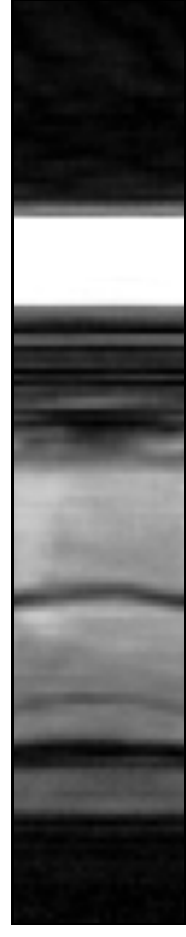}\hspace{-0.2cm}
\includegraphics[height=2.9cm]{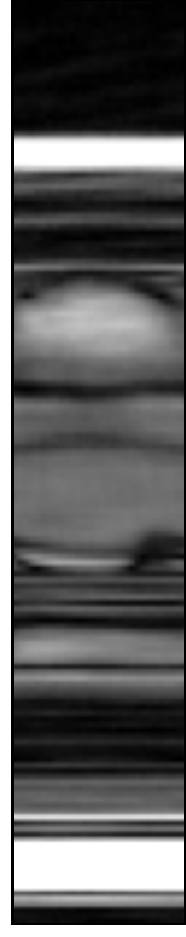}\hspace{-0.2cm}
\begin{overpic}[height=2.9cm,tics=10]{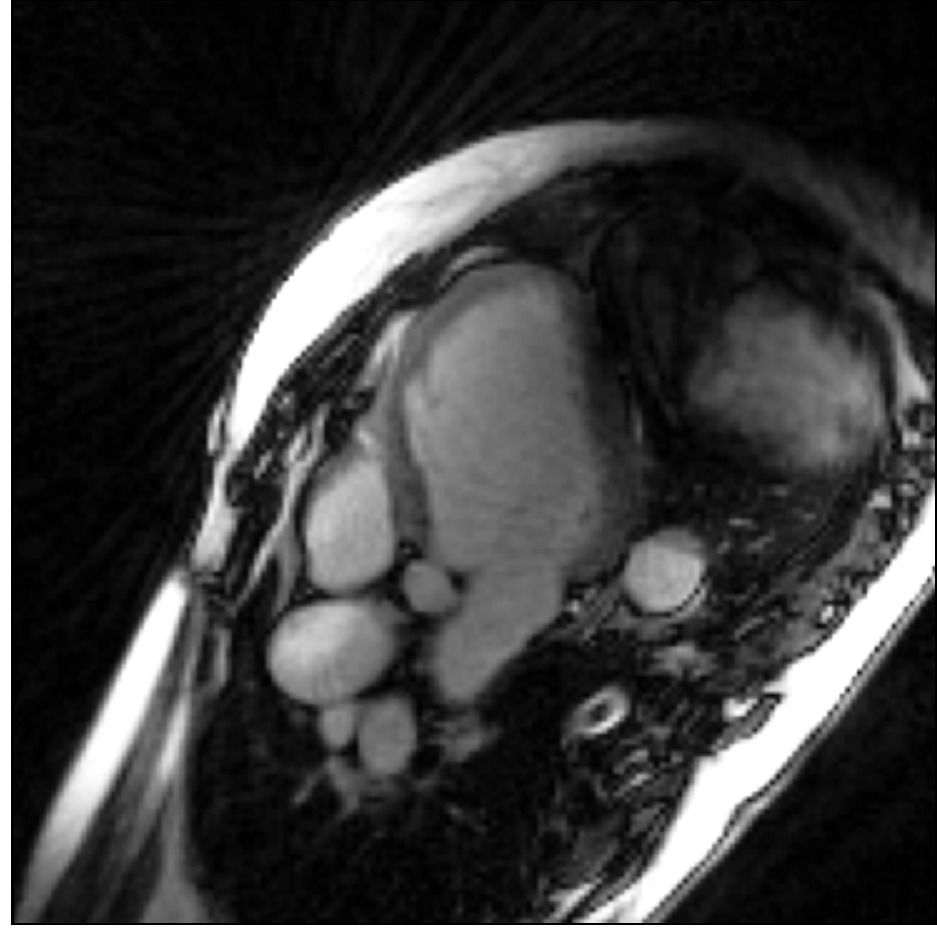}
\end{overpic} \hspace{-0.2cm}
\includegraphics[height=2.9cm]{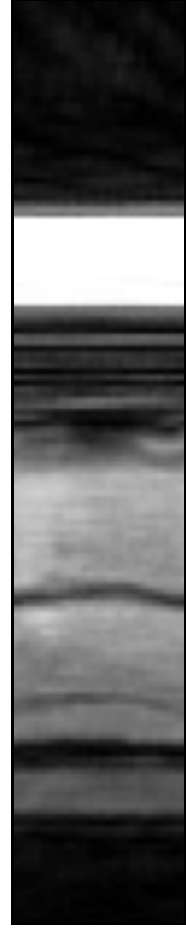}\hspace{-0.2cm}
\includegraphics[height=2.9cm]{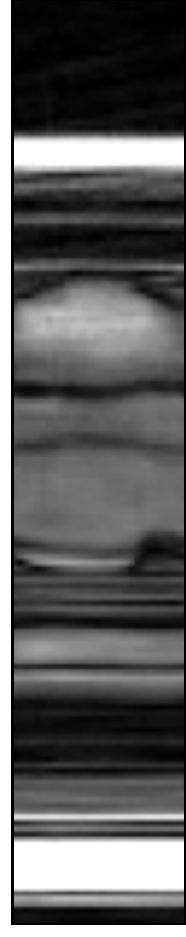}\hspace{-0.2cm}
\begin{overpic}[height=2.9cm,tics=10]{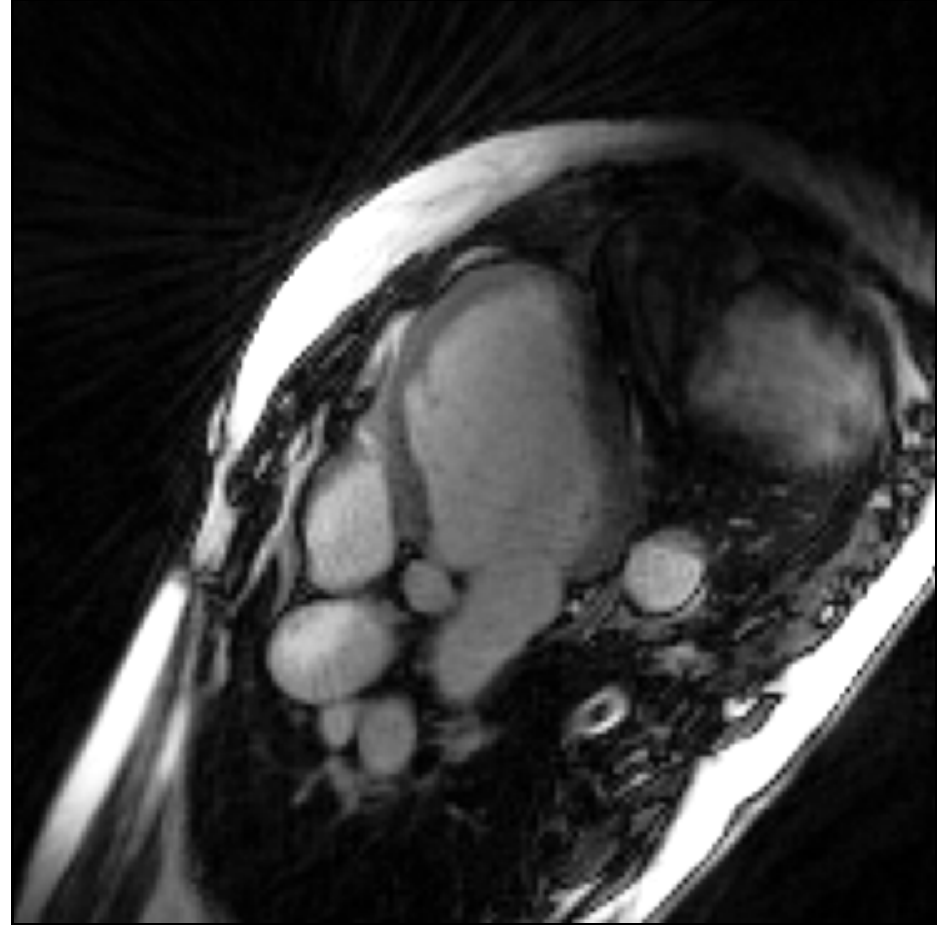}
\end{overpic} \hspace{-0.2cm}
}
\resizebox{\linewidth}{!}{
\includegraphics[height=2.9cm]{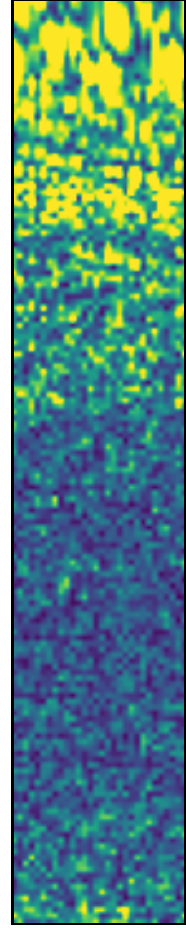}\hspace{-0.2cm}
\includegraphics[height=2.9cm]{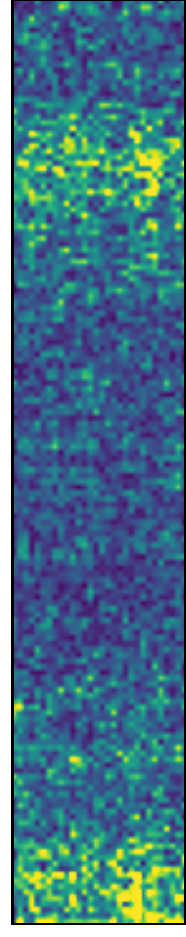}\hspace{-0.2cm}
\begin{overpic}[height=2.9cm,tics=10]{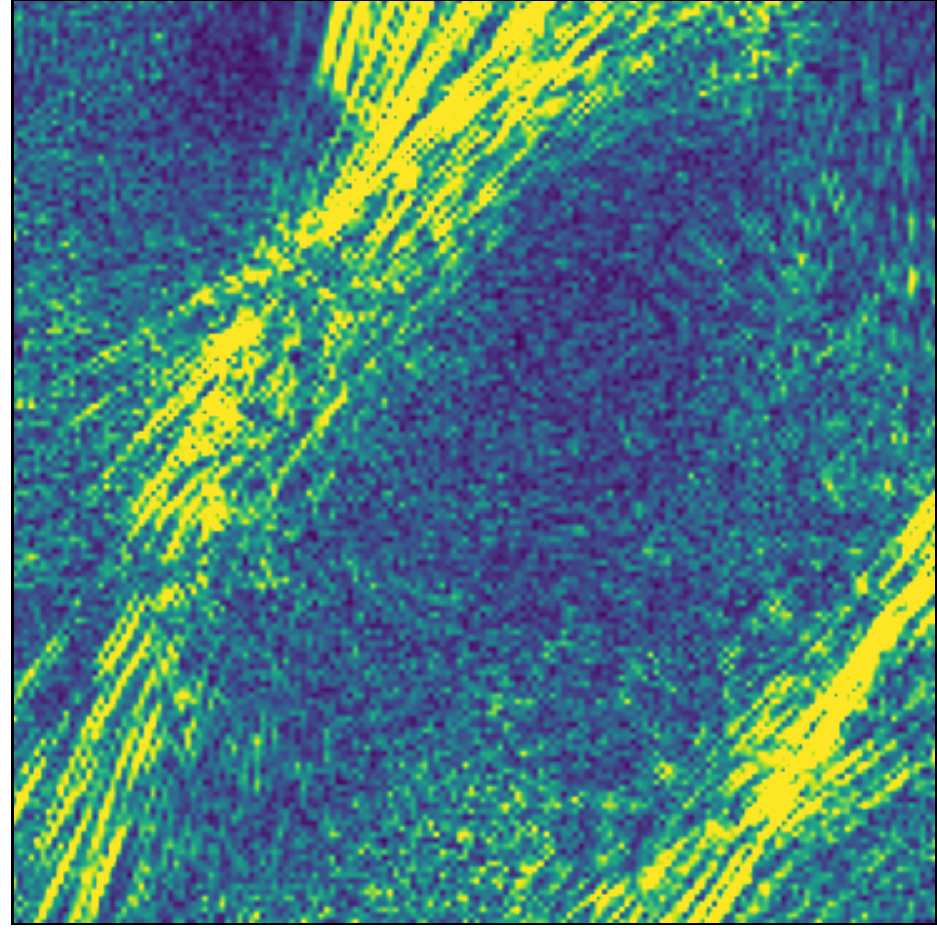}
\put (7,87) {\small\textcolor{white}{(A)}}
\end{overpic} \hspace{-0.2cm}
\includegraphics[height=2.9cm]{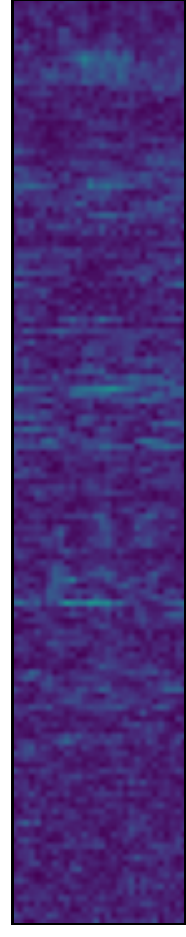}\hspace{-0.2cm}
\includegraphics[height=2.9cm]{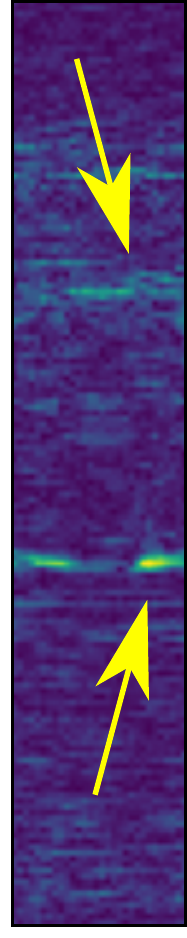}\hspace{-0.2cm}
\begin{overpic}[height=2.9cm,tics=10]{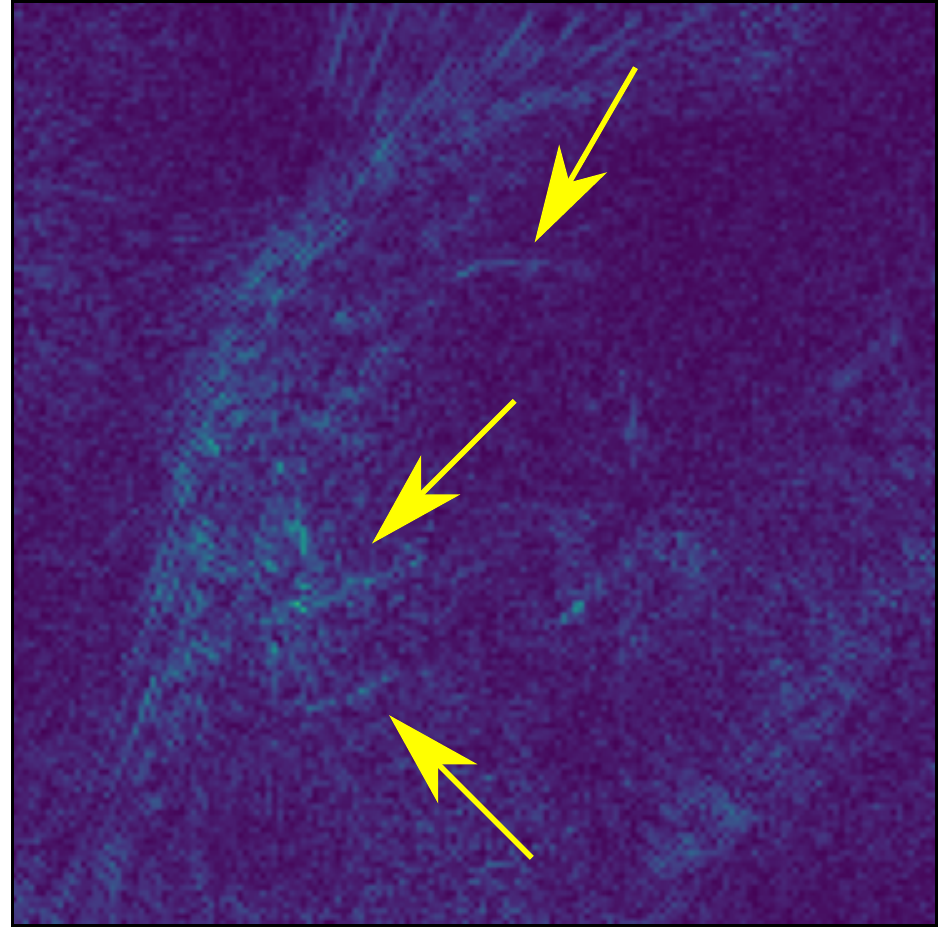}
\put (7,87) {\small\textcolor{white}{(B)}}
\end{overpic} \hspace{-0.2cm}
\includegraphics[height=2.9cm]{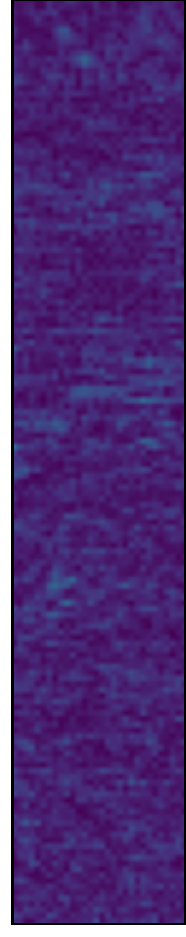}\hspace{-0.2cm}
\includegraphics[height=2.9cm]{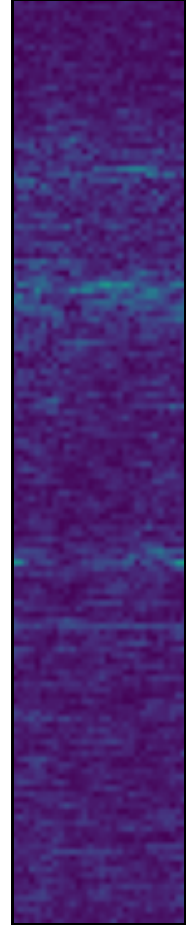}\hspace{-0.2cm}
\begin{overpic}[height=2.9cm,tics=10]{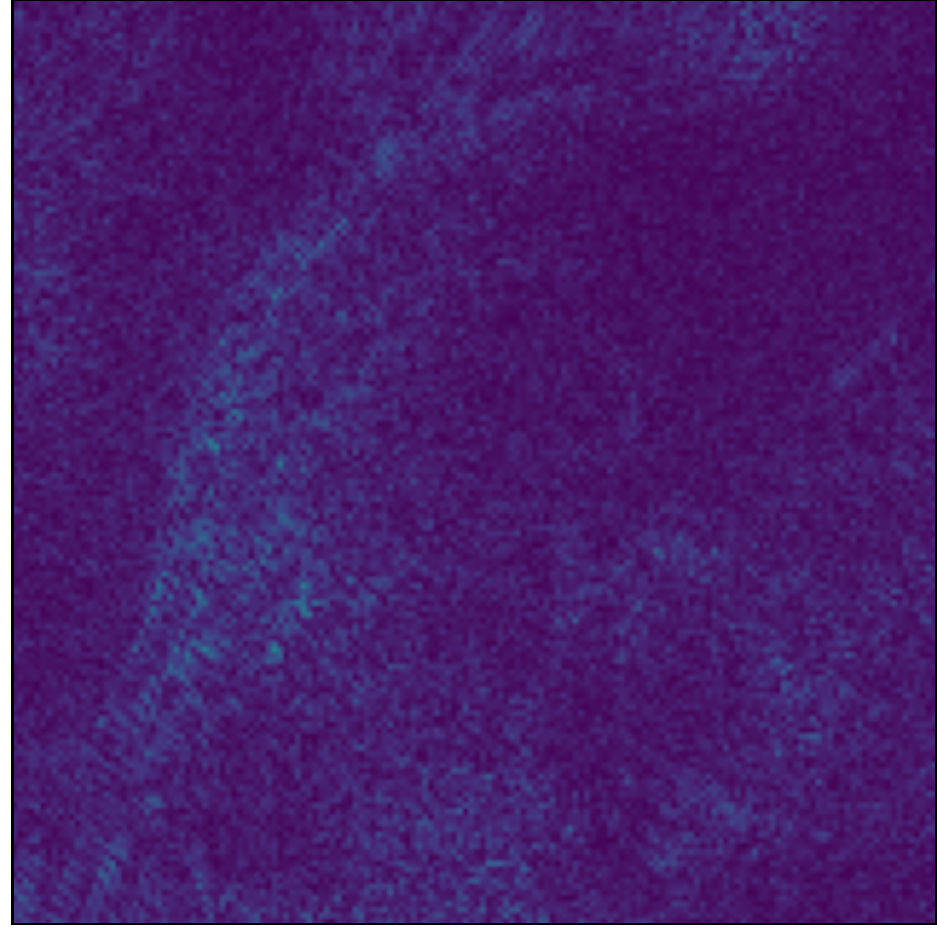}
 \put (7,87) {\small\textcolor{white}{(C)}}
\end{overpic} \hspace{-0.2cm}
\includegraphics[height=2.9cm]{images/results/training_stages_results/retrospective/Ntheta560/white_xt_yt.pdf}\hspace{-0.2cm}
\includegraphics[height=2.9cm]{images/results/training_stages_results/retrospective/Ntheta560/white_xt_yt.pdf}\hspace{-0.2cm}
\begin{overpic}[height=2.9cm,tics=10]{images/results/training_stages_results/retrospective/Ntheta560/white_xy.pdf}

 \put (7,87) {\small\textcolor{black}{(D)}}
\end{overpic} \hspace{-0.2cm}
}
\end{minipage}
\caption{Results and point-wise error-images for our proposed method, once fine-tuned with different $M=3$ and  $n_{\mathrm{CG}}=2$ (B) and once with $M=1$ and  $n_{\mathrm{CG}}=8$ (C). At test time, we used $M=12$ and $n_{\mathrm{CG}}=4$ for both methods. The network fine-tuned with $M=3$ and $n_{\mathrm{CG}}=2$ seems to  deliver slightly smoother results which lead to higher errors in regions where the cardiac motion is visible as indicated by the yellow arrows. The NUFFT-reconstruction and the ground-truth image are shown in (A) and (D), respectively.}\label{shallow_vs_deep_fig}\vspace{0.5cm}
\end{figure}

Figure \ref{shallow_vs_deep_fig} shows a comparison of the two best configurations of $M$ and $n_{\mathrm{CG}}$ for our proposed method. The figure well-reflects the measures in the sense that the point-wise error is clearly lower for the network fine-tuned with $M=1$ and $n_{\mathrm{CG}}=8$. Further, for the more shallow network fine-tuned $M=3$ and $n_{\mathrm{CG}}=2$, the results are clearly smoother in the regions where cardiac motion is visible. This comparison shows that sacrificing expressiveness of the CNN-block in order to be able to fine-tune with $M>1$ might not be necessary, see the discussion in the paper for more details. 

Table \ref{3dunet_table} shows a similar variation of $M$ and $n_{\mathrm{CG}}$ for the 3D U-Net. Interestingly, increasing $M$ while keeping  $n_{\mathrm{CG}}=4$ does not consistently improve the results. Thus, we attribute the possibility to increase $M$ at test time to the fact that in our proposed method, the forward model is part of the reconstruction network while training.

\begin{table*}[h]
	\centering
	\renewcommand{\arraystretch}{1.3}
	\footnotesize{
		\caption{Quantitative results for the 3D U-Net for which we also varied $M$ and $n_{\mathrm{CG}}$.} \label{3dunet_table}
		\vspace{0.2cm}
		\centering
		\begin{tabular}{ |l|
				c
				c
				c
				c
				c
				c
				c
				c|
			}
	\bottomrule 

	& \textbf{$M=1$} & \textbf{$M=2$} & \textbf{$M=3$} &\textbf{$M=4$}  &  \textbf{$M=6$} & \textbf{$M=8$} & \textbf{$M=10$}  & \textbf{$M=12$} \\
	& \textbf{$n_{\mathrm{CG}}=12$} & \textbf{$n_{\mathrm{CG}}=6$} & \textbf{$n_{\mathrm{CG}}=4$} &\textbf{$n_{\mathrm{CG}}=4$}  &  \textbf{$n_{\mathrm{CG}}=4$} & \textbf{$n_{\mathrm{CG}}=4$} & \textbf{$n_{\mathrm{CG}}=4$}  & \textbf{$n_{\mathrm{CG}}=4$} \\
			\midrule
& \multicolumn{8}{c|}{\textbf{Number of Radial Spokes:} $N_{\theta} = 560$}\\
\midrule	 
\textbf{PSNR} 		&46.0845 &\bf{46.1763} &46.0516 &46.0163 &45.8730 &45.6998 &45.5028 &45.2718   \\
\textbf{NRMSE} 		& 0.08 & \bf{0.0793} & 0.0803 & 0.0806 & 0.0818 & 0.0833 & 0.0851 & 0.0873  \\
\textbf{SSIM} 		& \bf{0.9880} & 0.9870 & 0.9869 & 0.9871 & 0.9870 & 0.9869 & 0.9866 & 0.9863\\
\textbf{MS-SSIM} 	&  \bf{0.9979} &0.9977 & 0.9976 & 0.9976 & 0.9976 & 0.9976 & 0.9976 & 0.9975\\
\textbf{UQI} 		& \bf{0.9288} & 0.9108 & 0.9112 & 0.9133 & 0.9144 & 0.9143 & 0.9137 & 0.9131\\
\textbf{VIQP}		& 0.934 & 0.9420 & 0.9419 & 0.9428 & 0.9438 & \bf{0.9441} & 0.9438 & 0.9429\\
\textbf{HPSI} 		& 0.9949 & \bf{0.9952} & 0.9950 & 0.9950 & 0.9949 & 0.9949 & 0.9948 & 0.9948\\

\midrule
& \multicolumn{8}{c|}{\textbf{Number of Radial Spokes:} $N_{\theta} = 1130$}\\
\textbf{PSNR} 		&\bf{47.5094} &47.3569 &47.2694 &47.2097  &47.0758 &46.9273  &46.7545 &46.5461  \\
\textbf{NRMSE} 		& \bf{0.0681} & 0.0693 & 0.0699 & 0.0704  & 0.0714 & 0.0725  & 0.0738 & 0.0755   \\
\textbf{SSIM} 		& \bf{0.9906} & 0.9895 & 0.9896 &0.9896  & 0.9896 & 0.9894  & 0.9893 & 0.9891   \\
\textbf{MS-SSIM} 	& \bf{0.9985} & 0.9982 & 0.9982 & 0.9982  & 0.9982 & 0.9981  & 0.9981 & 0.9981   \\
\textbf{UQI} 		& \bf{0.9420} & 0.9236 & 0.9251 & 0.9259  & 0.9261 & 0.9258  & 0.9255 & 0.9250  \\
\textbf{VIQP}		& 0.9547 & \bf{0.9558} & 0.9549 & 0.9550  & 0.9553 & 0.9552  & 0.9548 & 0.9541   \\
\textbf{HPSI} 		&  \bf{0.9966} &0.9965 & 0.9964 & 0.9964  & 0.9963 & 0.9963  & 0.9962 & 0.9963    \\

			\bottomrule 
		\end{tabular}
		
	}

	\vspace{0.3cm}
\end{table*}


\begin{thebibliography}{10}

\bibitem{puntmann2018society}
V.~O. Puntmann et~al.,
\newblock Society for Cardiovascular Magnetic Resonance (SCMR) expert consensus
  for CMR imaging endpoints in clinical research: part I-analytical validation
  and clinical qualification,
\newblock Journal of Cardiovascular Magnetic Resonance {\bf 20}, 67 (2018).

\bibitem{wang2018image}
G.~Wang, J.~C. Ye, K.~Mueller, and J.~A. Fessler,
\newblock Image reconstruction is a new frontier of machine learning,
\newblock IEEE Transactions on medical imaging {\bf 37}, 1289--1296 (2018).

\bibitem{ongie2020deep}
G.~Ongie, A.~Jalal, C.~A. M. R.~G. Baraniuk, A.~G. Dimakis, and R.~Willett,
\newblock Deep learning techniques for inverse problems in imaging,
\newblock IEEE Journal on Selected Areas in Information Theory  (2020).

\bibitem{jin2017deep}
K.~H. Jin, M.~T. McCann, E.~Froustey, and M.~Unser,
\newblock Deep convolutional neural network for inverse problems in imaging,
\newblock IEEE Transactions on Image Processing {\bf 26}, 4509--4522 (2017).

\bibitem{sandino2017deep}
C.~M. Sandino, N.~Dixit, J.~Y. Cheng, and S.~S. Vasanawala,
\newblock {Deep convolutional neural networks for accelerated dynamic magnetic
  resonance imaging},
\newblock Proceedings of 31st Conference of Neural Information Processing
  Systems (NIPS), Medical Imaigng meets NIPS Workshop.  (2017. Online.
  Available at
  \url{www.doc.ic.ac.uk/bglocker/public/mednips2017/mednips_2017_paper_19.pdf}).

\bibitem{Hauptmann2019}
A.~Hauptmann, S.~Arridge, F.~Lucka, V.~Muthurangu, and J.~A. Steeden,
\newblock Real-time cardiovascular MR with spatio-temporal artifact suppression
  using deep learning--proof of concept in congenital heart disease,
\newblock Magnetic Resonance in Medicine {\bf 81}, 1143--1156 (2019).

\bibitem{kofler2019spatio}
A.~Kofler, M.~Dewey, T.~Schaeffter, C.~Wald, and C.~Kolbitsch,
\newblock Spatio-temporal deep learning-based undersampling artefact reduction
  for 2D radial cine {MRI} with limited training data,
\newblock IEEE Transactions on Medical Imaging {\bf 39}, 703--717 (2020).

\bibitem{el2020multi}
H.~El-Rewaidy, A.~S. Fahmy, F.~Pashakhanloo, X.~Cai, S.~Kucukseymen, I.~Csecs,
  U.~Neisius, H.~Haji-Valizadeh, B.~Menze, and R.~Nezafat,
\newblock Multi-domain convolutional neural network (MD-CNN) for radial
  reconstruction of dynamic cardiac MRI,
\newblock Magnetic Resonance in Medicine  (2020).

\bibitem{hyun2018deep}
C.~M. Hyun, H.~P. Kim, S.~M. Lee, S.~Lee, and J.~K. Seo,
\newblock Deep learning for undersampled {MRI} reconstruction,
\newblock Physics in Medicine \& Biology {\bf 63}, 135007 (2018).

\bibitem{kofler2020neural}
A.~Kofler, M.~Haltmeier, T.~Schaeffter, M.~Kachelriess, M.~Dewey, C.~Wald, and
  C.~Kolbitsch,
\newblock Neural networks-based regularization for large-scale medical image
  reconstruction,
\newblock Physics in Medicine \& Biology {\bf 65}, 135003 (2020).

\bibitem{Antun201907377}
V.~Antun, F.~Renna, C.~Poon, B.~Adcock, and A.~C. Hansen,
\newblock On instabilities of deep learning in image reconstruction and the
  potential costs of AI,
\newblock Proceedings of the National Academy of Sciences  (2020).

\bibitem{maier2019learning}
A.~K. Maier, C.~Syben, B.~Stimpel, T.~W{\"u}rfl, M.~Hoffmann, F.~Schebesch,
  W.~Fu, L.~Mill, L.~Kling, and S.~Christiansen,
\newblock Learning with known operators reduces maximum error bounds,
\newblock Nature machine intelligence {\bf 1}, 373--380 (2019).

\bibitem{schlemper2017deep}
J.~Schlemper, J.~Caballero, J.~V. Hajnal, A.~N. Price, and D.~Rueckert,
\newblock A deep cascade of convolutional neural networks for dynamic {MR}
  image reconstruction,
\newblock IEEE Transactions on Medical Imaging {\bf 37}, 491--503 (2018).

\bibitem{hammernik2018learning}
K.~Hammernik, T.~Klatzer, E.~Kobler, M.~P. Recht, D.~K. Sodickson, T.~Pock, and
  F.~Knoll,
\newblock Learning a variational network for reconstruction of accelerated
  {MRI} data,
\newblock Magnetic Resonance in Medicine {\bf 79}, 3055--3071 (2018).

\bibitem{kobler2017variational}
E.~Kobler, T.~Klatzer, K.~Hammernik, and T.~Pock,
\newblock Variational networks: connecting variational methods and deep
  learning,
\newblock in {\em German conference on pattern recognition}, pages 281--293,
  Springer, 2017.

\bibitem{qin2018convolutional}
C.~Qin, J.~Schlemper, J.~Caballero, A.~N. Price, J.~V. Hajnal, and D.~Rueckert,
\newblock Convolutional recurrent neural networks for dynamic {MR} image
  reconstruction,
\newblock IEEE Transactions on Medical Imaging {\bf 38}, 280--290 (2018).

\bibitem{aggarwal2018modl}
H.~K. Aggarwal, M.~P. Mani, and M.~Jacob,
\newblock Modl: Model-based deep learning architecture for inverse problems,
\newblock IEEE Transactions on Medical Imaging {\bf 38}, 394--405 (2018).

\bibitem{qin2019k}
C.~Qin, J.~Schlemper, J.~Duan, G.~Seegoolam, A.~Price, J.~Hajnal, and
  D.~Rueckert,
\newblock k-t NEXT: Dynamic MR Image Reconstruction Exploiting Spatio-Temporal
  Correlations,
\newblock in {\em International Conference on Medical Image Computing and
  Computer-Assisted Intervention}, pages 505--513, Springer, 2019.

\bibitem{gilton2019neumann}
D.~Gilton, G.~Ongie, and R.~Willett,
\newblock Neumann networks for linear inverse problems in imaging,
\newblock IEEE Transactions on Computational Imaging {\bf 6}, 328--343 (2019).

\bibitem{kustner2020cinenet}
T.~K{\"u}stner et~al.,
\newblock CINENet: deep learning-based 3D cardiac CINE MRI reconstruction with
  multi-coil complex-valued 4D spatio-temporal convolutions,
\newblock Scientific reports {\bf 10}, 1--13 (2020).

\bibitem{schlemper2019nonuniform}
J.~Schlemper, S.~S.~M. Salehi, P.~Kundu, C.~Lazarus, H.~Dyvorne, D.~Rueckert,
  and M.~Sofka,
\newblock Nonuniform Variational Network: Deep Learning for Accelerated
  Nonuniform MR Image Reconstruction,
\newblock in {\em International Conference on Medical Image Computing and
  Computer-Assisted Intervention}, pages 57--64, Springer, 2019.

\bibitem{malave2020reconstruction}
M.~O. Malav{\'e}, C.~A. Baron, S.~P. Koundinyan, C.~M. Sandino, F.~Ong, J.~Y.
  Cheng, and D.~G. Nishimura,
\newblock Reconstruction of undersampled 3D non-Cartesian image-based
  navigators for coronary MRA using an unrolled deep learning model,
\newblock Magnetic Resonance in Medicine {\bf 84}, 800--812 (2020).

\bibitem{duan2019vs}
J.~Duan, J.~Schlemper, C.~Qin, C.~Ouyang, W.~Bai, C.~Biffi, G.~Bello,
  B.~Statton, D.~P. O’Regan, and D.~Rueckert,
\newblock VS-Net: Variable splitting network for accelerated parallel MRI
  reconstruction,
\newblock in {\em International Conference on Medical Image Computing and
  Computer-Assisted Intervention}, pages 713--722, Springer, 2019.

\bibitem{lustig2008compressed}
M.~Lustig, D.~L. Donoho, J.~M. Santos, and J.~M. Pauly,
\newblock Compressed sensing MRI,
\newblock IEEE Signal Processing magazine {\bf 25}, 72--82 (2008).

\bibitem{smith2019trajectory}
D.~S. Smith, S.~Sengupta, S.~A. Smith, and E.~Brian~Welch,
\newblock Trajectory optimized NUFFT: Faster non-Cartesian MRI reconstruction
  through prior knowledge and parallel architectures,
\newblock Magnetic resonance in medicine {\bf 81}, 2064--2071 (2019).

\bibitem{knoll2020deep}
F.~Knoll, K.~Hammernik, C.~Zhang, S.~Moeller, T.~Pock, D.~K. Sodickson, and
  M.~Akcakaya,
\newblock Deep-learning methods for parallel magnetic resonance imaging
  reconstruction: A survey of the current approaches, trends, and issues,
\newblock IEEE Signal Processing Magazine {\bf 37}, 128--140 (2020).

\bibitem{winkelmann2006optimal}
S.~Winkelmann, T.~Schaeffter, T.~Koehler, H.~Eggers, and O.~Doessel,
\newblock An optimal radial profile order based on the Golden Ratio for
  time-resolved {MRI},
\newblock IEEE Transactions on Medical Imaging {\bf 26}, 68--76 (2006).

\bibitem{qu2005convergence}
P.~Qu, K.~Zhong, B.~Zhang, J.~Wang, and G.~X. Shen,
\newblock Convergence behavior of iterative SENSE reconstruction with
  non-Cartesian trajectories,
\newblock Magnetic Resonance in Medicine: An Official Journal of the
  International Society for Magnetic Resonance in Medicine {\bf 54}, 1040--1045
  (2005).

\bibitem{le1991mpeg}
D.~Le~Gall,
\newblock MPEG: A video compression standard for multimedia applications,
\newblock Communications of the ACM {\bf 34}, 46--58 (1991).

\bibitem{Tsao2003}
J.~Tsao, P.~Boesiger, and K.~P. Pruessmann,
\newblock {k-t BLAST and k-t SENSE: Dynamic {MRI} With High Frame Rate
  Exploiting Spatiotemporal Correlations},
\newblock Magnetic Resonance in Medicine  (2003).

\bibitem{Ronneberger2015}
O.~Ronneberger, P.~Fischer, and T.~Brox,
\newblock U-net: Convolutional networks for biomedical image segmentation,
\newblock in {\em International Conference on Medical image computing and
  computer-assisted intervention}, pages 234--241, Springer, 2015.

\bibitem{hestenes1952methods}
M.~R. Hestenes et~al.,
\newblock Methods of conjugate gradients for solving linear systems,
\newblock Journal of research of the National Bureau of Standards {\bf 49},
  409--436 (1952).

\bibitem{feng_mrm_2012}
L.~Feng, M.~B. Srichai, R.~P. Lim, A.~Harrison, W.~King, G.~Adluru, E.~V.~R.
  Dibella, D.~K. Sodickson, R.~Otazo, and D.~Kim,
\newblock {Highly accelerated real-time cardiac cine MRI using k-t
  SPARSE-SENSE.},
\newblock Magnetic Resonance Imaging  (2012).

\bibitem{malik2005gridding}
W.~Q. Malik, H.~A. Khan, D.~J. Edwards, and C.~J. Stevens,
\newblock A gridding algorithm for efficient density compensation of
  arbitrarily sampled Fourier-domain data,
\newblock in {\em IEEE/Sarnoff Symposium on Advances in Wired and Wireless
  Communication, 2005.}, pages 125--128, IEEE, 2005.

\bibitem{wang2004image}
Z.~Wang, A.~C. Bovik, H.~R. Sheikh, and E.~P. Simoncelli,
\newblock Image quality assessment: from error visibility to structural
  similarity,
\newblock IEEE Transactions on Image Processing {\bf 13}, 600--612 (2004).

\bibitem{wang2003multiscale}
Z.~Wang, E.~P. Simoncelli, and A.~C. Bovik,
\newblock Multiscale structural similarity for image quality assessment,
\newblock in {\em The Thrity-Seventh Asilomar Conference on Signals, Systems \&
  Computers, 2003}, volume~2, pages 1398--1402, Ieee, 2003.

\bibitem{wang2002universal}
Z.~Wang and A.~C. Bovik,
\newblock A universal image quality index,
\newblock IEEE Signal Processing Letters {\bf 9}, 81--84 (2002).

\bibitem{sheikh2006image}
H.~R. Sheikh and A.~C. Bovik,
\newblock Image information and visual quality,
\newblock IEEE Transactions on image processing {\bf 15}, 430--444 (2006).

\bibitem{reisenhofer2018haar}
R.~Reisenhofer, S.~Bosse, G.~Kutyniok, and T.~Wiegand,
\newblock A {H}aar wavelet-based perceptual similarity index for image quality
  assessment,
\newblock Signal Processing: Image Communication {\bf 61}, 33--43 (2018).

\bibitem{Muckley2019}
M.~e.~a. Muckley,
\newblock Torch KB-NUFFT,
\newblock \url{https://github.com/mmuckley/torchkbnufft}, 2019.

\bibitem{Muckley2020}
M.~J. Muckley, R.~Stern, T.~Murrell, and F.~Knoll,
\newblock {TorchKbNufft}: A High-Level, Hardware-Agnostic Non-Uniform Fast
  Fourier Transform,
\newblock in {\em ISMRM Workshop on Data Sampling \& Image Reconstruction},
  2020.

\bibitem{kofler2018u}
A.~Kofler, M.~Haltmeier, C.~Kolbitsch, M.~Kachelrie{\ss}, and M.~Dewey,
\newblock {A U-nets cascade for sparse view computed tomography},
\newblock in {\em International Workshop on Machine Learning for Medical Image
  Reconstruction}, page 91–99, Springer, 2018.

\bibitem{pruessmann2001advances}
K.~P. Pruessmann, M.~Weiger, P.~B{\"o}rnert, and P.~Boesiger,
\newblock Advances in sensitivity encoding with arbitrary k-space trajectories,
\newblock Magnetic Resonance in Medicine: An Official Journal of the
  International Society for Magnetic Resonance in Medicine {\bf 46}, 638--651
  (2001).

\bibitem{block2007}
K.~T. Block, M.~Uecker, and J.~Frahm,
\newblock Undersampled radial {MRI} with multiple coils. {I}terative image
  reconstruction using a total variation constraint,
\newblock Magnetic Resonance in Medicine {\bf 57}, 1086--1098 (2007).

\bibitem{wang2014compressed}
Y.~Wang and L.~Ying,
\newblock Compressed sensing dynamic cardiac cine {MRI} using learned
  spatiotemporal dictionary,
\newblock IEEE Transactions on Biomedical Engineering {\bf 61}, 1109--1120
  (2014).

\bibitem{caballero2014dictionary}
J.~Caballero, A.~N. Price, D.~Rueckert, and J.~V. Hajnal,
\newblock Dictionary learning and time sparsity for dynamic MR data
  reconstruction,
\newblock IEEE Transactions on Medical Imaging {\bf 33}, 979--994 (2014).

\bibitem{pali2020}
M.~C. Pali, T.~Schaeffter, C.~Kolbitsch, and A.~Kofler,
\newblock Adaptive Sparsity Level and Dictionary Size Estimation for Image
  Reconstruction in Accelerated 2D Radial Cine MRI,
\newblock Medical Physics  (In Press, 2020).

\bibitem{schwab2018deep}
J.~Schwab, S.~Antholzer, and M.~Haltmeier,
\newblock Deep null space learning for inverse problems: convergence analysis
  and rates,
\newblock Inverse Problems {\bf 35}, 025008 (2019).

\bibitem{biswas2019dynamic}
S.~Biswas, H.~K. Aggarwal, and M.~Jacob,
\newblock Dynamic MRI using model-based deep learning and SToRM priors:
  MoDL-SToRM,
\newblock Magnetic resonance in medicine {\bf 82}, 485--494 (2019).

\end{thebibliography}
\end{document}